\documentclass[
	a4paper,
	abstracton,
	emulatestandardclasses
]{scrartcl}

\usepackage[english]{babel}
\usepackage{xcolor}
\usepackage{a4wide}
\usepackage{authblk}
\usepackage{amsthm}
\usepackage{amsmath}
\usepackage{amssymb}
\usepackage{physics}
\usepackage{csquotes}
\usepackage{paralist}
\usepackage[
	backend=bibtex,
	sorting=none
]{biblatex}
\usepackage{graphicx}
\usepackage{hyperref}
\usepackage{cleveref}
\usepackage[binary-units]{siunitx}
\usepackage[
	margin=10pt,
	labelfont=bf
]{caption}
\usepackage{booktabs}
\usepackage{pgfplots}
\usepackage{adjustbox}
\usepackage[acronym]{glossaries}

\title{MRI to CT Translation with GANs}
\author[1]{Bodo Kaiser}
\author[2]{Shadi Albarqouni}
\affil[1]{\textit{bodo.kaiser@physik.uni-muenchen.de}}
\affil[2]{\textit{shadi.albarqouni@tum.de}}

\DeclareMathOperator{\mae}{MAE}
\DeclareMathOperator{\mse}{MSE}
\DeclareMathOperator{\gdl}{GDL}
\DeclareMathOperator{\psnr}{PSNR}

\pgfplotsset{compat=newest}
\glsdisablehyper
\newacronym{ct}{CT}{X-ray computed tomography}
\newacronym{mri}{MRI}{magnetic resonance imaging}
\newacronym{pd}{PD}{proton density}
\newacronym{t1}{T1}{spin-lattice relaxation time}
\newacronym{t2}{T2}{spin-spin relaxation time}
\newacronym{mp}{MP}{magnetization-prepared}
\newacronym{itk}{ITK}{Insight Segmentation and Registration Toolkit}
\newacronym{rage}{RAGE}{RF pulse and rapid gradient echo}
\newacronym{mhd}{MHD}{MetaImage Header}
\newacronym{pet}{PET}{positron emission tomography}
\newacronym{rire}{RIRE}{Retrospective Image Registration Evaluation}
\newacronym{nifti}{NIfTI}{Neuroimaging Informatics Technology Institute}
\newacronym{oasis}{OASIS}{Open Access Series of Imaging Studies}
\newacronym{adni}{ADNI}{Alzheimer's Disease Neuroimaging Initiative}
\newacronym{mae}{MAE}{mean absolute error}
\newacronym{mse}{MSE}{mean squared error}
\newacronym{gan}{GAN}{generative adversarial network}
\newacronym{gdl}{GDL}{gradient distance loss}
\newacronym{psnr}{PSNR}{peak signal-to-noise ratio}
\newacronym{cnn}{CNN}{convolutional neural network}

\newcommand\inputpgf[2]{{
\let\pgfimageWithoutPath\pgfimage
\renewcommand{\pgfimage}[2][]{\pgfimageWithoutPath[##1]{#1/##2}}
\input{#1/#2}
}}

\begin{document}

\makeatletter
\maketitle
\begin{abstract}
  We present a detailed description and reference implementation of
  preprocessing steps necessary to prepare the public \gls{rire} dataset
  for the task of \gls{mri} to \gls{ct} translation. Furthermore we describe
  and implement three state of the art \gls{cnn} and \gls{gan} models where
  we report statistics and visual results of two of them.
\end{abstract}

\makeatother

\section{Introduction}

\gls{ct} and \gls{mri} are the essential medical imaging modalities for
clinical diagnosis and cancer monitoring. Inside the clinical framework
\gls{mri} is the more informative and safer modality~\cite{Hartwig09}. Instead
of x-rays which are known to contribute to carcinogenesis~\cite{Martin06},
\gls{mri} exploits the magnetic properties of the hydrogen nucleus and is not
associated with to have negative impact on the patients health. In addition
\gls{mri} provides more detailed visual information on soft tissue. These
benefitial characteristics suggest that \gls{mri} supersedes \gls{ct} in the
long-term. One of many remaining obstacles is, however, the requirement of
\gls{ct} for image guided radiation therapy planning. Although \gls{mri} and
\gls{ct} differ significant in the applied physics, the high entropy of
\gls{mri} data suggests the existence of a surjective transform from
\gls{mri} to \gls{ct} space. With the recent advents in computer vision
techniques based on \gls{gan} we seem to close to finding such a mapping
emprically.

\section{Related Work}

Since the early days of \gls{ct}, health manufacturer were attempted to reduce
radiation exposure in \gls{ct} scans by using, for instance, more sensible
detection electronics, and more sophisticated scanning sequences. Through the
growing availability of computing power we also find evermore computer vision
techniques being utilized, for example, in the enhancement of image quality of
low-dose \gls{ct}s~\cite{Xu12}. Altough these efforts have lead to an
impressive and steady evolution of the \gls{ct} apparatus, they still require
the patient to be irradiated nevertheless.
First approaches which dispense the radiation exposure, through the
computational transformation of \gls{mri} to \gls{ct}, relay on the
atlas-based transformations applied to \gls{mri} to predict \gls{ct}, see
Ref.~\cite{Hofmann08}. Further improvements thereto include, for instance,
random forests~\cite{Andreasen13}. Finally it has been shown that these
\gls{ct} prediction methods can in fact already replace physical \gls{ct}
for treatment planning~\cite{Andreasen2017}.
At the same time, we have seen an incredible progress with deep learning
techniques in computer science~\cite{LeCun15}. Recent efforts with \gls{gan}s,
see Ref.~\cite{Goodfellow14}, seem to be a promising path towards finding
a global optimum in training neural networks through the use of game theory.
Furthermore \gls{gan}s proved significant improvements over the former state
of art in the task of image to image translation~\cite{Isola16} but also the
generalization of three dimensional structures inside the so called latent
space~\cite{ZXFT16}.
Keeping this in mind, the medical computer vision community rapidly adapted
\gls{gan}s for their own specific tasks. In comparison to datasets common in
general computer vision, medical datasets typical comprise volumetric single
channel images with high bit depth. Bearing the challenge of \gls{ct} from
\gls{mri} prediction in mind, the expectations towards \gls{gan}s have been
lately shown increased performance to the previous approaches~\cite{Nie16}.
Yet, the full potential of \gls{gan}s have not been exhausted. For example,
it has been shown that \gls{gan}s are capable of being trained with
unregistered modalities~\cite{Wolterink17}.
Beside the enourmous breakthroughs made in medical computer vision we still
see a shortage in a reproducable comparison of recent methods with publicly
available data. Not to mention the open questions with regard to best
practices in choosing good \gls{gan} model parameters for the task of
\gls{ct} prediction which we hope to address in the subsequent sections.

\section{Methods}

\subsection{Dataset}

Though many datasets involving \gls{mri} and \gls{ct} data exist, for instance,
\gls{oasis}~\cite{OASIS} or \gls{adni}~\cite{ADNI}, public datasets in which
both modalities are obtainable for the same subject are, to date, rare. To our
knowledge only the \gls{rire} project~\cite{RIRE} and the Cancer Imaging
Archive~\cite{CIA} provide \gls{mri} and \gls{ct} from the same subject. For
the present work we used the data from the \gls{rire} project because it uses
an uniform data fromat.
\begin{table}[h]
  \centering
  \begin{tabular}{*{6}{c}}
    \toprule
    \acrshort{ct} &
		\acrshort{mri} \acrshort{pd} &
		\acrshort{mri} \acrshort{t1} &
		\acrshort{mri} \acrshort{t2} &
		\acrshort{mri} \acrshort{mp} \acrshort{rage} &
		\acrshort{pet} \\
    \midrule
    \num{17} & \num{14} & \num{19} & \num{18} & \num{9} & \num{8} \\
             & \num{12} & \num{17} & \num{16} & \num{9} & \num{6} \\
    \bottomrule
  \end{tabular}
  \caption{Subject counts of the \gls{rire} dataset with respect to the
    available imaging modalities. In the second table row we only consider
    subjects with \gls{ct} data present.
  }\label{tab:rire}
\end{table}
In \Cref{tab:rire} we list the aggregated modality count of the \gls{rire}
dataset in the first row. In the second row we list the aggregated modality
count for the subjects with \gls{ct} modality available. Beside of \gls{ct}
one can also obtain \gls{pet} images for some subjects. Next to the common
\gls{t1} and \gls{t2} weighted \gls{mri}, some subjects of the \gls{rire}
dataset also offer \gls{pd} and \gls{mp} \gls{rage} weighted \gls{mri}s. Some
\gls{mri}s can be obtained in a rectified version, which we did not use. We
used the \gls{t1} weighted \gls{mri} together with the \gls{ct} as input and
target data as these give us the highest subject count. However, it would be
an interesting experiment to supply different \gls{mri}s as multi-channel
input.

\subsection{Preprocessing}

\begin{figure}[h]
  \centering
  \includegraphics[page=1,width=.8\linewidth]{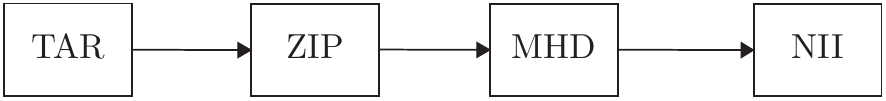}
  \caption{Image extraction and conversion from \gls{mhd} to \gls{nifti}
		format.
	}\label{fig:conversion}
\end{figure}

\begin{figure}[h]
  \centering
  \includegraphics[page=2,width=.8\linewidth]{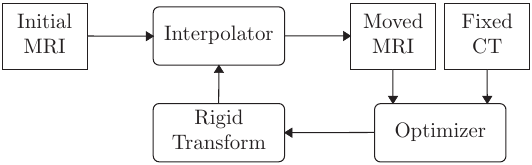}
  \caption{Multi-modal image coregistration using maximum mutual information
    optimization.
	}\label{fig:registration}
\end{figure}

The modality data for each subject can be downloaded from the website of the
\gls{rire} project, see Ref.~\cite{RIRE}. In \Cref{fig:conversion} we depicted
the first preprocessing protocol. It involves the extraction, decompression and
conversion of the volumetric data. After extraction and decompression the
volumetric data presents itself as \gls{mhd}. We converted the \gls{mhd} files
to the self-contained \gls{nifti} format through the Python front-end of the
\gls{itk} library.
\Cref{fig:registration} illustrates the coregistration procedure that follows
the first preprocessing procedure. The coregistration yields a rigid
transformation that aligns the moving volume with the fixed volume. Given a
rigid transformation, a linear interpolator returns a translated volume from
the sample points of the initial moving volume. The mutual information between
the moved \gls{mri} and the \gls{ct} is then used to optimize the
rigid transformation. This procedure is executed iteratively and stopped when
the maximum iterations steps are reached or the convergence condition is met.
As the implementation of the interpolator and the transformation optimizer
are complex, we used the registration toolset included in the \gls{itk}
library, see Ref.~\cite{Yaniv2018}.
For the present work we choose the \gls{ct} volume to be fixed, as the
\gls{ct} volumes are in general spatially normalized accross different
subjects. Because the different modalities have in general different
resolutions, we found that for the lower resolution modality the
coregistration produced artifacts at the boundaries of the transverse plane.
We manually removed these slices from the dataset.
Following the coregistration, we used the binary fill holes algorithm from
SciPy~\cite{SciPy} as an attempt to remove the \gls{ct} table present in some
\gls{ct} volumes as well as background noise.\footnote{The complete
preprocessing described so far is available at
\url{https://github/bodokaiser/mrtoct-scripts}.} Finally we converted the
preprocessed pairs of \gls{mri} and \gls{ct} volumes to the tfrecord format
in order to easily read the data into Tensorflow~\cite{Tensorflow15}. As part
of the data pipeline implemented with Tensorflow we perform a pad or crop to
either $384\times384$ for transverse 2D slices. In the 3D case we performed
patch extraction of target shape $32\times32\times32$ for \gls{mri} and
$16\times16\times16$ for \gls{ct}. The target shape for the 2D slices was
choosen as a compromise between compatibility with the convolution parameters
and reducing crop on the volumes. Furtheremore we applied a
min-max-normalization in order to keep floating range arthimetic in a range
of $[0,1]$.

\subsection{Models}

We attempted to implement three different neural network models for the
\gls{mri} to \gls{ct} synthesis task. The first and most simple model is based
on the popular u-net model~\cite{Ronneberger15} in combination with a
standard error metric, i.e.\ \gls{mae} and \gls{mse}. The second model is
based on pix2pix~\cite{Isola16}, which already has proven great success in the
task of image translation. It uses a u-net based model as generator in
addition to a simple discriminator model to calculate the adversarial loss. As
third and last model we attempted an implementation of the context-aware 3D
synthesis GAN from Nie~\cite{Nie16}. Unfortunately we found our implementation
of the patch reconstruction to be too resource intensive for practical
purpose. In comparison to the other two models, which operate on the
transverse 2D slices of the brain, it is applied to 3D patches. Training and
interferen were implemented using the Tensorflow~\cite{Tensorflow15}
framework.\footnote{The implementation is available at
\url{https://github/bodokaiser/mrtoct-tensorflow}.}

\subsubsection{u-net}

The original u-net model~\cite{Ronneberger15} was developed for the
segmentation of biomedical images. A central concept of the architecture is to
combine the capture of context and precise localization through interconnected
layers.
\begin{figure}[h]
  \centering
  \includegraphics[page=3,width=.8\linewidth]{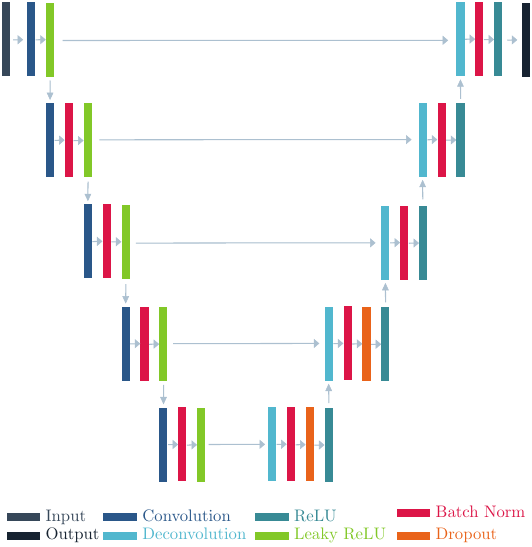}
  \caption{The u-net architecture.
  }\label{fig:unet:gen}
\end{figure}
In \Cref{fig:unet:gen} the u-net architecture is illustrated. We remark the
two paths of data flow: for one the image is passed through a sequence of
encoders and decoders, then again data can flow from one encoder stage
directly to the corresponding decoder stage. The encoder encode localized
features while the decoders decode an image from the previous layer and the
corresponding encoder stage.
We adapted the specific u-net based architecture from pixtopix. In comparison
to the original formulation the number convolution layers are reduced and
the max pooling in the decoder blocks were replaced by deconvolution (also known
as transposed convolution) layers. Furthermore we used Leaky ReLUs in the
encoders instead of usual ReLUs. Except for the first encoder we applied batch
normalization inbetween the convolution and activation layers. Another
difference relative to the original scheme there is a use of dropout layers
after the first and second decoder. Dropout layers are known to improve
network generalization by randomly suppressing features from the training
process~\cite{Srivastava2014}.
\begin{table}[h]
  \centering
  \begin{tabular}{cccc}
    \toprule
    Type & Kernel & Strides & Output Shape \\
    \midrule
    Input & & & \num{384x384x1} \\ 
    Convolution & \num{4x4} & \num{2} & \num{192x192x64} \\
    Convolution & \num{4x4} & \num{2} & \num{96x96x128} \\
    Convolution & \num{4x4} & \num{2} & \num{48x48x256} \\
    Convolution & \num{4x4} & \num{2} & \num{24x24x512} \\
    Convolution & \num{4x4} & \num{2} & \num{12x12x512} \\
    Deconvolution & \num{4x4} & \num{2} & \num{24x24x512} \\
    Deconvolution & \num{4x4} & \num{2} & \num{48x48x512} \\
    Deconvolution & \num{4x4} & \num{2} & \num{96x96x256} \\
    Deconvolution & \num{4x4} & \num{2} & \num{192x192x128} \\
    Deconvolution & \num{4x4} & \num{2} & \num{384x384x64} \\
    Deconvolution & \num{3x3} & \num{1} & \num{384x384x1} \\
    Output & & & \num{384x384x1} \\ 
    \bottomrule
  \end{tabular}
  \caption{Network parameters used in the u-net.
  }\label{tab:unet:gen}
\end{table}
\Cref{fig:unet:gen} lists the network parameters used for our u-net
architecture. The kernel paremeter specifies the shape of the convolution
kernel, the stride parameter describes the spacing between convolutions.
Weight initialization was performed using Xavier, see Ref.~\cite{Xavier2010},
if not noted otherwise.

\subsubsection{pixtopix}

The pixtopix model uses the previously introduced u-net architecture as
generator to translate an input \gls{mri} to \gls{ct}. In addition, pixtopix
utilizes a second network, the discriminator network, to output a score map
that distinguishes between real and fake \gls{ct}, wherein the term real
\gls{ct} corresponds to a \gls{ct} propably obtained from the ground truth
and fake \gls{ct} correspond to a probable output of the generation
In this sense one is able to add an adversarial loss term to the
standard metric loss, that maximizes the identification of real \gls{ct}s
while minimizing the missidentification of fake \gls{ct}s as real
ones~\cite{Goodfellow14}.
The pixtopix model has proven great success as general purpose solution
for translation experiments with color images~\cite{Isola16}. Recently
pixtopix was extended to support even training on unpaired
data~\cite{Zhu2017}. This approach has also successfully been applied to the
task of \gls{mri} to \gls{ct} translation~\cite{Wolterink17}.
\begin{figure}[h]
  \centering
  \includegraphics[page=4,width=.8\linewidth]{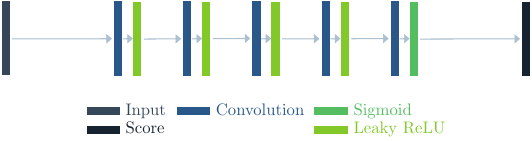}
  \caption{The pixtopix discriminator architecture.
	}\label{fig:pixtopix:disc}
\end{figure}
\Cref{fig:pixtopix:disc} depicts the pixtopix discriminator architecture. It
consists of five convolution layers with non-linear activation function. The
first four activation functions are Leaky ReLUs while the last one is of type
sigmoid.
\begin{table}[h]
  \centering
  \begin{tabular}{cccc}
    \toprule
    Type & Kernel & Strides & Output Shape \\
    \midrule
    Input & & & \num{384x384x2} \\ 
    Convolution & \num{4x4} & \num{2} & \num{192x192x64} \\
    Convolution & \num{4x4} & \num{2} & \num{96x96x128} \\
    Convolution & \num{4x4} & \num{2} & \num{48x48x256} \\
    Convolution & \num{4x4} & \num{2} & \num{24x24x512} \\
    Convolution & \num{4x4} & \num{1} & \num{1x24x512} \\
    Output & & & \num{1x24x512} \\ 
    \bottomrule
  \end{tabular}
  \caption{Network parameters used in the pixtopix discriminator.
  }\label{tab:pixtopix:disc}
\end{table}
\Cref{tab:pixtopix:disc} lists the network parameters used for the pixtopix
discriminator network. The input comprises the input \gls{mri} with either 
the real or fake \gls{ct} concated at the last dimension. The final output is
a score map of shape $1\times24\times512$.

\subsubsection{Context-aware 3D synthesis}

\begin{figure}[h]
  \centering
  \includegraphics[page=6,width=.8\linewidth]{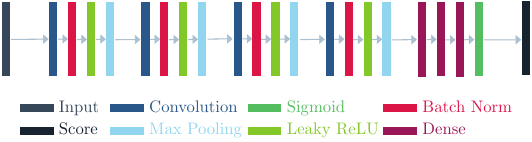}
  \caption{The context-aware 3D synthesis discriminator architecture.
	}\label{fig:synthesis:disc}
\end{figure}

The last model uses 3D patches of shape $32\times32\times32$ from the
\gls{mri} to synthesize \gls{ct} patches of shape $16\times16\times16$. By
using a larger volume for the input the network is able to perform
context-aware synthesis. Furthermore the patch-based data approach allows the
support of different sized brain volumes or even only specific subregions ---
as long as the voxel size correspond to the same world sizes. Even though
patch-based models give benefits under practical circumstances, they increase
the complexity of the pre- and postprocessing by requiring patch extraction
and aggregation.
\begin{figure}[h]
  \centering
  \includegraphics[page=5,width=.8\linewidth]{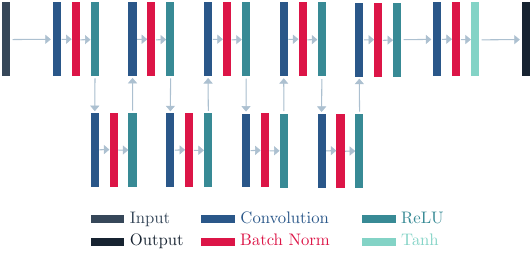}
  \caption{The contxt-aware 3D synthesis generator architecture.
  }\label{fig:synthesis:gen}
\end{figure}
In \Cref{fig:synthesis:gen} and \Cref{fig:synthesis:disc} we illustrated the
generator and discriminator architecture of the context-aware 3D synthesis
model. The generator convolves the input \gls{mri} patch to the target
\gls{ct} patch. In comparison to the u-net based generators there are no
interconnected layers.
The discriminator takes a similar approach and reduces the output or target
\gls{ct} patch to a score map of shape $8\times8\times8\times1$. In comparison
to pixtopix it does not consider the input \gls{mri}. Furthremore we note that
the lack of dropout layers and the preference of ReLUs over Leaky ReLUs as
well as max pooling over transposed convolution (deconvolution).
\begin{table}[h]
  \centering
  \begin{tabular}{cccc}
    \toprule
    Type & Kernel & Strides & Output Shape \\
    \midrule
    Input & & & \num{32x32x32x1} \\ 
    Convolution & \num{9x9x9} & \num{1} & \num{24x24x24x32} \\
    Convolution & \num{3x3x3} & \num{1} & \num{24x24x24x32} \\
    Convolution & \num{3x3x3} & \num{1} & \num{24x24x24x32} \\
    Convolution & \num{3x3x3} & \num{1} & \num{24x24x24x32} \\
    Convolution & \num{9x9x9} & \num{1} & \num{16x16x16x64} \\
    Convolution & \num{3x3x3} & \num{1} & \num{16x16x16x64} \\
    Convolution & \num{3x3x3} & \num{1} & \num{16x16x16x32} \\
    Convolution & \num{7x7x7} & \num{1} & \num{16x16x16x32} \\
    Convolution & \num{3x3x3} & \num{1} & \num{16x16x16x32} \\
    Convolution & \num{3x3x3} & \num{1} & \num{16x16x16x1} \\
    Output & & & \num{16x16x16x1} \\ 
    \bottomrule
  \end{tabular}
  \caption{Network parameters used in the context-aware 3D synthesis generator.
  }\label{tab:synthesis:gen}
\end{table}
\Cref{tab:synthesis:gen} discloses the network parameters of the generator.
Though the kernel size was given in Ref.~\cite{Nie16}, we had to experiment
with the padding algorithm and the stride parameter in order to reproduce the
dimension reduction to \num{16x16x16}.
\begin{table}[h]
  \centering
  \begin{tabular}{cccc}
    \toprule
    Type & Kernel & Strides & Output Shape \\
    \midrule
    Input & & & \num{16x16x16x1} \\ 
    Convolution & \num{5x5x5} & \num{1} & \num{16x16x16x32} \\
    Max Pooling & \num{3x3x3} & \num{1} & \num{14x14x14x32} \\
    Convolution & \num{5x5x5} & \num{1} & \num{14x14x14x64} \\
    Max Pooling & \num{3x3x3} & \num{1} & \num{12x12x12x64} \\
    Convolution & \num{5x5x5} & \num{1} & \num{12x12x12x128} \\
    Max Pooling & \num{3x3x3} & \num{1} & \num{10x10x10x128} \\
    Dense & \num{512} & & \num{8x8x8x512} \\
    Dense & \num{128} & & \num{8x8x8x128} \\
    Dense & \num{1} & & \num{8x8x8x1} \\
    Output & & & \num{8x8x8x1} \\ 
    \bottomrule
  \end{tabular}
  \caption{Network parameters used in the context-aware 3D synthesis
    discriminator.
  }\label{tab:synthesis:disc}
\end{table}
\Cref{tab:synthesis:disc} discloses the network parameters of the
discriminator. The dense layer, also known as fully connected layer, connects
each feature channel of the output of the last max pooling layer with each
other. The final output score map is of shape \num{8x8x8x1}.
\begin{figure}[h]
  \centering
  \includegraphics[page=7,width=.6\linewidth]{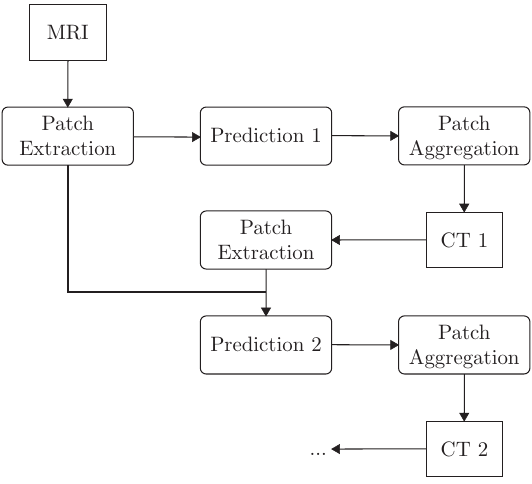}
  \caption{The auto-context model used in ontext-aware 3D synthesis for
    image refinement.
	}\label{fig:synthesis:refine}
\end{figure}
We already noted that the context-aware 3D synthesis generator lacks
interconnected layers in comparison to u-net. Instead, it uses the
auto-context model first introduced in Ref.~\cite{Tu2010}. The concept is
illustrated in \Cref{fig:synthesis:refine}. The idea is to first train a
single model instance on a pair of \gls{ct} and \gls{mri} patches. The
predicted \gls{ct} then are used as input together with the \gls{mri} patches
to train a second model instance. Applied iteratively this approach converges
after three iterations~\cite{Nie16}.

\subsection{Losses}

Beside the preprocessing and the network architecture another important part
in using neural network is the choice of a cost function. The cost function is
necessary in order to calculate a gradient with respect to the network
weights. The network weights are then updated according to their respective
gradient and a convergence parameter of the optimizer. In this manner one
hopes to find the optimal weights for a specific task.

In our experiments we relied on the Adam optimizer, see Ref.~\cite{Kingma2014},
with the parameters listed in \Cref{tab:optimizer}.
\begin{table}[h]
  \centering
  \begin{tabular}{ccc}
    \toprule
    Learning Rate & $\beta_1$ & $\beta_2$ \\
    \midrule
    \num{2e-4} & \num{5e-1} & \num{0.999} \\
    \bottomrule
  \end{tabular}
  \caption{Adam optimizer parameters used for our experiments.
  }\label{tab:optimizer}
\end{table}
These parameters were choosen empirically for fast convergence and good
results. That said there may of course exist better parameters. We did not
perform grid search or other hyperparameter optimization techniques.

\subsubsection{Distance}

Distance based losses are well-known from a wide range of scientific
disciplines and correspond to a distance between two pixel values. We will
present some distance losses now. Let $X,Y\in{[0,1]}^N$ be output and target
vectors, then we define the \gls{mae} to be
\begin{equation}
  \mae\left(X,Y\right)
  =
  \frac{1}{N}\sum_{i=1}^N
  \abs{X_i-Y_i}
  \label{eq:dist:mae}.
\end{equation}
The \gls{mse} we define via
\begin{equation}
  \mse\left(X,Y\right)
  =
  \frac{1}{N}\sum_{i=1}^N
  {\left(X_i-Y_j\right)}^2
  \label{eq:dist:mse}.
\end{equation}
Finally the \gls{gdl} disclosed in Ref.~\cite{Nie16} is defined as
\begin{equation}
  \gdl\left(X,Y\right)
  =
  \mse\left(\grad X,\grad Y\right)
  \label{eq:dist:gdl},
\end{equation}
wherein $\grad$ is the spatial gradient. We approximate the $i$th element
of the spatial gradient through
\begin{equation}
  \grad X_i
  \approx
  \begin{cases}
    X_i-X_{i+1}\qc\text{if }, $1<i<N$\\
    0\qc\text{otherwise}
  \end{cases}
  \label{eq:dist:grad}.
\end{equation}
The loss terms can of course be combined
\begin{equation}
  \lambda_\text{MAE}\mae\left(X,Y\right)
  +
  \lambda_\text{MSE}\mse\left(X,Y\right)
  +
  \lambda_\text{GDL}\gdl\left(X,Y\right),
\end{equation}
wherein the $\lambda$ denotes the weight of the respective loss term.

The \gls{mse} penalizes outliers stronger than the \gls{mae}. Furthermore the
\gls{mse} offers a continous derivative whereas the \gls{mae} has undefined
behaviour at $0$. The \gls{gdl} was reported to correct for strong
edges~\cite{Nie16}, as present at the tissue boundaries in the brain.

\subsubsection{Adversarial}

A major shortcoming of the distance based losses is that they only consider
pixel-wise deviations and thereby neglect more complex structures. With the
advent of \gls{gan} one can think of the adversarial loss as an embodiment of
more complex function that respects (local) structures. Given a discriminator
network that outputs a score map for real $D(X)$ and fake data $D(Y)=D(G(Z))$,
where $G(Z)$ is the generator output from the input vector $Z$, the
standard adversarial loss is defined as
\begin{equation}
  \log(D(X))+\log(1-D(G(Z)))
  \label{eq:adv:minmax}.
\end{equation}
Recently a modified least-squared adversarial loss
\begin{equation}
  \log{\left(D{(X)}^2\right)}+\log{\left({(1-D{\left(G{(Z)}\right)})}^2\right)}
  \label{eq:adv:lsquare},
\end{equation}
has been reported that yields superior results and more stable training
characteristics~\cite{Mao2016}. The least-squared adversarial loss is used
in the pixtopix model.

\section{Experiments}

As noted earlier, we ran into practical challenges with our implementation of
the patch aggregation algorithm required for the implementation of the
context-aware 3D synthesis. Though patch aggregation worked in general, it
occupied more computational resources than we could consume without the
interference with other projects. As a result we did not perform more than one
iteration of the auto-context model, which prevents us from a fair comparison,
however, we encourage everyone to test our implementation themselves.

Consequently we will only report results obtained with the u-net \gls{cnn} and
the pixtopix \gls{gan} model.

\subsection{u-net}

The \num{17} subjects of the dataset were divided into \num{13} subjects for
training and \num{4} subjects for validation. We tried to respect the
transverse resolution of the initial volumes, i.e.\ the number of transverse
slices, in the division process. In \Cref{tab:unet:dataset} we see the volume
shape of the input \gls{mri} and the target \gls{ct} of the respective
subject as well as their assignment to the training or validation dataset.
\begin{table}[h]
  \centering
  \begin{tabular}{ccc}
    \toprule
    Dataset & Subject & Shape \\
    \midrule
    Training & \num{1} & \num{161x320x250x1} \\
    Training & \num{2} & \num{149x328x250x1} \\
    Training & \num{3} & \num{112x303x281x1} \\
    Training & \num{4} & \num{155x291x259x1} \\
    Training & \num{5} & \num{143x307x284x1} \\
    Training & \num{6} & \num{149x278x267x1} \\
    Training & \num{7} & \num{200x289x268x1} \\
    Training & \num{8} & \num{218x282x238x1} \\
    Training & \num{9} & \num{191x322x252x1} \\
    Training & \num{10} & \num{200x303x243x1} \\
    Training & \num{11} & \num{181x317x239x1} \\
    Training & \num{12} & \num{186x310x248x1} \\
    Training & \num{13} & \num{112x313x238x1} \\
    Validation & \num{1} & \num{112x298x227x1} \\
    Validation & \num{2} & \num{223x328x282x1} \\
    Validation & \num{3} & \num{223x307x276x1} \\
    Validation & \num{4} & \num{204x329x262x1} \\
    \bottomrule
  \end{tabular}
  \caption{Training and validation dataset volumes used in this section. The
    dimensions of the shape correspond to depth, height and width.
  }\label{tab:unet:dataset}
\end{table}
The subjects assigned to the training dataset were processed in transverse
slices. We trained the u-net model once with the \gls{mae} and once with the
\gls{mae} and \gls{gdl} loss in order to estimate the impact of the \gls{gdl}.
The training parameters are summarized in \Cref{tab:unet:params}. The image
slices correspond to the total number of 2D images extracted from the
transverse (depth) plane of the volumes. The batch size denotes the number of
images proccessed in one step. For convenience we estimated the number of
epochs from the total training step number. The training was stopped when the
gradients vanished and the total loss stabilized. We found that these criteria
were met for the u-net model at around \num{20000} steps or \num{140} epochs.
\begin{table}[h]
  \centering
  \begin{tabular}{cccccc}
    \toprule
    & &
    \multicolumn{2}{c}{\acrshort{mae}} &
    \multicolumn{2}{c}{\acrshort{mae}+\acrshort{gdl}} \\
    Image Slices & Batch Size & Steps & Epochs & Steps & Epochs \\
    \midrule
    \num{2157} & \num{16} & \num{20542} & \num{152} & \num{19388} & \num{143} \\
    \bottomrule
  \end{tabular}
  \caption{Training parameters used for the distance metrics experiments.
  }\label{tab:unet:params}
\end{table}
The appropriate loss term weights $\lambda_\text{mae}$ and
$\lambda_\text{mse}$ were chosen such that the gradient with respect to the
loss terms yields a similar magnitude. We found that to be the case for
\num{1e-7}. In an early attempt we also tried to compare \gls{mae} and
\gls{mse} as loss functions, yet, we did not find significant differences and
stuck with \gls{mae} which is the distance loss used in the original pixtopix. 
\begin{table}[h]
  \centering
  \begin{tabular}{ccccc}
    \toprule
    $\lambda_\text{mae}$ &
    $\lambda_\text{gdl}$ &
    \acrshort{mae} &
    \acrshort{mse} &
    \acrshort{psnr} \\
    \midrule
    \num{1} & \num{0} & \num{31.58} & \num{6577} & \num{59.5} \\
    \num{1} & \num{1e-7} & \num{37.15} & \num{7945} & \num{58.1} \\
    \bottomrule
  \end{tabular}
  \caption{Distance metrics for the u-net model trained with different loss
    functions, evaluated on the training dataset.
  }\label{tab:unet:training}
\end{table}
\Cref{tab:unet:training} lists the metrics for the u-net model trained with
different loss functions evaluated on the training dataset. The \gls{psnr}
metric is defined as
\begin{equation}
	\psnr{\left(X,Y\right)}
	=
	10\log{\left(\frac{M^2}{\mse}\right)}
	\label{eq:psnr},
\end{equation}
wherein $M$ denotes the maximum pixel value, in our case $2^{16}-1$. It is
useful to quantify the noise level present in the generated \gls{ct} images,
with a higher \gls{psnr} usually corresponding to lower noise. We remark
that the \gls{gdl} yields a slightly better result on the \gls{psnr} metric,
but yields inferior values on \gls{mae} and \gls{mse}.
\begin{table}[h]
  \centering
  \begin{tabular}{ccccc}
    \toprule
    $\lambda_\text{mae}$ &
    $\lambda_\text{gdl}$ &
    \acrshort{mae} &
    \acrshort{mse} &
    \acrshort{psnr} \\
    \midrule
    \num{1} & \num{0} & \num{123.6} & \num{70846} & \num{47.93} \\
    \num{1} & \num{1e-7} & \num{129.0} & \num{72704} & \num{47.90} \\
    \bottomrule
  \end{tabular}
  \caption{Distance metrics for the u-net model trained with different loss
    functions, evaluated on the validation dataset.
  }\label{tab:unet:validation}
\end{table}
In \Cref{tab:unet:validation} we see the same metrics evaluated on the
validation dataset. These metrics are in general more informative than the
metrics from the training dataset as we expect the networks to overfit. In
comparison to the \Cref{tab:unet:training} the \gls{mae} are nearly four times
the \gls{mae} for the training dataset. The \gls{mse} is of one magntiude
higher which also confirms overfitting. The \gls{psnr} metric obtained from
the validation dataset is lower than for the training dataset. We should keep
in mind that for the \gls{psnr} a higher value is usually better and also that
the \gls{psnr} is scaled logarithmicly. Overall the metrics suggest that our
network overfits and that the \gls{gdl} slightly decreases the performance.
Nevertheless we should keep in mind that these metrics are based on
pixel-wise measures, therefore we need to examine the visual results to draw
final conclusions and sort out, for instance, possible bias in the subject
selection of the datasets.
\begin{figure}[h]
  \centering
  \begin{adjustbox}{width=\linewidth}
\begingroup%
\makeatletter%
\begin{pgfpicture}%
\pgfpathrectangle{\pgfpointorigin}{\pgfqpoint{12.000000in}{16.000000in}}%
\pgfusepath{use as bounding box, clip}%
\begin{pgfscope}%
\pgfsetbuttcap%
\pgfsetmiterjoin%
\pgfsetlinewidth{0.000000pt}%
\definecolor{currentstroke}{rgb}{1.000000,1.000000,1.000000}%
\pgfsetstrokecolor{currentstroke}%
\pgfsetdash{}{0pt}%
\pgfpathmoveto{\pgfqpoint{0.000000in}{0.000000in}}%
\pgfpathlineto{\pgfqpoint{12.000000in}{0.000000in}}%
\pgfpathlineto{\pgfqpoint{12.000000in}{16.000000in}}%
\pgfpathlineto{\pgfqpoint{0.000000in}{16.000000in}}%
\pgfpathclose%
\pgfusepath{}%
\end{pgfscope}%
\begin{pgfscope}%
\pgfpathrectangle{\pgfqpoint{0.622560in}{11.374975in}}{\pgfqpoint{2.863300in}{3.665025in}}%
\pgfusepath{clip}%
\pgfsys@transformshift{0.623333in}{11.373333in}%
\pgftext[left,bottom]{\pgfimage[interpolate=true,width=2.860000in,height=3.666667in]{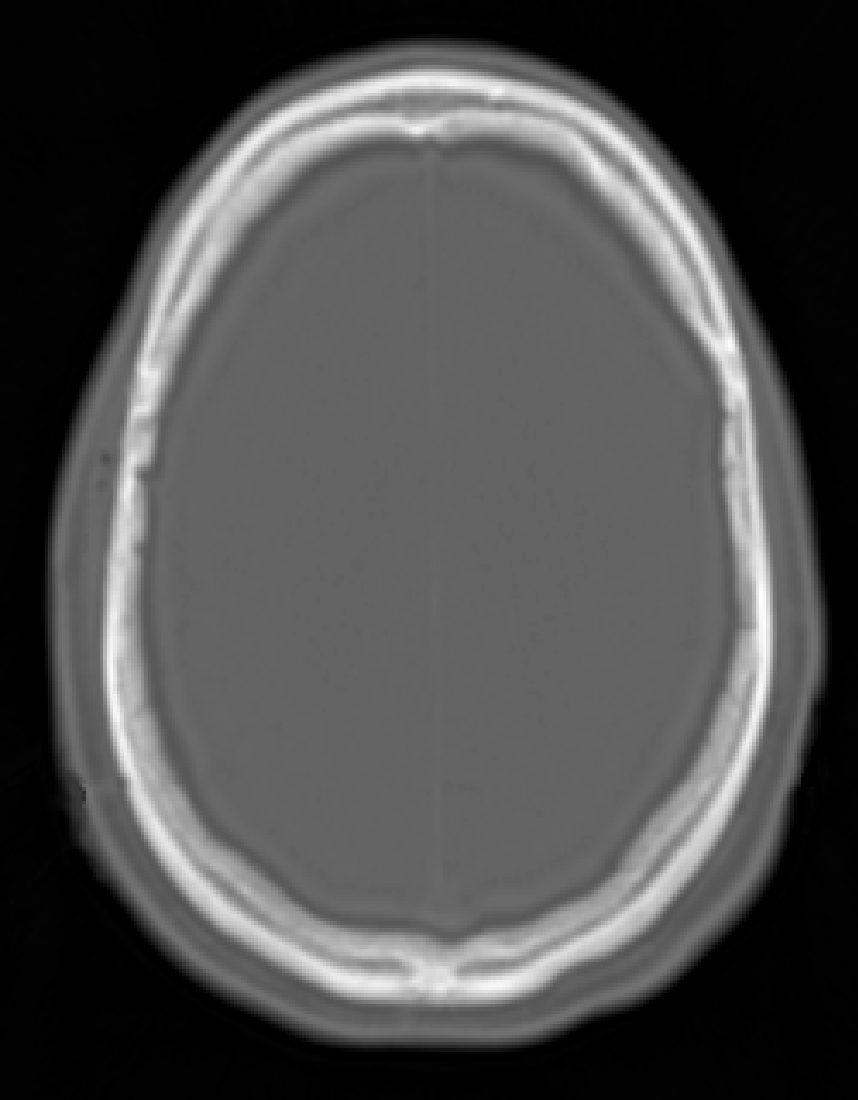}}%
\end{pgfscope}%
\begin{pgfscope}%
\pgftext[x=2.054211in,y=15.317778in,,base]{\fontsize{16.000000}{19.200000}\selectfont Ground Truth}%
\end{pgfscope}%
\begin{pgfscope}%
\pgfpathrectangle{\pgfqpoint{4.568350in}{11.374975in}}{\pgfqpoint{2.863300in}{3.665025in}}%
\pgfusepath{clip}%
\pgfsys@transformshift{4.570000in}{11.373333in}%
\pgftext[left,bottom]{\pgfimage[interpolate=true,width=2.863333in,height=3.666667in]{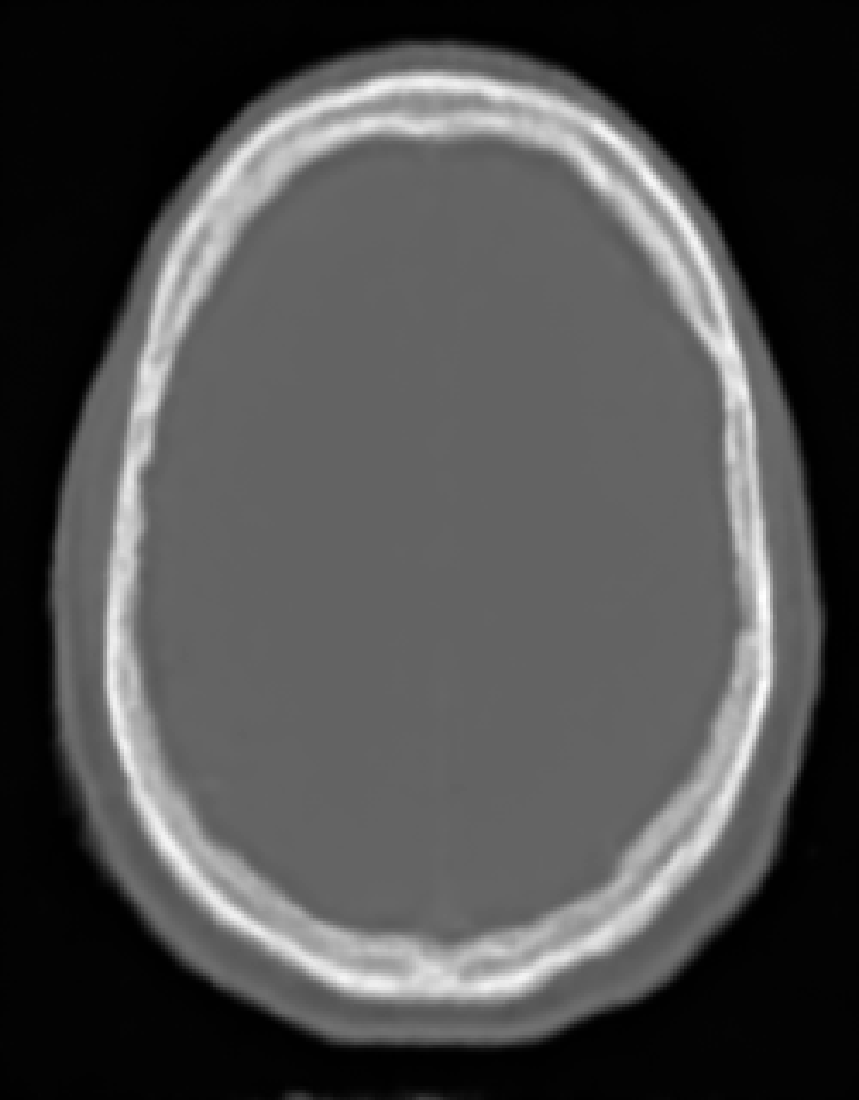}}%
\end{pgfscope}%
\begin{pgfscope}%
\pgftext[x=6.000000in,y=15.317778in,,base]{\fontsize{16.000000}{19.200000}\selectfont \(\displaystyle \lambda_\text{mae}=1,\lambda_\text{gdl}=0\)}%
\end{pgfscope}%
\begin{pgfscope}%
\pgfpathrectangle{\pgfqpoint{8.514139in}{11.374975in}}{\pgfqpoint{2.863300in}{3.665025in}}%
\pgfusepath{clip}%
\pgfsys@transformshift{8.513333in}{11.373333in}%
\pgftext[left,bottom]{\pgfimage[interpolate=true,width=2.863333in,height=3.666667in]{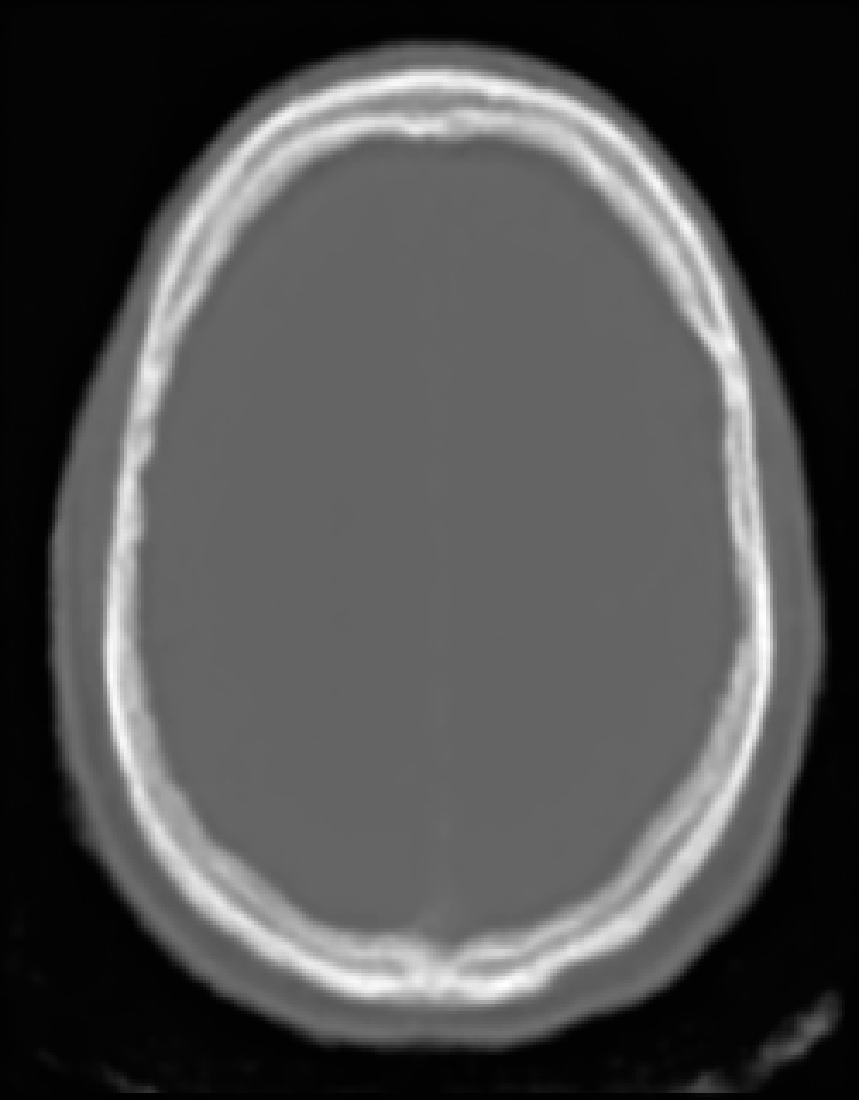}}%
\end{pgfscope}%
\begin{pgfscope}%
\pgftext[x=9.945789in,y=15.317778in,,base]{\fontsize{16.000000}{19.200000}\selectfont \(\displaystyle \lambda_\text{mae}=1,\lambda_\text{gdl}=\num{1e-7}\)}%
\end{pgfscope}%
\begin{pgfscope}%
\pgfpathrectangle{\pgfqpoint{0.657479in}{7.636650in}}{\pgfqpoint{2.793464in}{3.665025in}}%
\pgfusepath{clip}%
\pgfsys@transformshift{0.656667in}{7.636667in}%
\pgftext[left,bottom]{\pgfimage[interpolate=true,width=2.796667in,height=3.666667in]{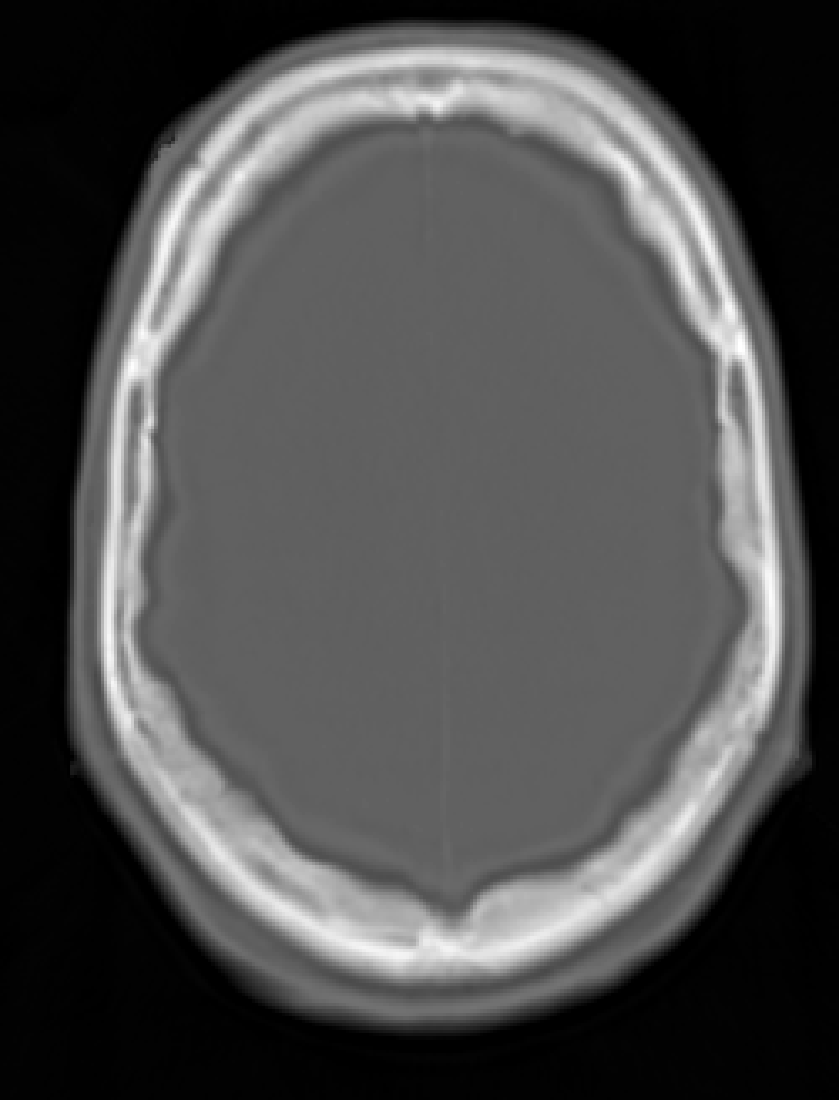}}%
\end{pgfscope}%
\begin{pgfscope}%
\pgfpathrectangle{\pgfqpoint{4.603268in}{7.636650in}}{\pgfqpoint{2.793464in}{3.665025in}}%
\pgfusepath{clip}%
\pgfsys@transformshift{4.603333in}{7.636667in}%
\pgftext[left,bottom]{\pgfimage[interpolate=true,width=2.796667in,height=3.666667in]{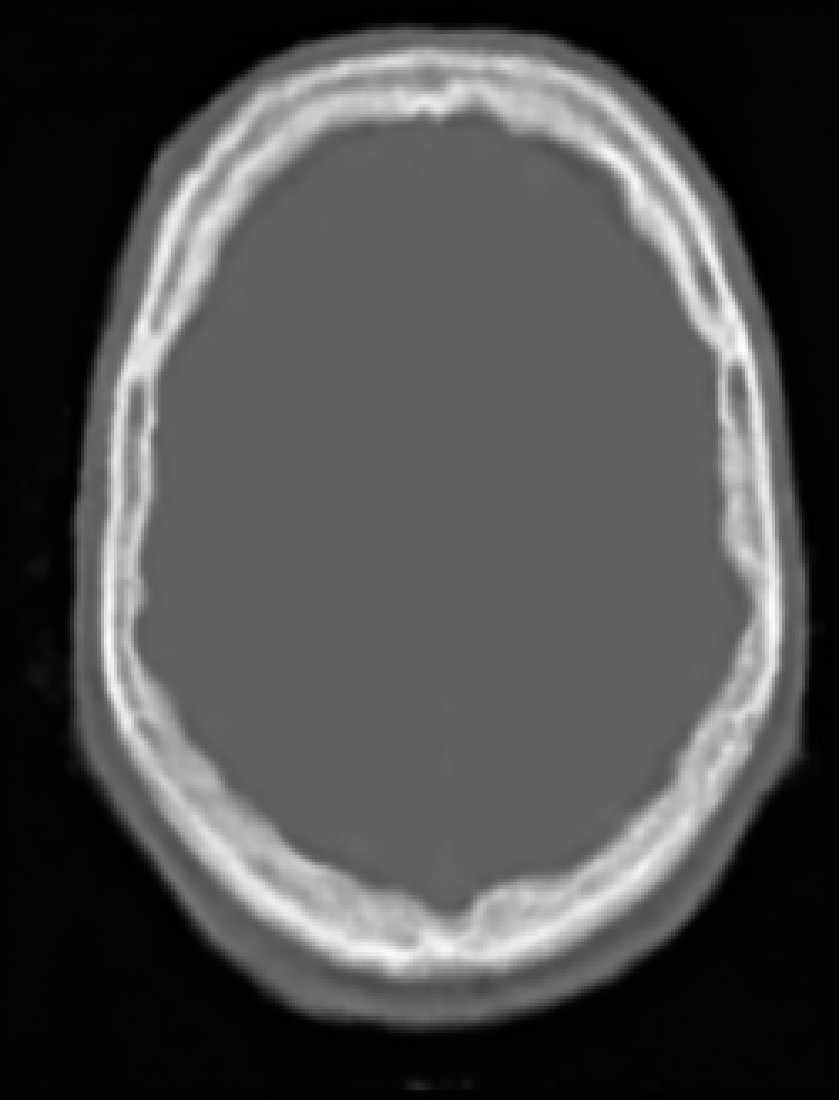}}%
\end{pgfscope}%
\begin{pgfscope}%
\pgfpathrectangle{\pgfqpoint{8.549058in}{7.636650in}}{\pgfqpoint{2.793464in}{3.665025in}}%
\pgfusepath{clip}%
\pgfsys@transformshift{8.550000in}{7.636667in}%
\pgftext[left,bottom]{\pgfimage[interpolate=true,width=2.796667in,height=3.666667in]{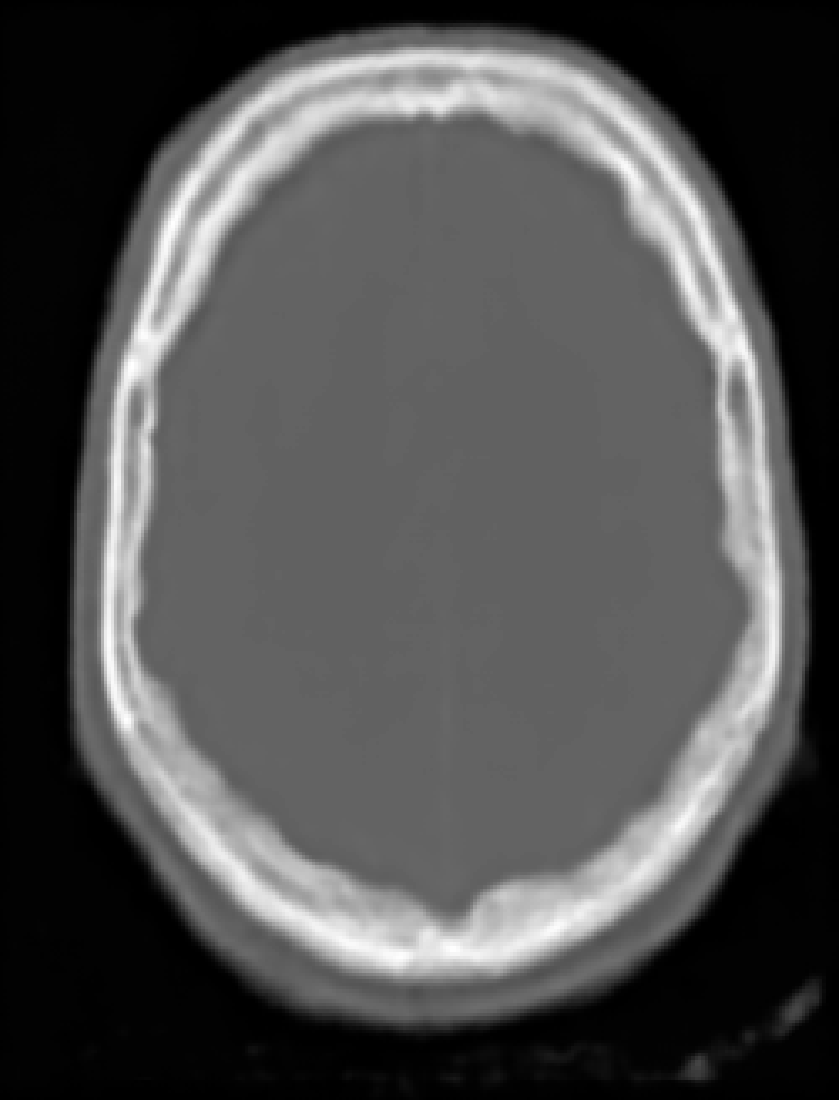}}%
\end{pgfscope}%
\begin{pgfscope}%
\pgfpathrectangle{\pgfqpoint{0.354752in}{3.898325in}}{\pgfqpoint{3.398917in}{3.665025in}}%
\pgfusepath{clip}%
\pgfsys@transformshift{0.353333in}{3.896667in}%
\pgftext[left,bottom]{\pgfimage[interpolate=true,width=3.400000in,height=3.663333in]{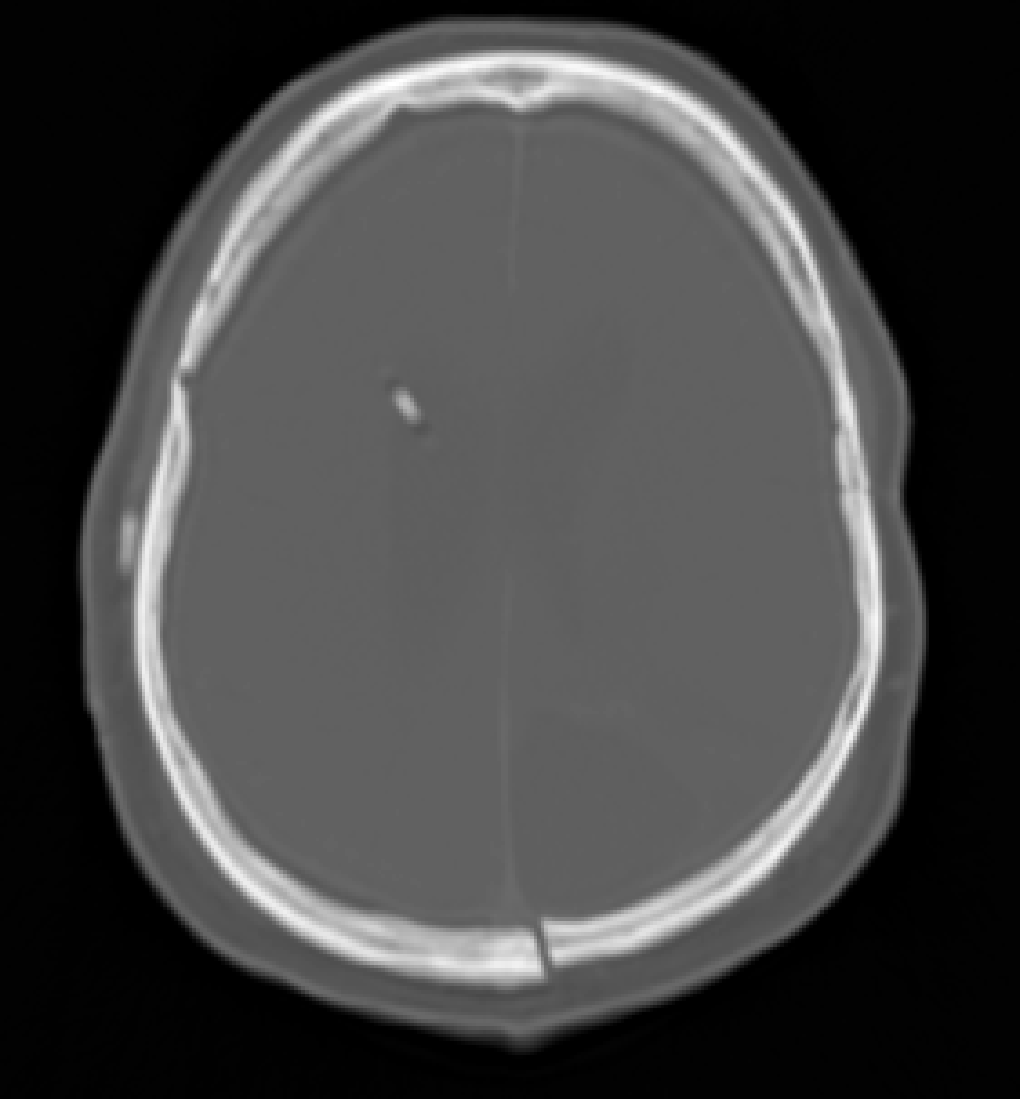}}%
\end{pgfscope}%
\begin{pgfscope}%
\pgfpathrectangle{\pgfqpoint{4.300541in}{3.898325in}}{\pgfqpoint{3.398917in}{3.665025in}}%
\pgfusepath{clip}%
\pgfsys@transformshift{4.300000in}{3.896667in}%
\pgftext[left,bottom]{\pgfimage[interpolate=true,width=3.400000in,height=3.663333in]{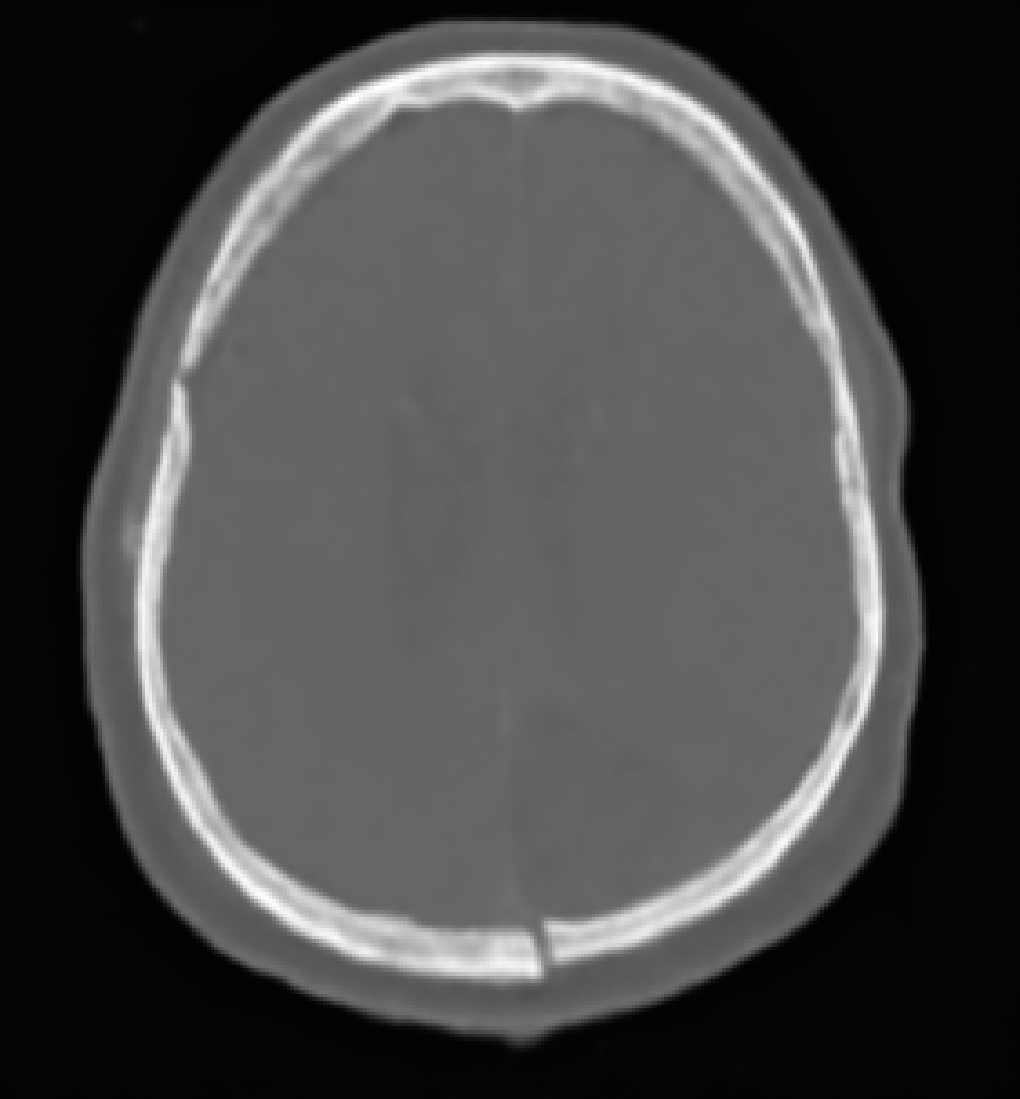}}%
\end{pgfscope}%
\begin{pgfscope}%
\pgfpathrectangle{\pgfqpoint{8.246331in}{3.898325in}}{\pgfqpoint{3.398917in}{3.665025in}}%
\pgfusepath{clip}%
\pgfsys@transformshift{8.246667in}{3.896667in}%
\pgftext[left,bottom]{\pgfimage[interpolate=true,width=3.400000in,height=3.663333in]{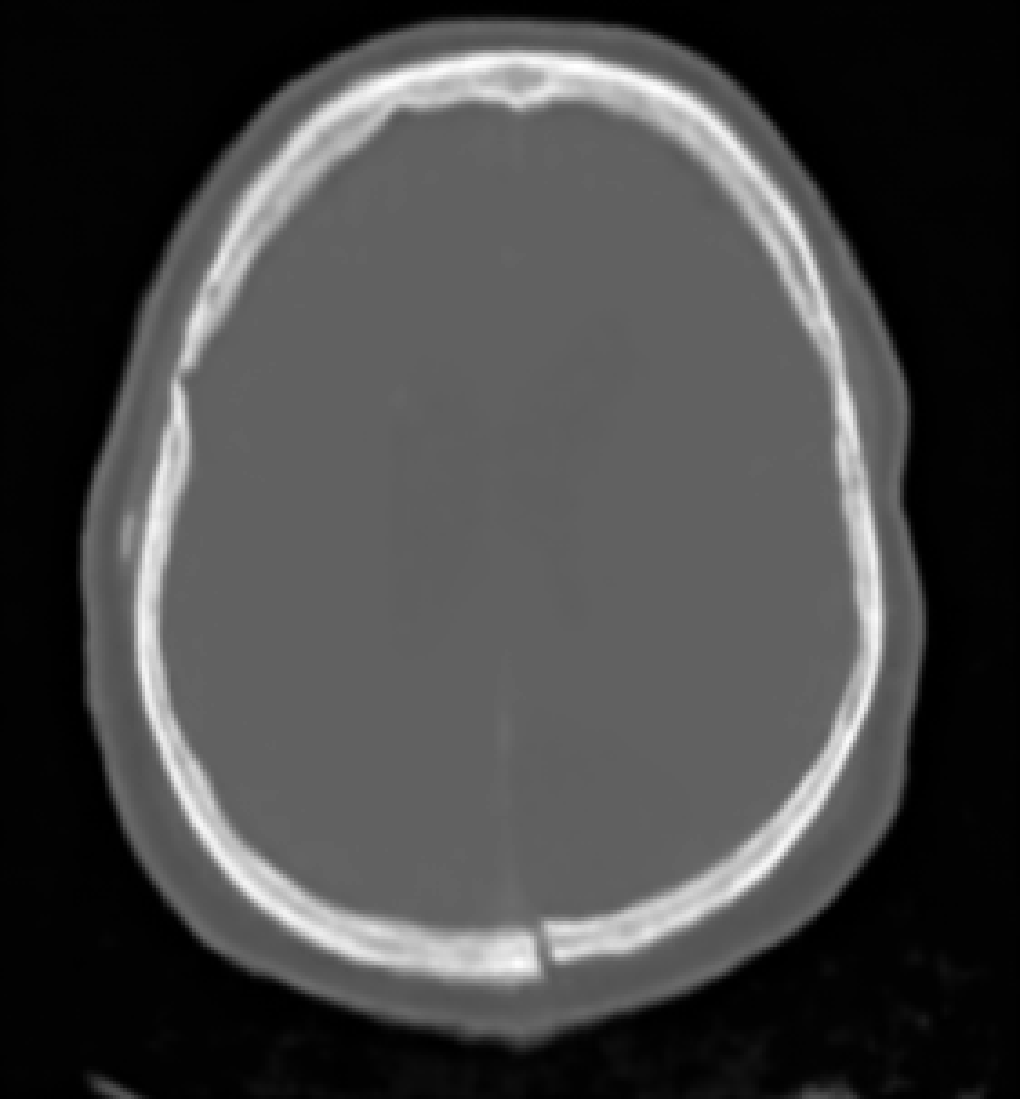}}%
\end{pgfscope}%
\begin{pgfscope}%
\pgfpathrectangle{\pgfqpoint{0.423212in}{0.160000in}}{\pgfqpoint{3.261998in}{3.665025in}}%
\pgfusepath{clip}%
\pgfsys@transformshift{0.423333in}{0.160000in}%
\pgftext[left,bottom]{\pgfimage[interpolate=true,width=3.263333in,height=3.666667in]{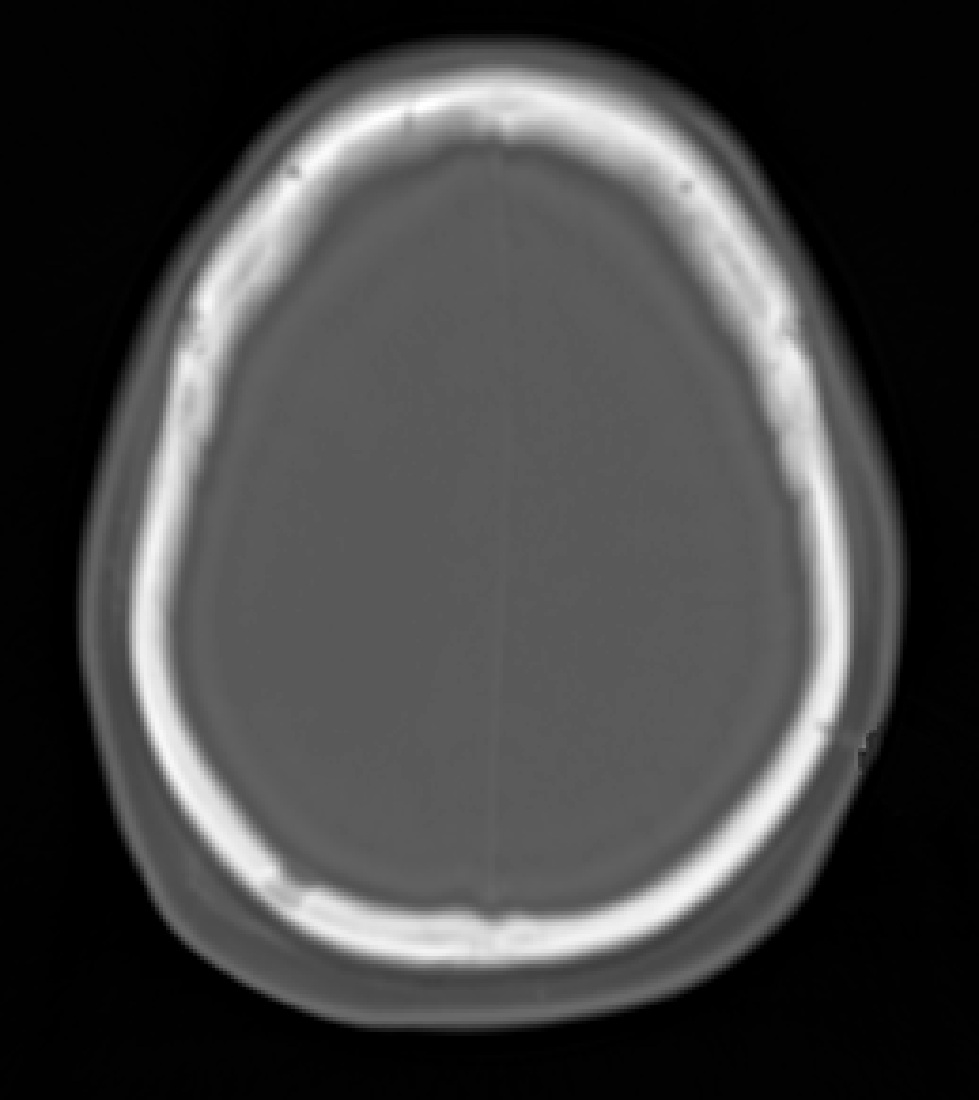}}%
\end{pgfscope}%
\begin{pgfscope}%
\pgfpathrectangle{\pgfqpoint{4.369001in}{0.160000in}}{\pgfqpoint{3.261998in}{3.665025in}}%
\pgfusepath{clip}%
\pgfsys@transformshift{4.370000in}{0.160000in}%
\pgftext[left,bottom]{\pgfimage[interpolate=true,width=3.260000in,height=3.666667in]{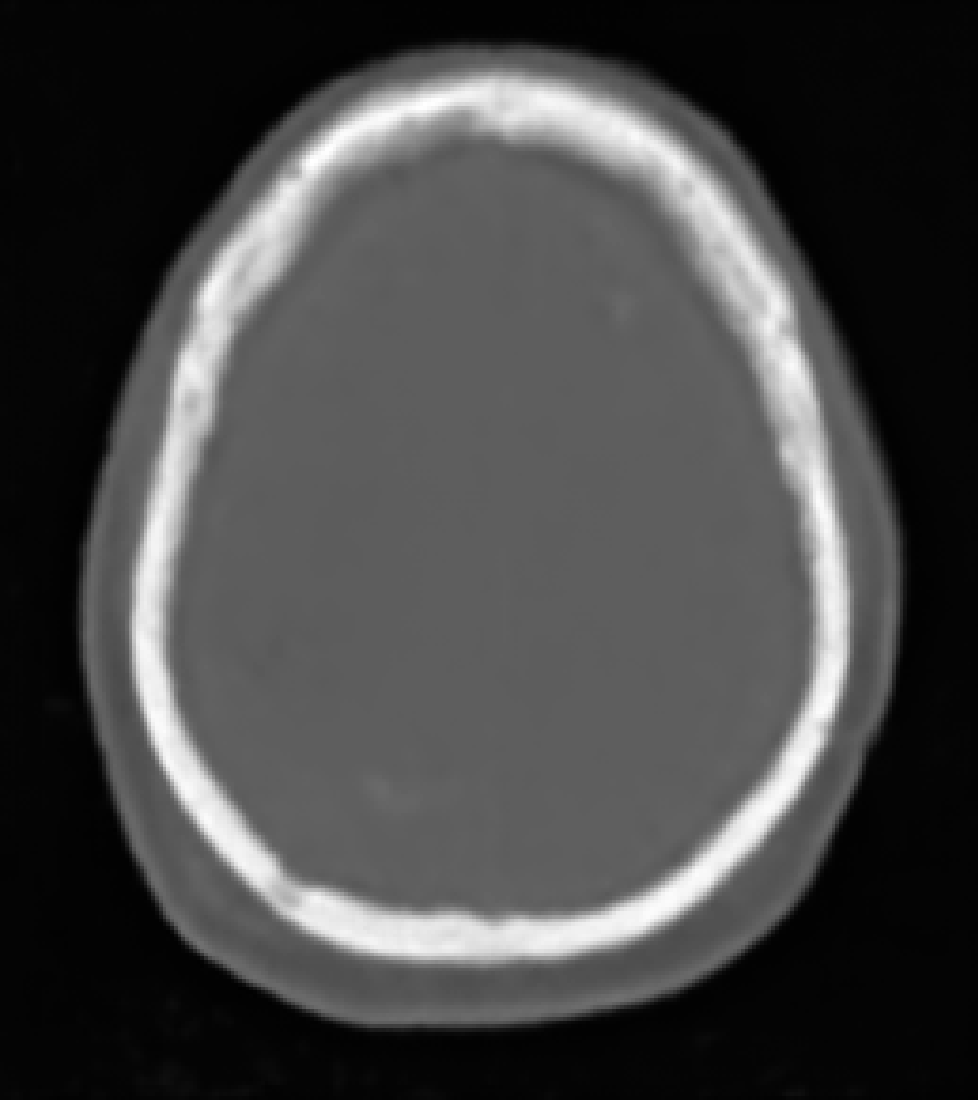}}%
\end{pgfscope}%
\begin{pgfscope}%
\pgfpathrectangle{\pgfqpoint{8.314791in}{0.160000in}}{\pgfqpoint{3.261998in}{3.665025in}}%
\pgfusepath{clip}%
\pgfsys@transformshift{8.313333in}{0.160000in}%
\pgftext[left,bottom]{\pgfimage[interpolate=true,width=3.263333in,height=3.666667in]{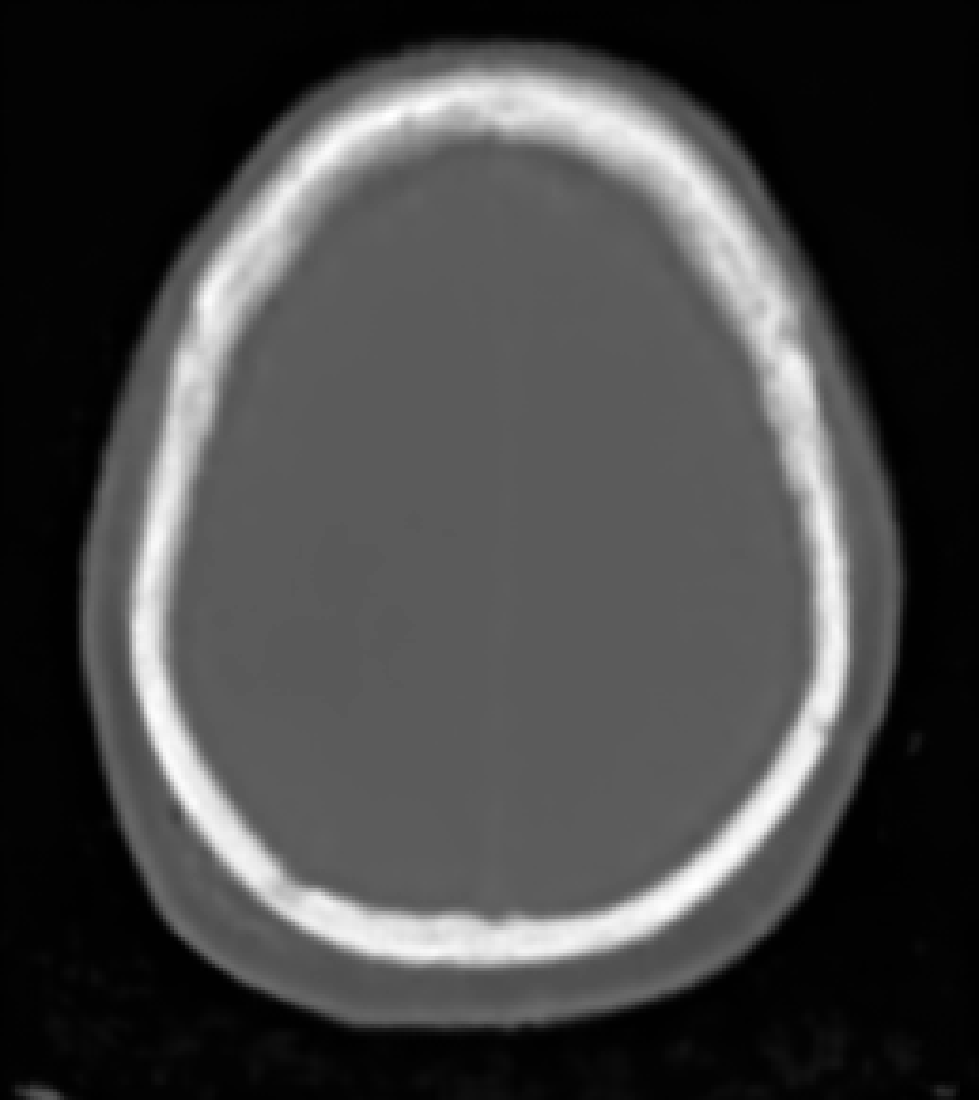}}%
\end{pgfscope}%
\end{pgfpicture}%
\makeatother%
\endgroup%
  \end{adjustbox}
  \caption{Transverse views for the u-net model trained with different
    loss functions, evaluated on the training dataset.
  }\label{fig:unet:training}
\end{figure}
In \Cref{fig:unet:training} we present the the transverse views for the
differently trained u-net models evaluated on the training dataset including
the ground truth on the left. We note that the u-net instance trained with
\gls{gdl} shows some artifacts outside of the head. Furthermore the soft
matter structure seems more coarse.
\begin{figure}[h]
  \centering
  \begin{adjustbox}{width=\linewidth}
\begingroup%
\makeatletter%
\begin{pgfpicture}%
\pgfpathrectangle{\pgfpointorigin}{\pgfqpoint{12.000000in}{16.000000in}}%
\pgfusepath{use as bounding box, clip}%
\begin{pgfscope}%
\pgfsetbuttcap%
\pgfsetmiterjoin%
\pgfsetlinewidth{0.000000pt}%
\definecolor{currentstroke}{rgb}{1.000000,1.000000,1.000000}%
\pgfsetstrokecolor{currentstroke}%
\pgfsetdash{}{0pt}%
\pgfpathmoveto{\pgfqpoint{0.000000in}{0.000000in}}%
\pgfpathlineto{\pgfqpoint{12.000000in}{0.000000in}}%
\pgfpathlineto{\pgfqpoint{12.000000in}{16.000000in}}%
\pgfpathlineto{\pgfqpoint{0.000000in}{16.000000in}}%
\pgfpathclose%
\pgfusepath{}%
\end{pgfscope}%
\begin{pgfscope}%
\pgfpathrectangle{\pgfqpoint{0.658303in}{11.374975in}}{\pgfqpoint{2.791814in}{3.665025in}}%
\pgfusepath{clip}%
\pgfsys@transformshift{0.656667in}{11.373333in}%
\pgftext[left,bottom]{\pgfimage[interpolate=true,width=2.793333in,height=3.666667in]{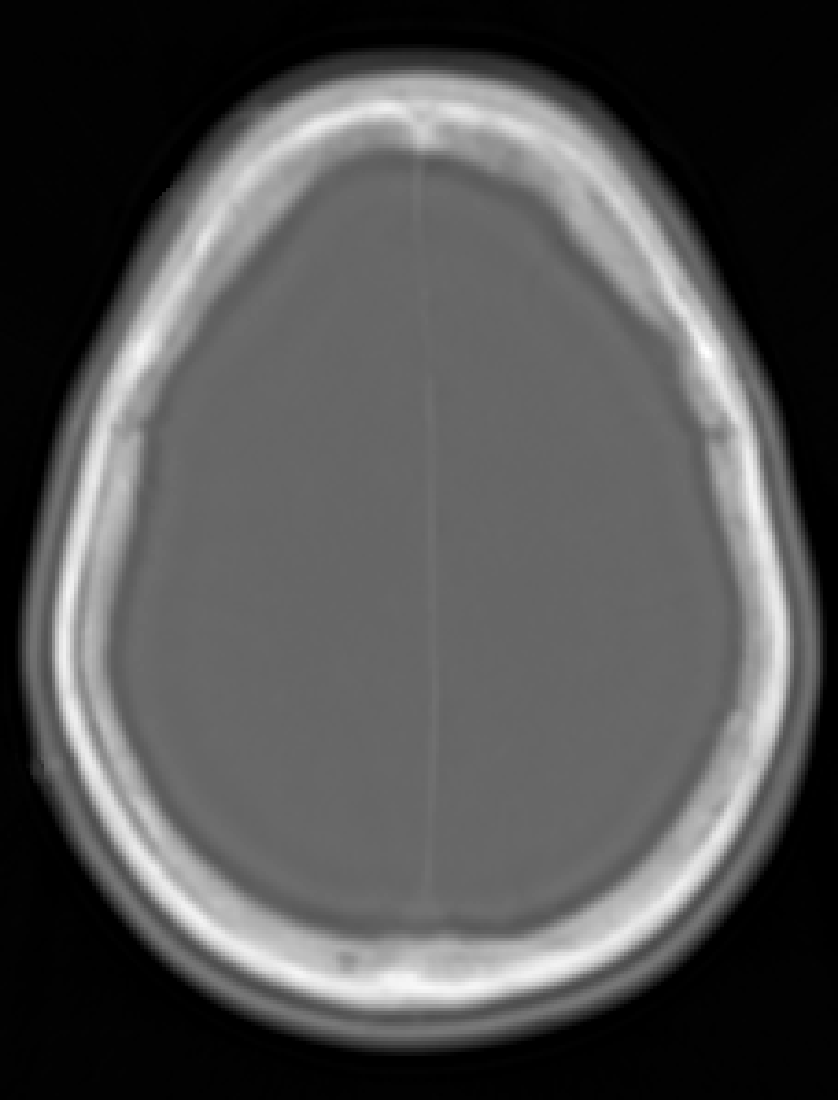}}%
\end{pgfscope}%
\begin{pgfscope}%
\pgftext[x=2.054211in,y=15.317778in,,base]{\fontsize{16.000000}{19.200000}\selectfont Ground Truth}%
\end{pgfscope}%
\begin{pgfscope}%
\pgfpathrectangle{\pgfqpoint{4.604093in}{11.374975in}}{\pgfqpoint{2.791814in}{3.665025in}}%
\pgfusepath{clip}%
\pgfsys@transformshift{4.603333in}{11.373333in}%
\pgftext[left,bottom]{\pgfimage[interpolate=true,width=2.793333in,height=3.666667in]{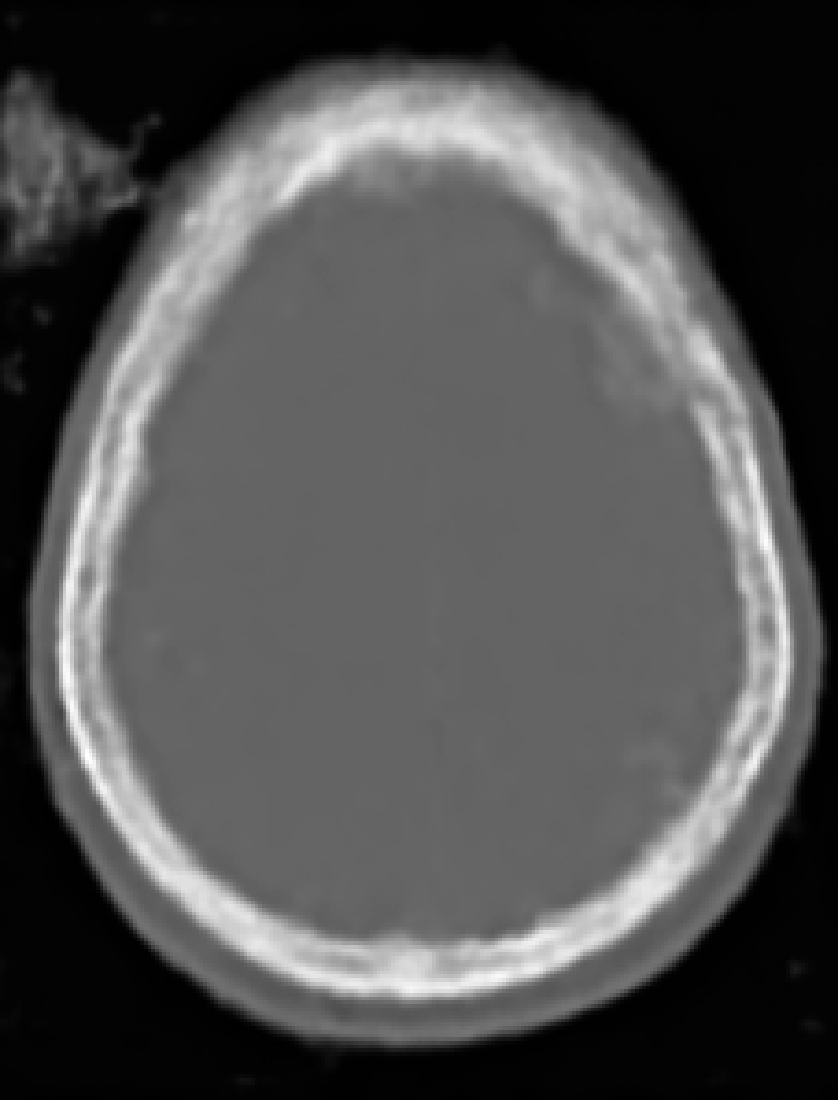}}%
\end{pgfscope}%
\begin{pgfscope}%
\pgftext[x=6.000000in,y=15.317778in,,base]{\fontsize{16.000000}{19.200000}\selectfont \(\displaystyle \lambda_\text{mae}=1,\lambda_\text{gdl}=0\)}%
\end{pgfscope}%
\begin{pgfscope}%
\pgfpathrectangle{\pgfqpoint{8.549882in}{11.374975in}}{\pgfqpoint{2.791814in}{3.665025in}}%
\pgfusepath{clip}%
\pgfsys@transformshift{8.550000in}{11.373333in}%
\pgftext[left,bottom]{\pgfimage[interpolate=true,width=2.793333in,height=3.666667in]{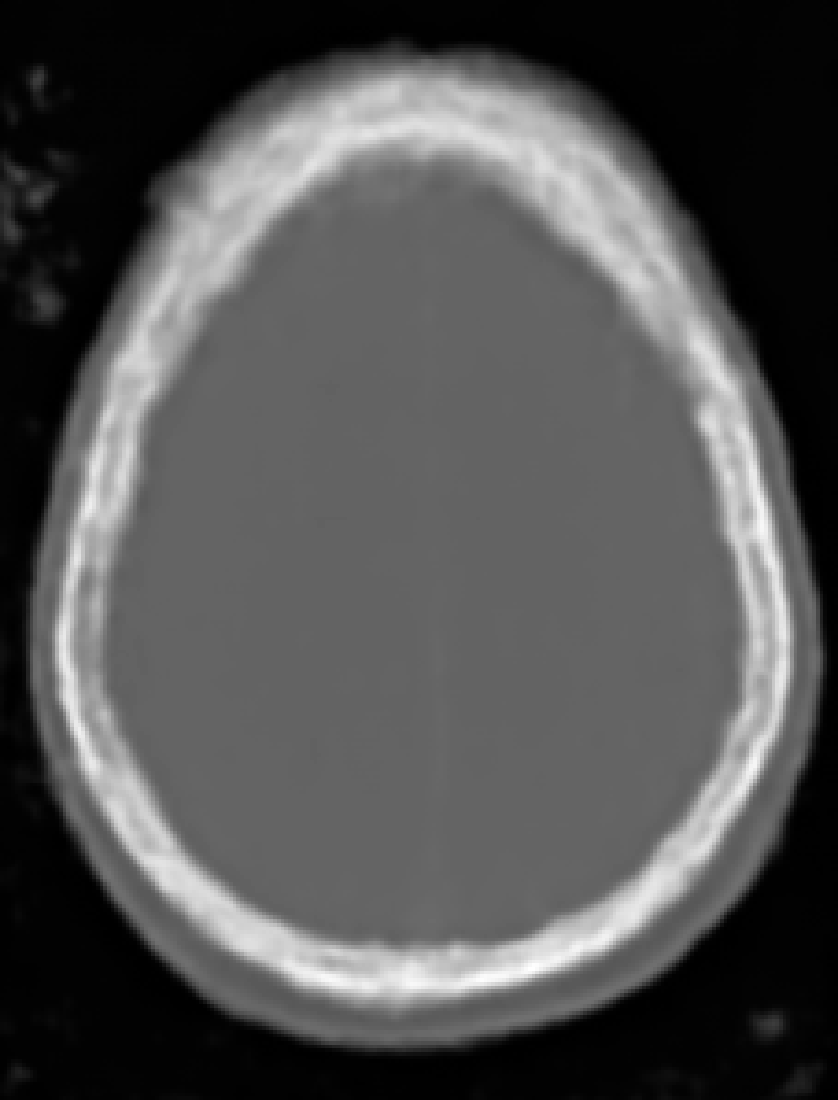}}%
\end{pgfscope}%
\begin{pgfscope}%
\pgftext[x=9.945789in,y=15.317778in,,base]{\fontsize{16.000000}{19.200000}\selectfont \(\displaystyle \lambda_\text{mae}=1,\lambda_\text{gdl}=\num{1e-7}\)}%
\end{pgfscope}%
\begin{pgfscope}%
\pgfpathrectangle{\pgfqpoint{0.478697in}{7.636650in}}{\pgfqpoint{3.151027in}{3.665025in}}%
\pgfusepath{clip}%
\pgfsys@transformshift{0.480000in}{7.636667in}%
\pgftext[left,bottom]{\pgfimage[interpolate=true,width=3.153333in,height=3.666667in]{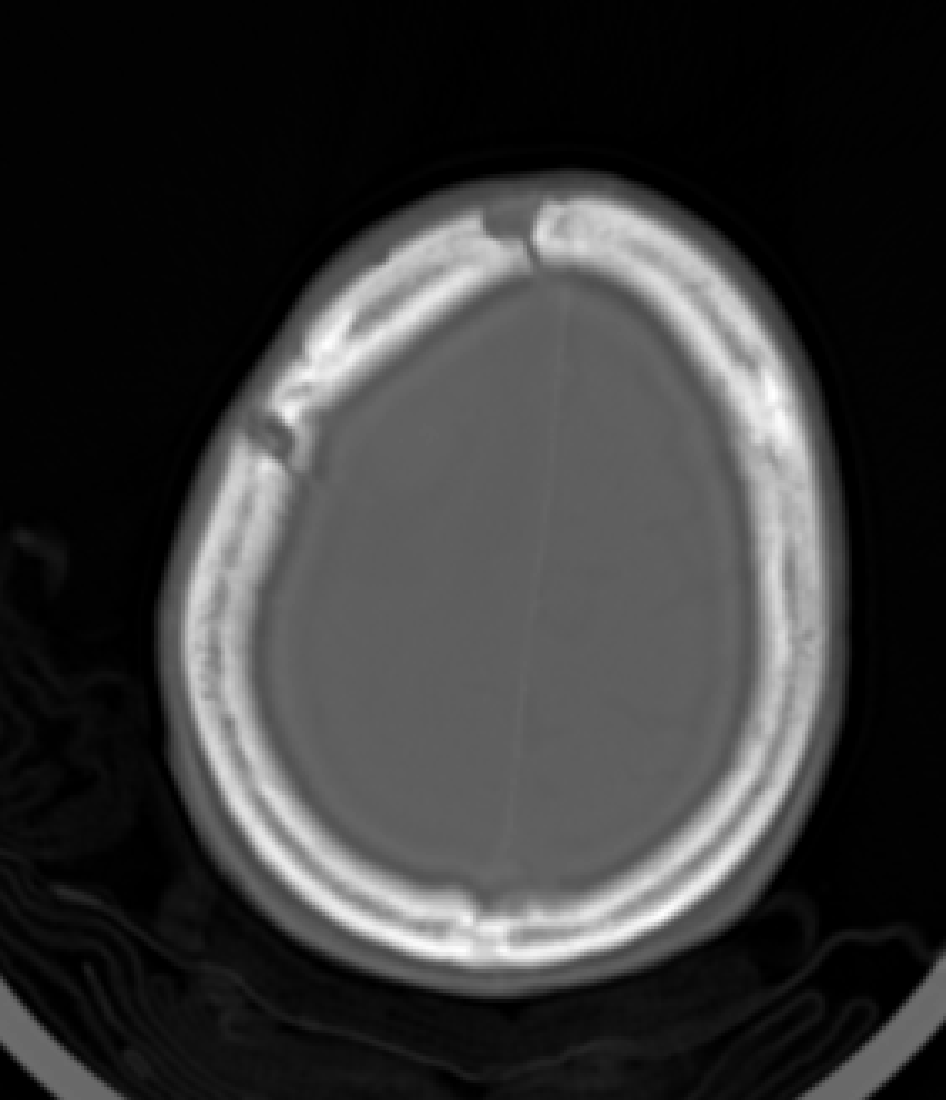}}%
\end{pgfscope}%
\begin{pgfscope}%
\pgfpathrectangle{\pgfqpoint{4.424486in}{7.636650in}}{\pgfqpoint{3.151027in}{3.665025in}}%
\pgfusepath{clip}%
\pgfsys@transformshift{4.423333in}{7.636667in}%
\pgftext[left,bottom]{\pgfimage[interpolate=true,width=3.153333in,height=3.666667in]{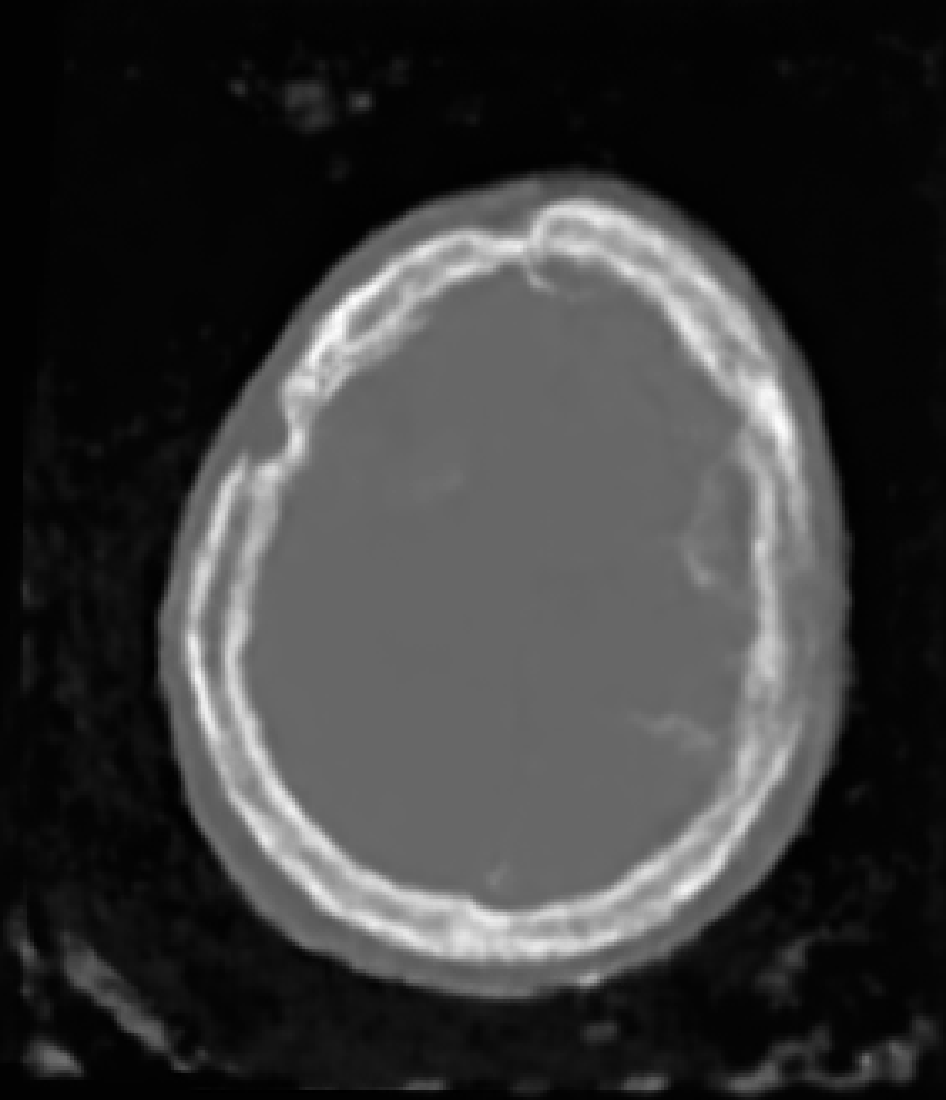}}%
\end{pgfscope}%
\begin{pgfscope}%
\pgfpathrectangle{\pgfqpoint{8.370276in}{7.636650in}}{\pgfqpoint{3.151027in}{3.665025in}}%
\pgfusepath{clip}%
\pgfsys@transformshift{8.370000in}{7.636667in}%
\pgftext[left,bottom]{\pgfimage[interpolate=true,width=3.153333in,height=3.666667in]{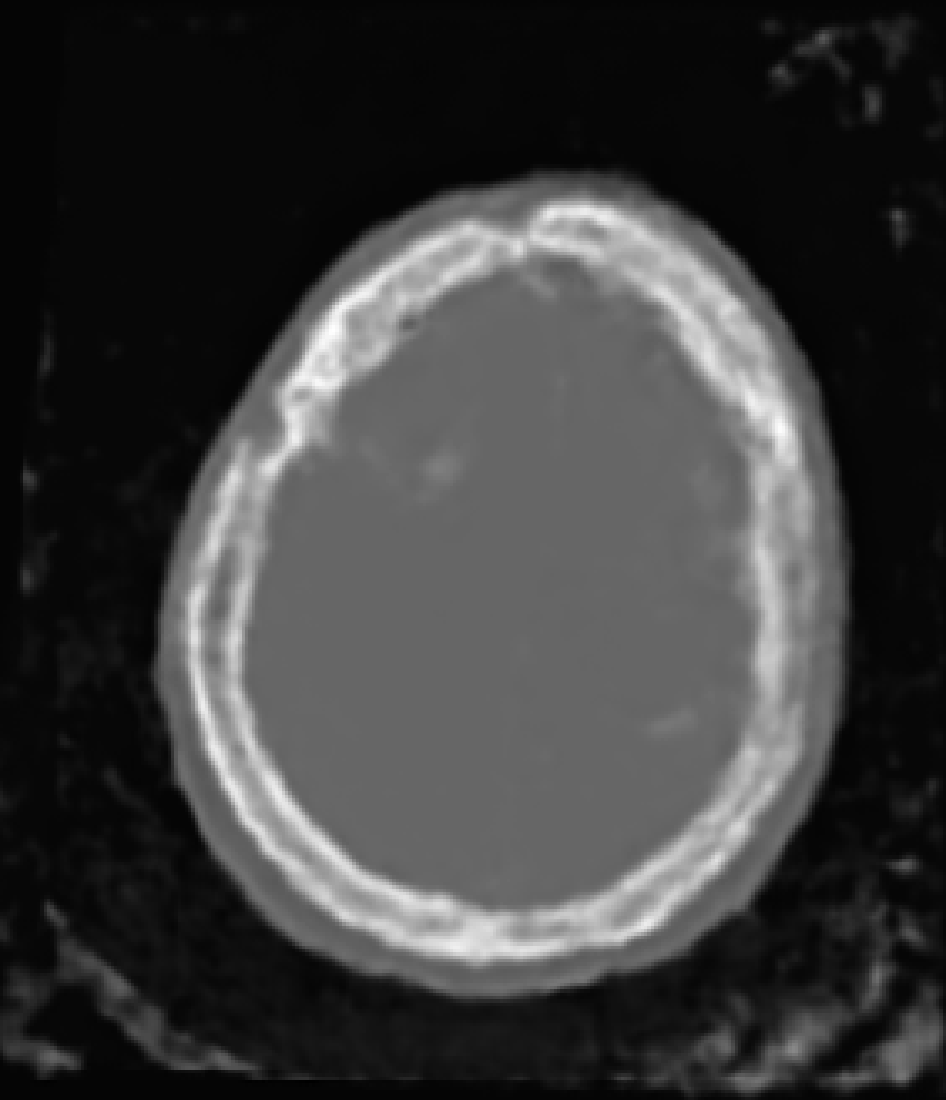}}%
\end{pgfscope}%
\begin{pgfscope}%
\pgfpathrectangle{\pgfqpoint{0.406740in}{3.898325in}}{\pgfqpoint{3.294941in}{3.665025in}}%
\pgfusepath{clip}%
\pgfsys@transformshift{0.406667in}{3.896667in}%
\pgftext[left,bottom]{\pgfimage[interpolate=true,width=3.296667in,height=3.663333in]{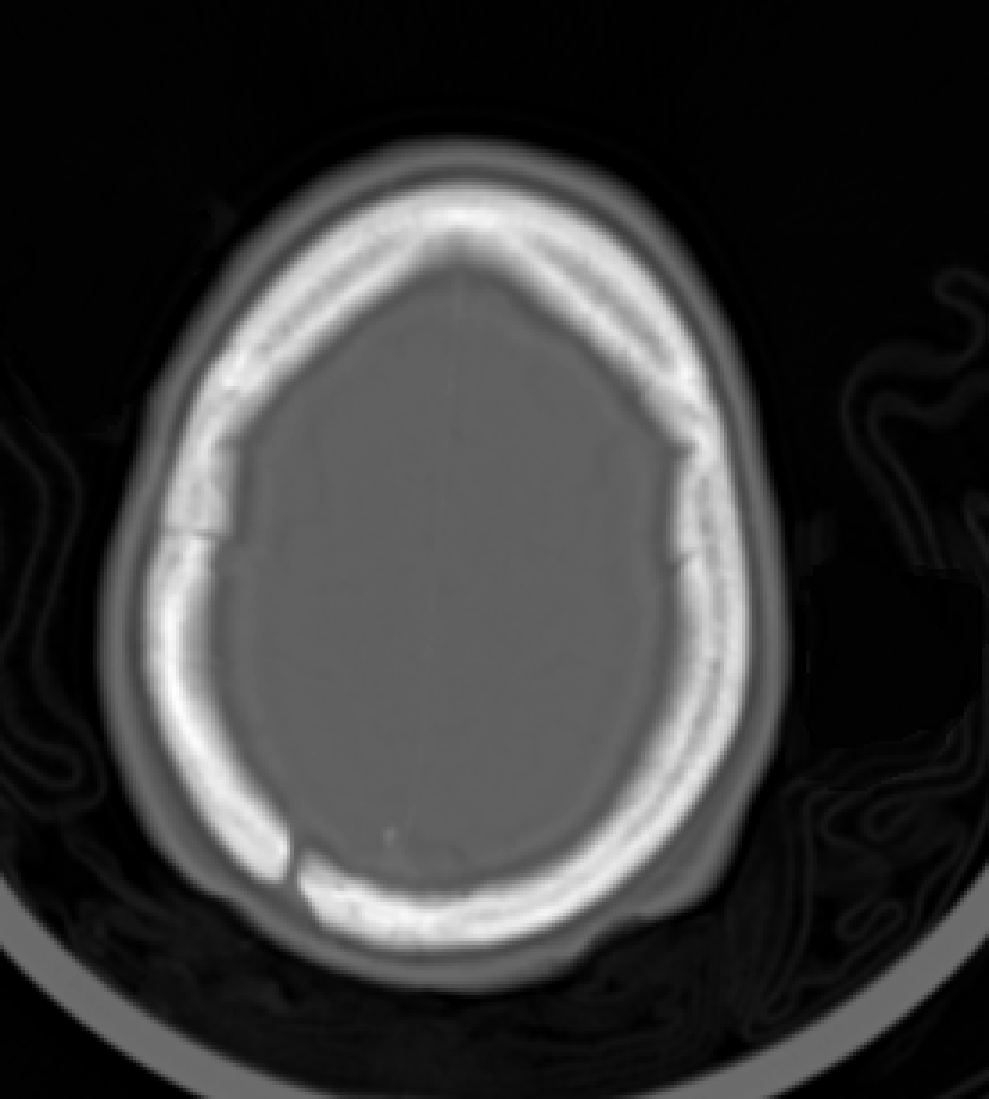}}%
\end{pgfscope}%
\begin{pgfscope}%
\pgfpathrectangle{\pgfqpoint{4.352530in}{3.898325in}}{\pgfqpoint{3.294941in}{3.665025in}}%
\pgfusepath{clip}%
\pgfsys@transformshift{4.353333in}{3.896667in}%
\pgftext[left,bottom]{\pgfimage[interpolate=true,width=3.296667in,height=3.663333in]{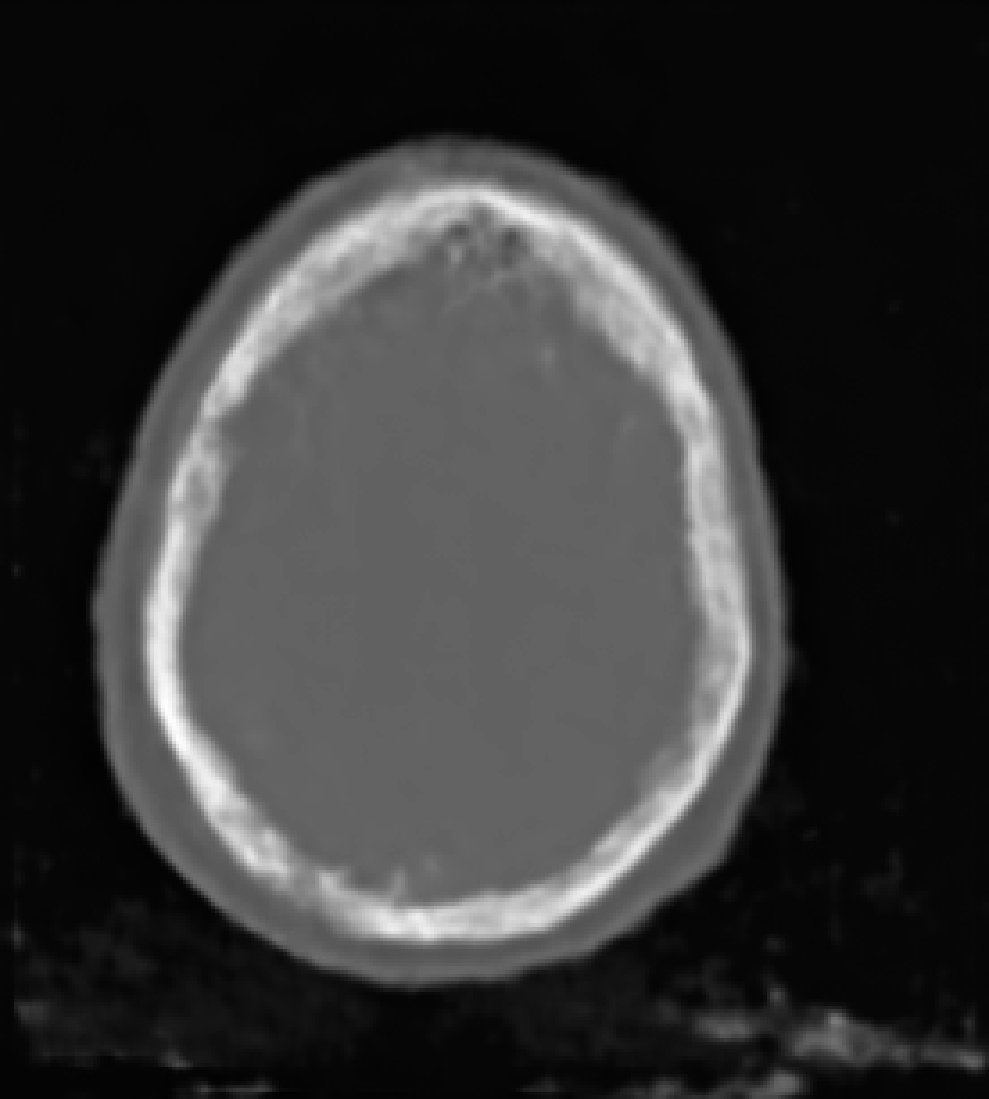}}%
\end{pgfscope}%
\begin{pgfscope}%
\pgfpathrectangle{\pgfqpoint{8.298319in}{3.898325in}}{\pgfqpoint{3.294941in}{3.665025in}}%
\pgfusepath{clip}%
\pgfsys@transformshift{8.296667in}{3.896667in}%
\pgftext[left,bottom]{\pgfimage[interpolate=true,width=3.296667in,height=3.663333in]{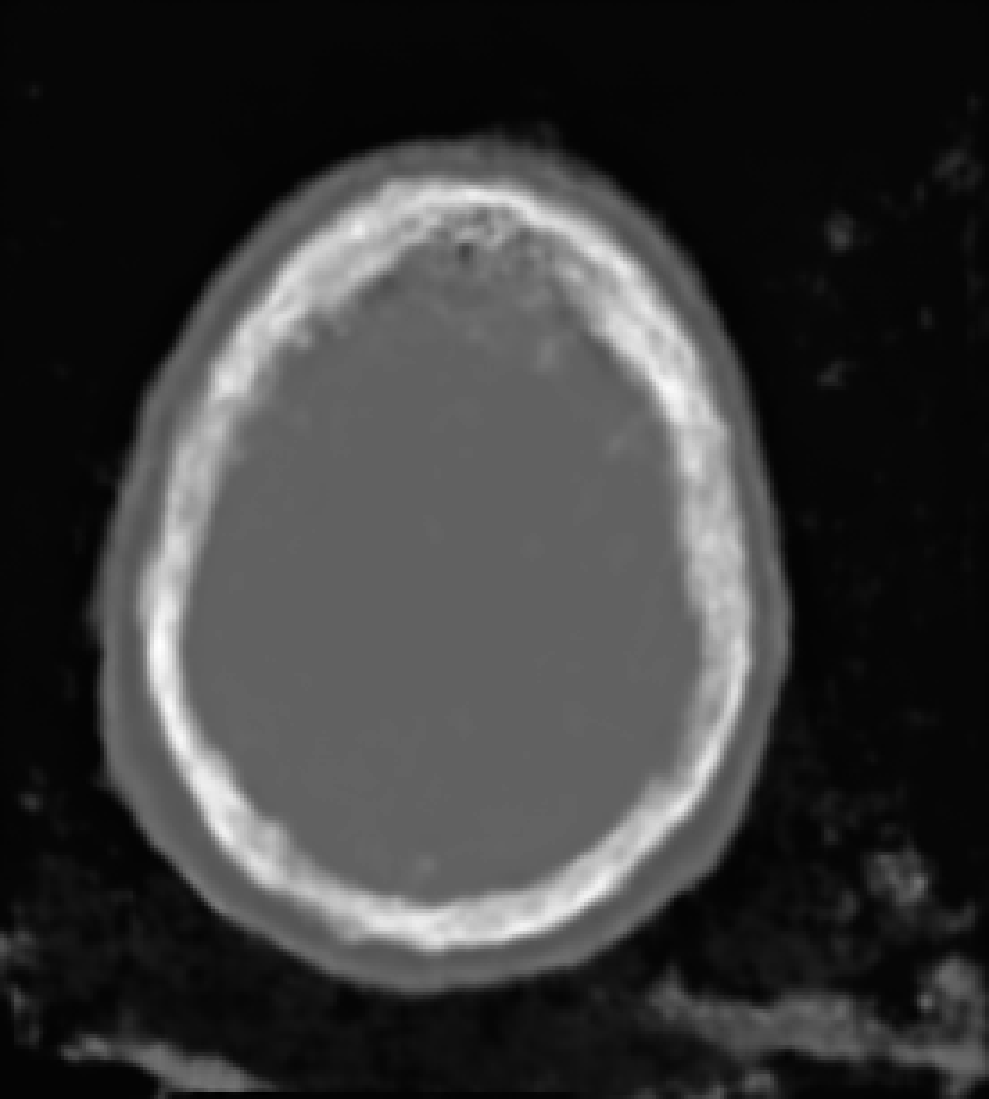}}%
\end{pgfscope}%
\begin{pgfscope}%
\pgfpathrectangle{\pgfqpoint{0.594885in}{0.160000in}}{\pgfqpoint{2.918652in}{3.665025in}}%
\pgfusepath{clip}%
\pgfsys@transformshift{0.593333in}{0.160000in}%
\pgftext[left,bottom]{\pgfimage[interpolate=true,width=2.916667in,height=3.666667in]{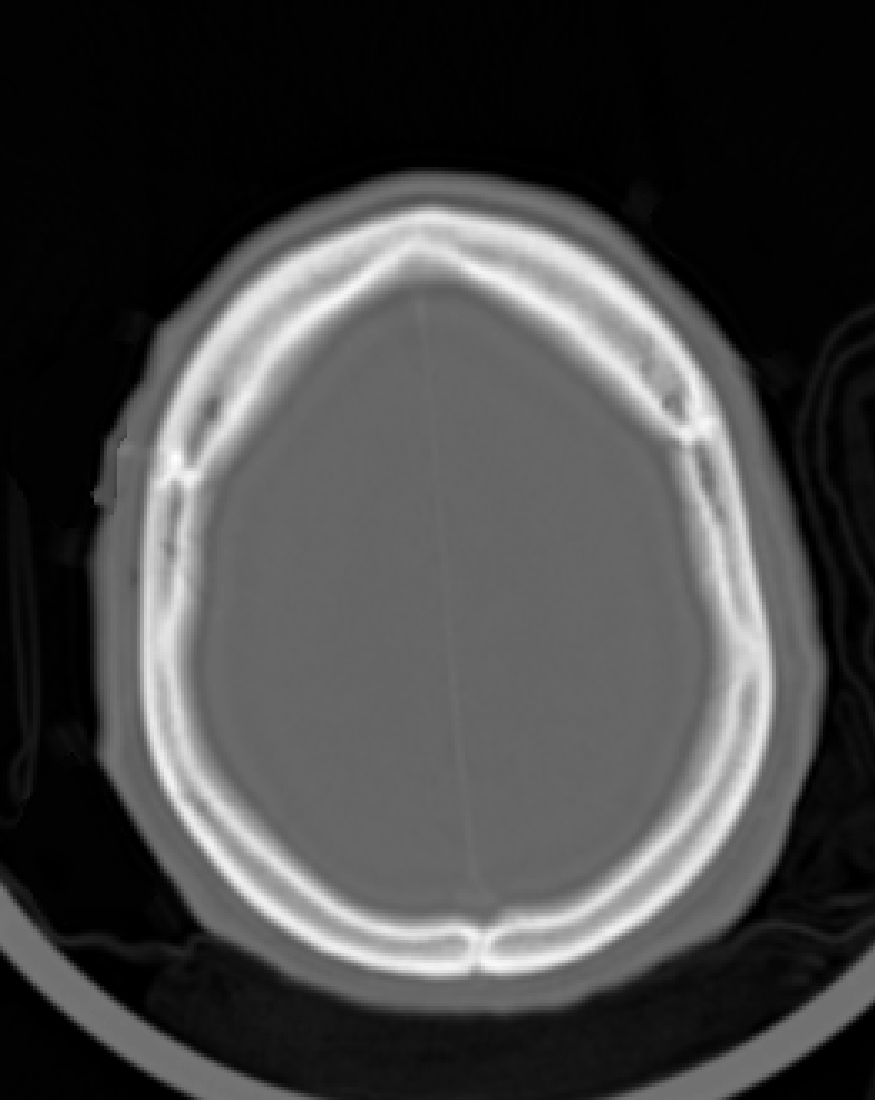}}%
\end{pgfscope}%
\begin{pgfscope}%
\pgfpathrectangle{\pgfqpoint{4.540674in}{0.160000in}}{\pgfqpoint{2.918652in}{3.665025in}}%
\pgfusepath{clip}%
\pgfsys@transformshift{4.540000in}{0.160000in}%
\pgftext[left,bottom]{\pgfimage[interpolate=true,width=2.920000in,height=3.666667in]{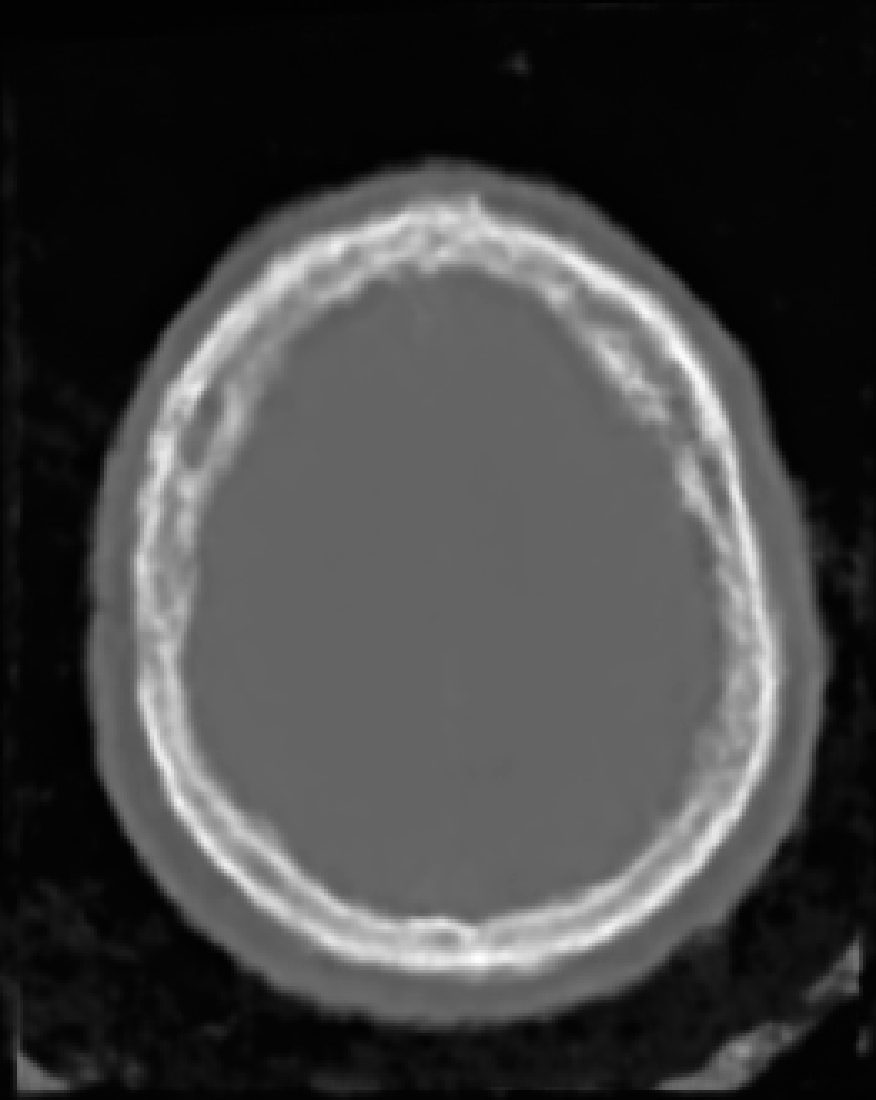}}%
\end{pgfscope}%
\begin{pgfscope}%
\pgfpathrectangle{\pgfqpoint{8.486464in}{0.160000in}}{\pgfqpoint{2.918652in}{3.665025in}}%
\pgfusepath{clip}%
\pgfsys@transformshift{8.486667in}{0.160000in}%
\pgftext[left,bottom]{\pgfimage[interpolate=true,width=2.920000in,height=3.666667in]{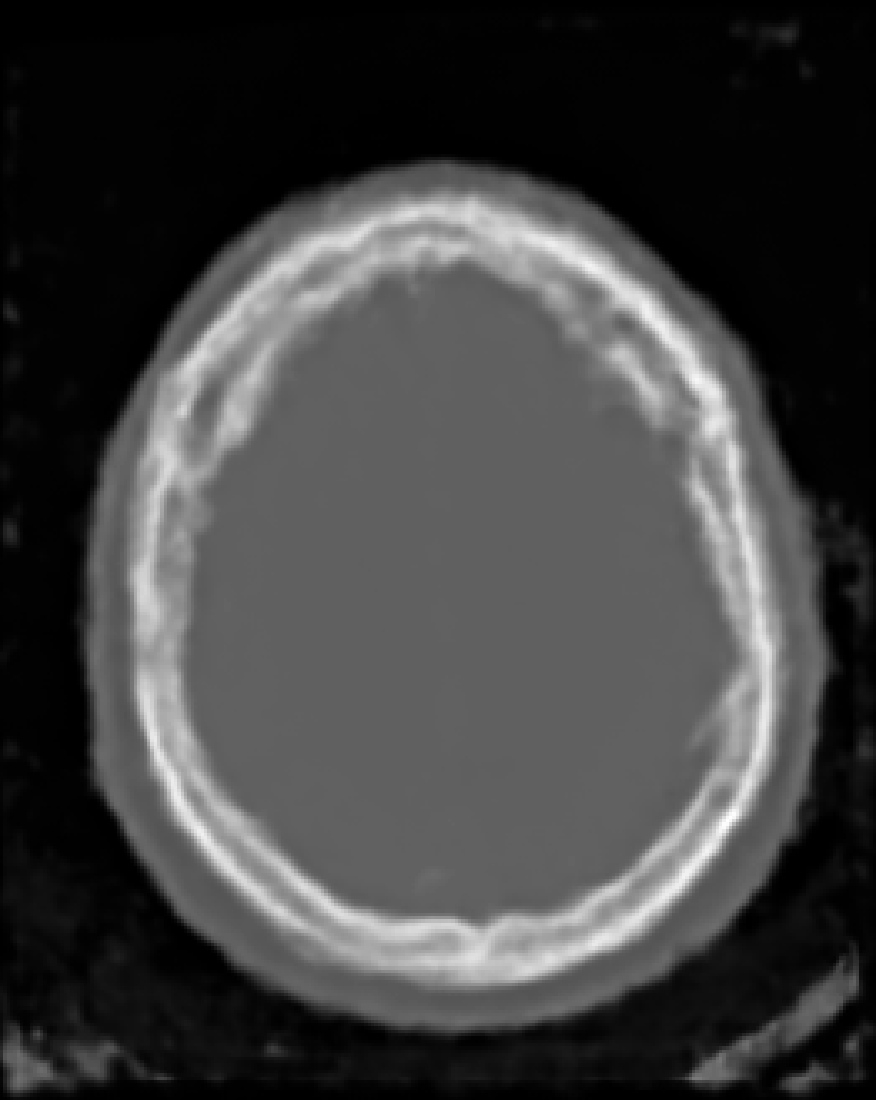}}%
\end{pgfscope}%
\end{pgfpicture}%
\makeatother%
\endgroup%
  \end{adjustbox}
  \caption{Transverse views for the u-net model trained with different
    loss functions, evaluated on the validation dataset.
  }\label{fig:unet:validation}
\end{figure}
In \Cref{fig:unet:validation} we show the transverse views for the differently
trained u-net models evaluated on the validation dataset. We can see that
the overall performance is much worse to unknown data which again suggests
overfitting. For the first two rows we note that the u-net trained with
\gls{gdl} loss seems more robust. We conclude that the \gls{gdl} improves
subjective performance on the validation dataset by a small amount, but
performance by standard metric seems to be decreased by a small amount.
Furthermore we want to mention, that the use of the \gls{gdl} requires high
computational cost.

\subsection{pixtopix}

In a second part we want to compare the u-net model trained on the \gls{mae}
with the pixtopix model. As already noted both models differ in that the
pixtopix uses a discriminator network in order to calculate a least-square
adverarial loss term. The adversarial least-square loss term was waited with
$\lambda_\text{adv}=0.01$ and the \gls{mae} term with $\lambda_\text{mae}=1$.

\begin{figure}[h]
  \centering
  \begin{adjustbox}{width=\linewidth}
\begingroup%
\makeatletter%
\begin{pgfpicture}%
\pgfpathrectangle{\pgfpointorigin}{\pgfqpoint{12.000000in}{16.000000in}}%
\pgfusepath{use as bounding box, clip}%
\begin{pgfscope}%
\pgfsetbuttcap%
\pgfsetmiterjoin%
\pgfsetlinewidth{0.000000pt}%
\definecolor{currentstroke}{rgb}{1.000000,1.000000,1.000000}%
\pgfsetstrokecolor{currentstroke}%
\pgfsetdash{}{0pt}%
\pgfpathmoveto{\pgfqpoint{0.000000in}{0.000000in}}%
\pgfpathlineto{\pgfqpoint{12.000000in}{0.000000in}}%
\pgfpathlineto{\pgfqpoint{12.000000in}{16.000000in}}%
\pgfpathlineto{\pgfqpoint{0.000000in}{16.000000in}}%
\pgfpathclose%
\pgfusepath{}%
\end{pgfscope}%
\begin{pgfscope}%
\pgfpathrectangle{\pgfqpoint{0.622560in}{11.374975in}}{\pgfqpoint{2.863300in}{3.665025in}}%
\pgfusepath{clip}%
\pgfsys@transformshift{0.623333in}{11.373333in}%
\pgftext[left,bottom]{\pgfimage[interpolate=true,width=2.860000in,height=3.666667in]{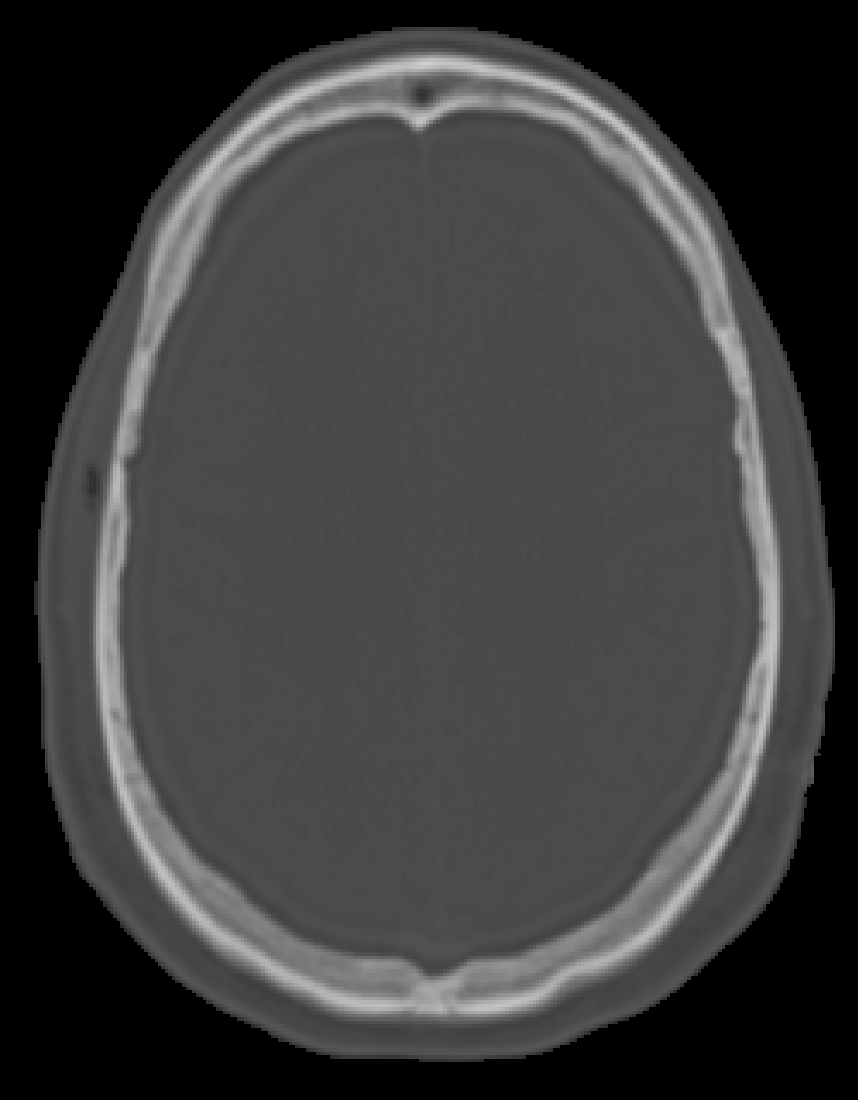}}%
\end{pgfscope}%
\begin{pgfscope}%
\pgftext[x=2.054211in,y=15.317778in,,base]{\fontsize{16.000000}{19.200000}\selectfont Ground Truth}%
\end{pgfscope}%
\begin{pgfscope}%
\pgfpathrectangle{\pgfqpoint{4.568350in}{11.374975in}}{\pgfqpoint{2.863300in}{3.665025in}}%
\pgfusepath{clip}%
\pgfsys@transformshift{4.570000in}{11.373333in}%
\pgftext[left,bottom]{\pgfimage[interpolate=true,width=2.863333in,height=3.666667in]{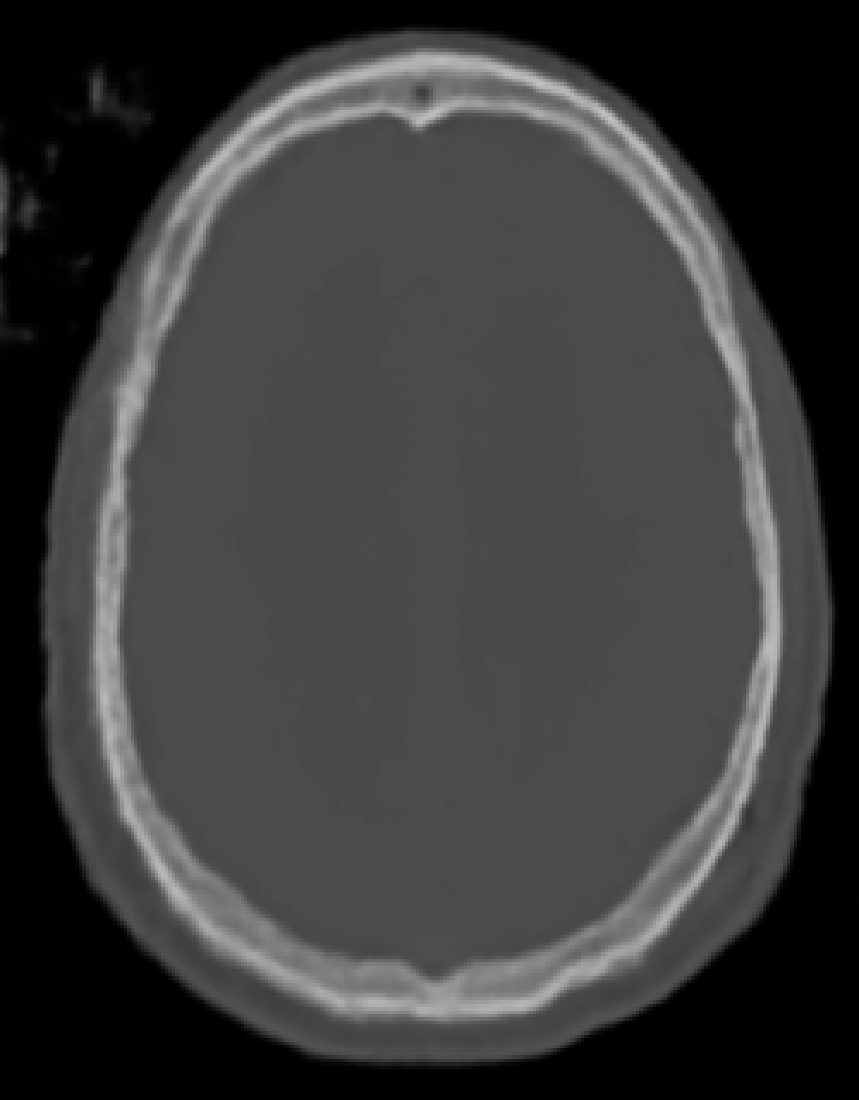}}%
\end{pgfscope}%
\begin{pgfscope}%
\pgftext[x=6.000000in,y=15.317778in,,base]{\fontsize{16.000000}{19.200000}\selectfont u-net}%
\end{pgfscope}%
\begin{pgfscope}%
\pgfpathrectangle{\pgfqpoint{8.514139in}{11.374975in}}{\pgfqpoint{2.863300in}{3.665025in}}%
\pgfusepath{clip}%
\pgfsys@transformshift{8.513333in}{11.373333in}%
\pgftext[left,bottom]{\pgfimage[interpolate=true,width=2.863333in,height=3.666667in]{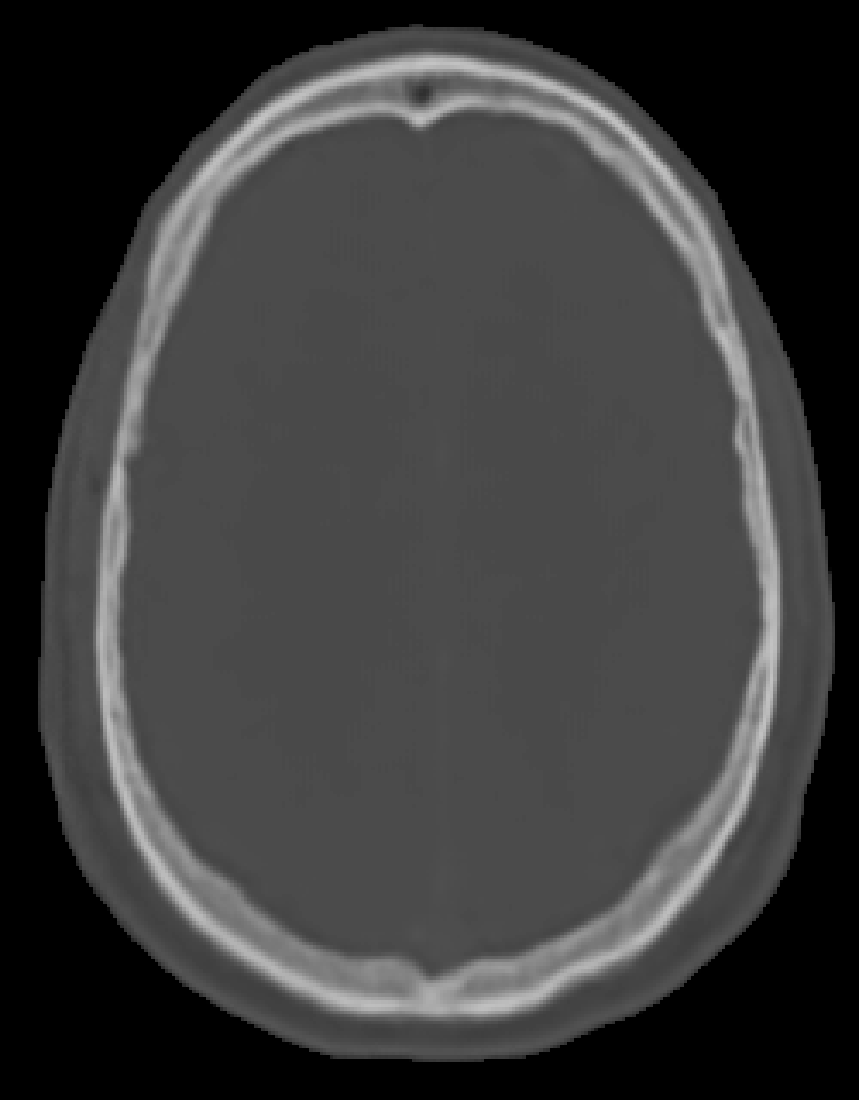}}%
\end{pgfscope}%
\begin{pgfscope}%
\pgftext[x=9.945789in,y=15.317778in,,base]{\fontsize{16.000000}{19.200000}\selectfont pixtopix}%
\end{pgfscope}%
\begin{pgfscope}%
\pgfpathrectangle{\pgfqpoint{0.657479in}{7.636650in}}{\pgfqpoint{2.793464in}{3.665025in}}%
\pgfusepath{clip}%
\pgfsys@transformshift{0.656667in}{7.636667in}%
\pgftext[left,bottom]{\pgfimage[interpolate=true,width=2.796667in,height=3.666667in]{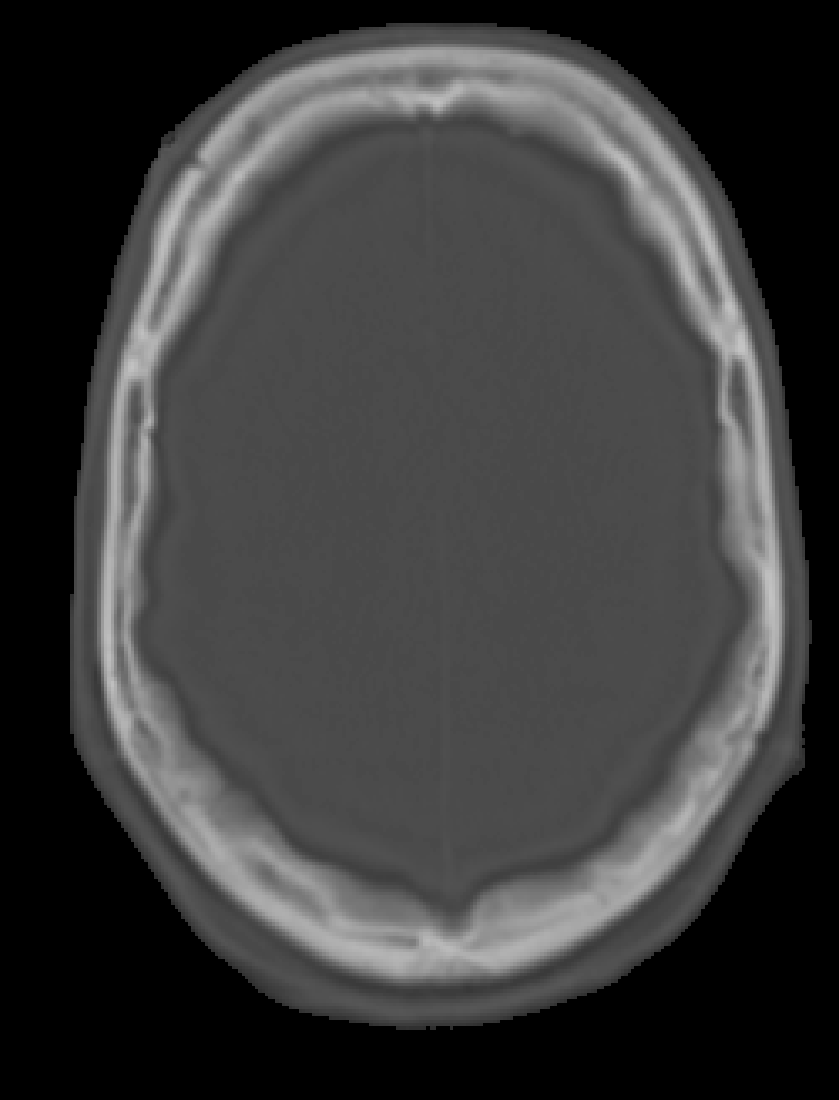}}%
\end{pgfscope}%
\begin{pgfscope}%
\pgfpathrectangle{\pgfqpoint{4.603268in}{7.636650in}}{\pgfqpoint{2.793464in}{3.665025in}}%
\pgfusepath{clip}%
\pgfsys@transformshift{4.603333in}{7.636667in}%
\pgftext[left,bottom]{\pgfimage[interpolate=true,width=2.796667in,height=3.666667in]{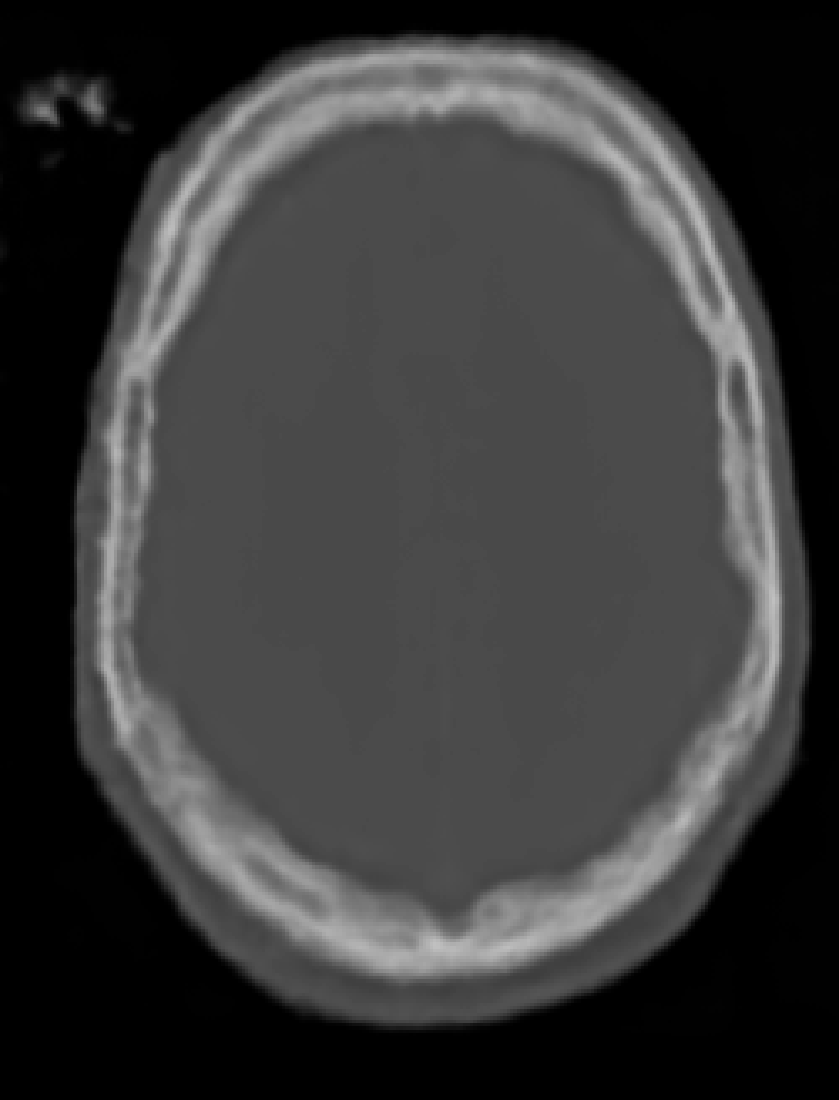}}%
\end{pgfscope}%
\begin{pgfscope}%
\pgfpathrectangle{\pgfqpoint{8.549058in}{7.636650in}}{\pgfqpoint{2.793464in}{3.665025in}}%
\pgfusepath{clip}%
\pgfsys@transformshift{8.550000in}{7.636667in}%
\pgftext[left,bottom]{\pgfimage[interpolate=true,width=2.796667in,height=3.666667in]{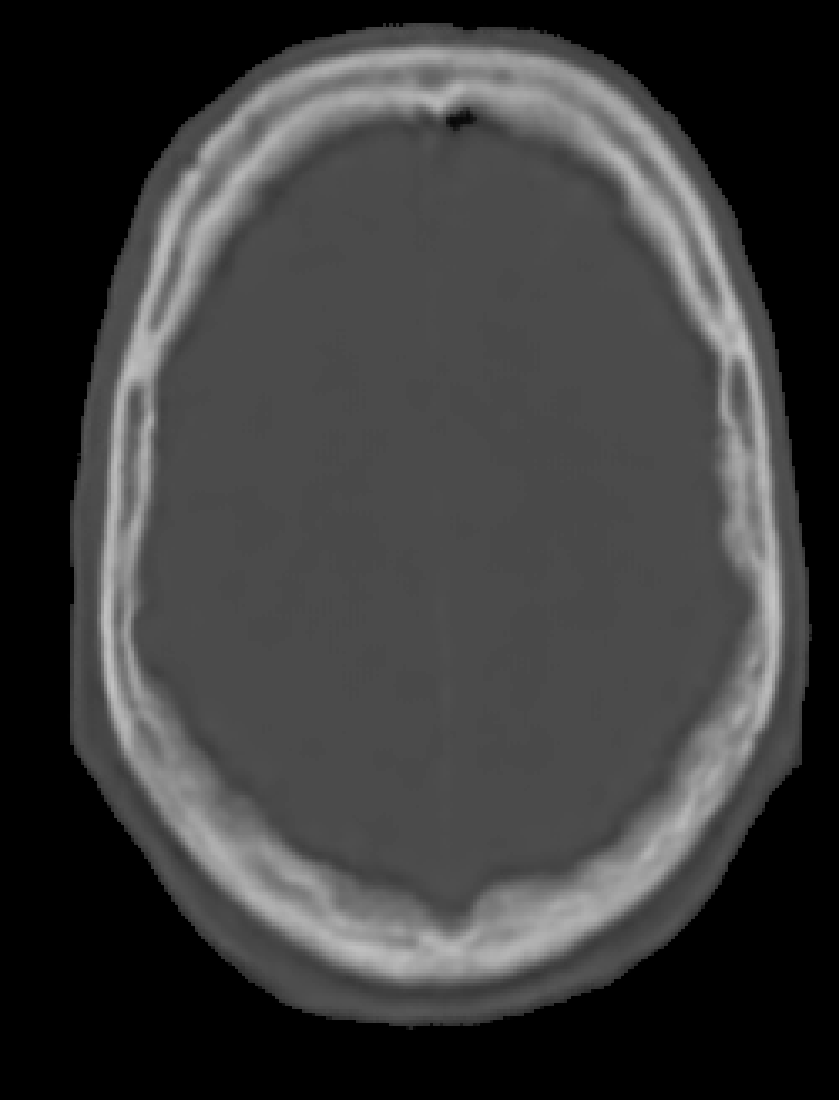}}%
\end{pgfscope}%
\begin{pgfscope}%
\pgfpathrectangle{\pgfqpoint{0.354752in}{3.898325in}}{\pgfqpoint{3.398917in}{3.665025in}}%
\pgfusepath{clip}%
\pgfsys@transformshift{0.353333in}{3.896667in}%
\pgftext[left,bottom]{\pgfimage[interpolate=true,width=3.400000in,height=3.663333in]{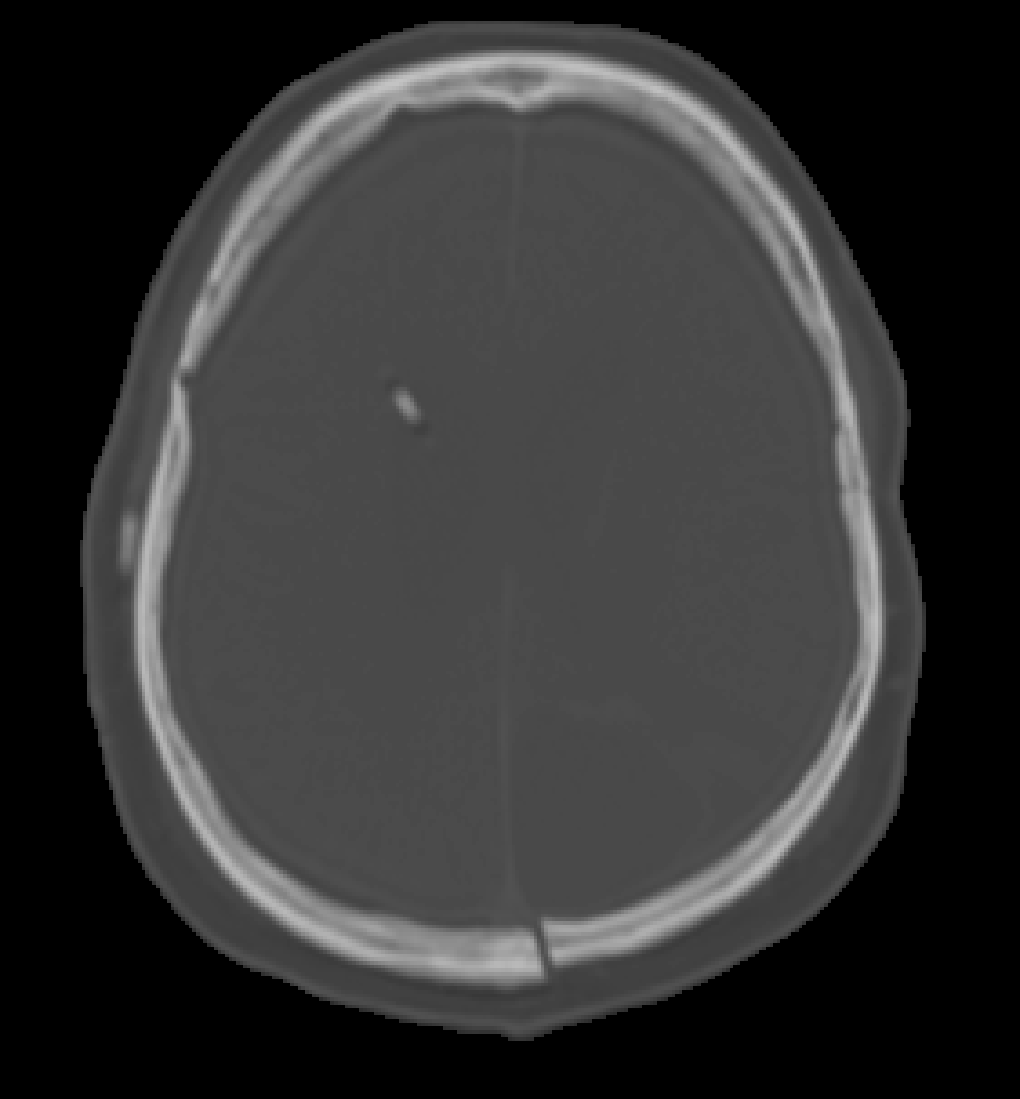}}%
\end{pgfscope}%
\begin{pgfscope}%
\pgfpathrectangle{\pgfqpoint{4.300541in}{3.898325in}}{\pgfqpoint{3.398917in}{3.665025in}}%
\pgfusepath{clip}%
\pgfsys@transformshift{4.300000in}{3.896667in}%
\pgftext[left,bottom]{\pgfimage[interpolate=true,width=3.400000in,height=3.663333in]{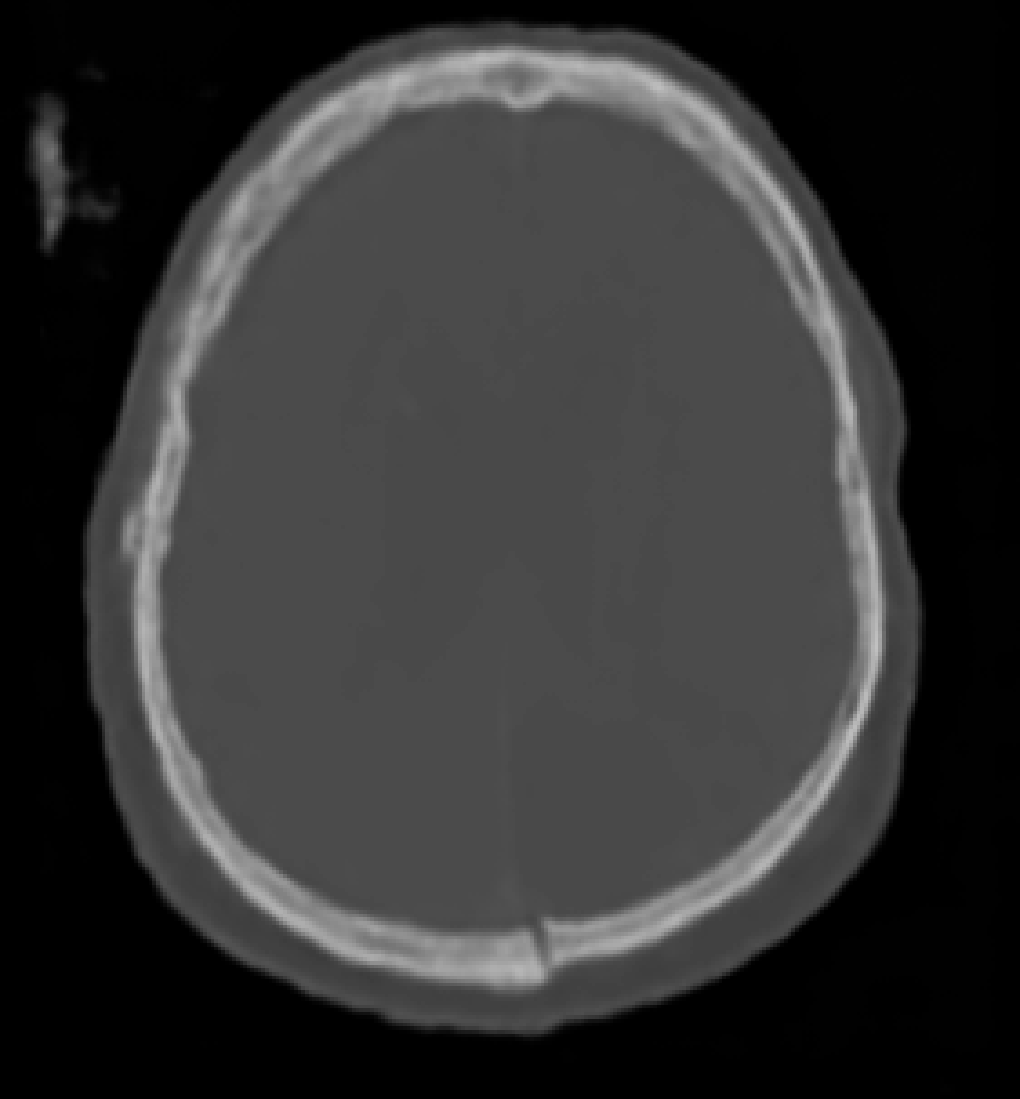}}%
\end{pgfscope}%
\begin{pgfscope}%
\pgfpathrectangle{\pgfqpoint{8.246331in}{3.898325in}}{\pgfqpoint{3.398917in}{3.665025in}}%
\pgfusepath{clip}%
\pgfsys@transformshift{8.246667in}{3.896667in}%
\pgftext[left,bottom]{\pgfimage[interpolate=true,width=3.400000in,height=3.663333in]{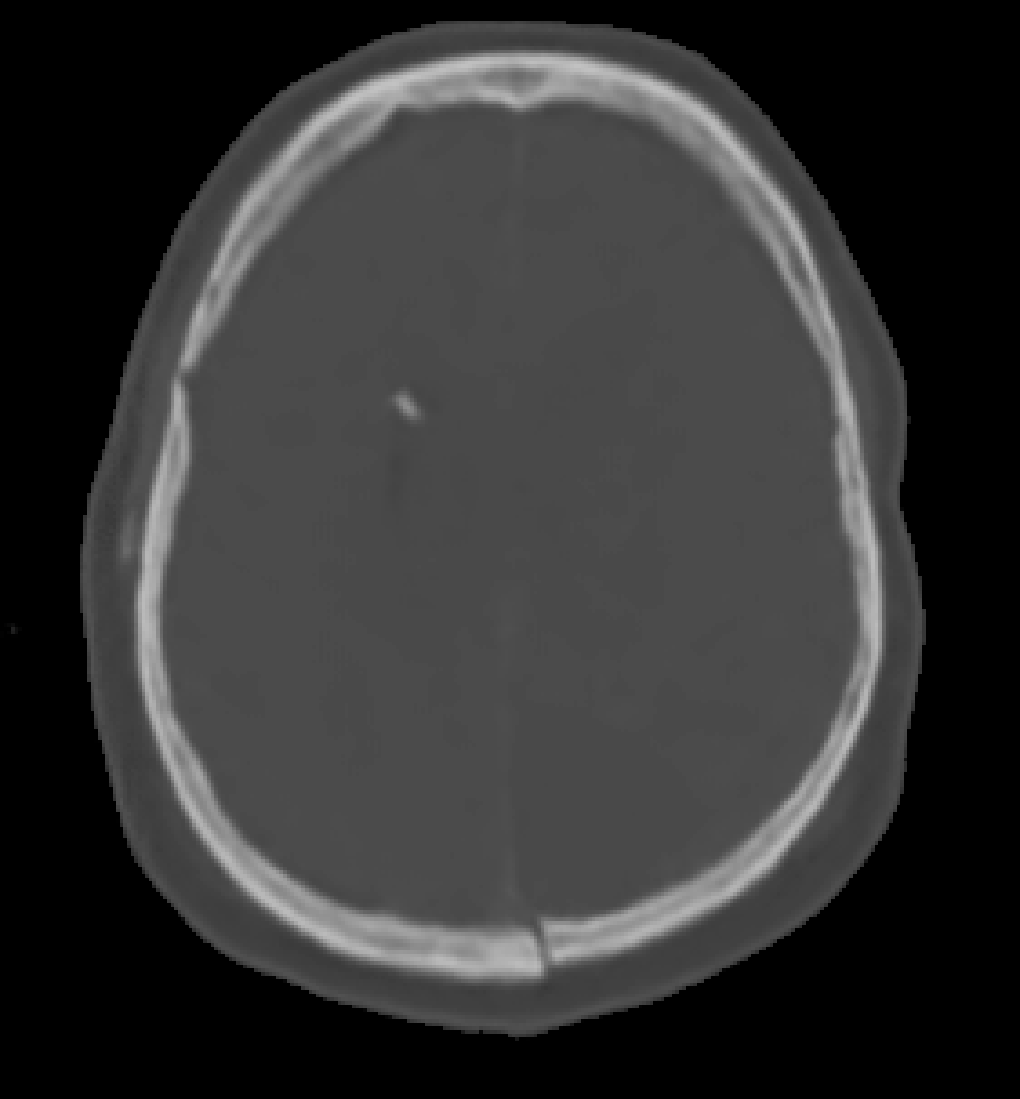}}%
\end{pgfscope}%
\begin{pgfscope}%
\pgfpathrectangle{\pgfqpoint{0.423212in}{0.160000in}}{\pgfqpoint{3.261998in}{3.665025in}}%
\pgfusepath{clip}%
\pgfsys@transformshift{0.423333in}{0.160000in}%
\pgftext[left,bottom]{\pgfimage[interpolate=true,width=3.263333in,height=3.666667in]{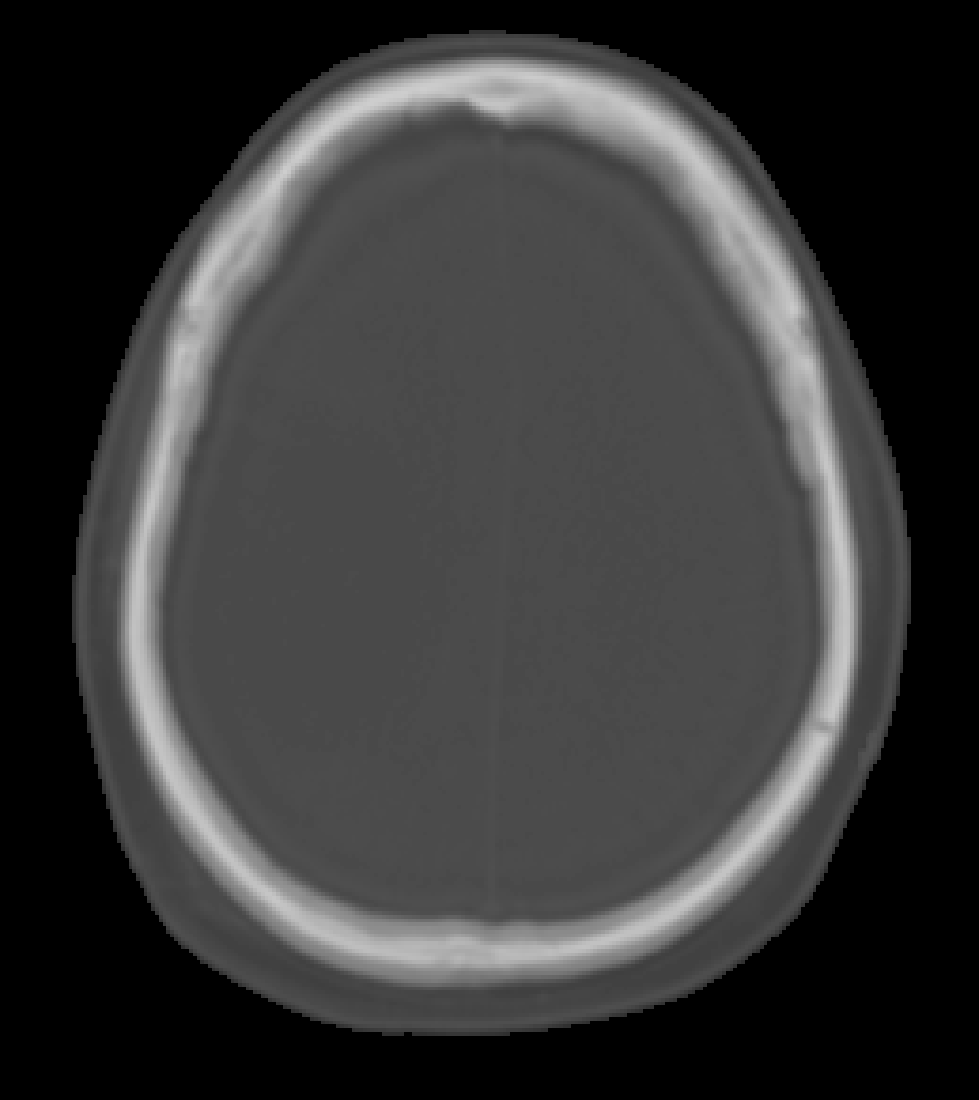}}%
\end{pgfscope}%
\begin{pgfscope}%
\pgfpathrectangle{\pgfqpoint{4.369001in}{0.160000in}}{\pgfqpoint{3.261998in}{3.665025in}}%
\pgfusepath{clip}%
\pgfsys@transformshift{4.370000in}{0.160000in}%
\pgftext[left,bottom]{\pgfimage[interpolate=true,width=3.260000in,height=3.666667in]{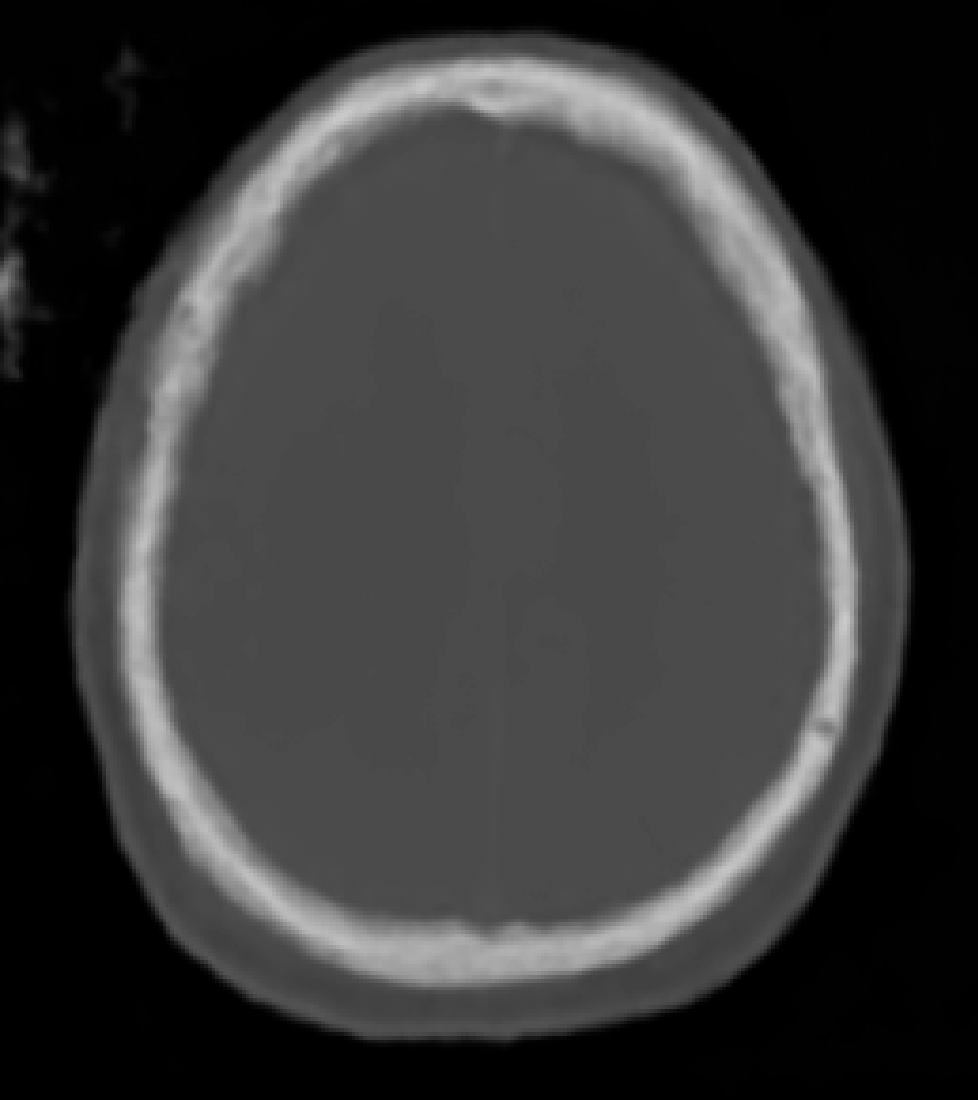}}%
\end{pgfscope}%
\begin{pgfscope}%
\pgfpathrectangle{\pgfqpoint{8.314791in}{0.160000in}}{\pgfqpoint{3.261998in}{3.665025in}}%
\pgfusepath{clip}%
\pgfsys@transformshift{8.313333in}{0.160000in}%
\pgftext[left,bottom]{\pgfimage[interpolate=true,width=3.263333in,height=3.666667in]{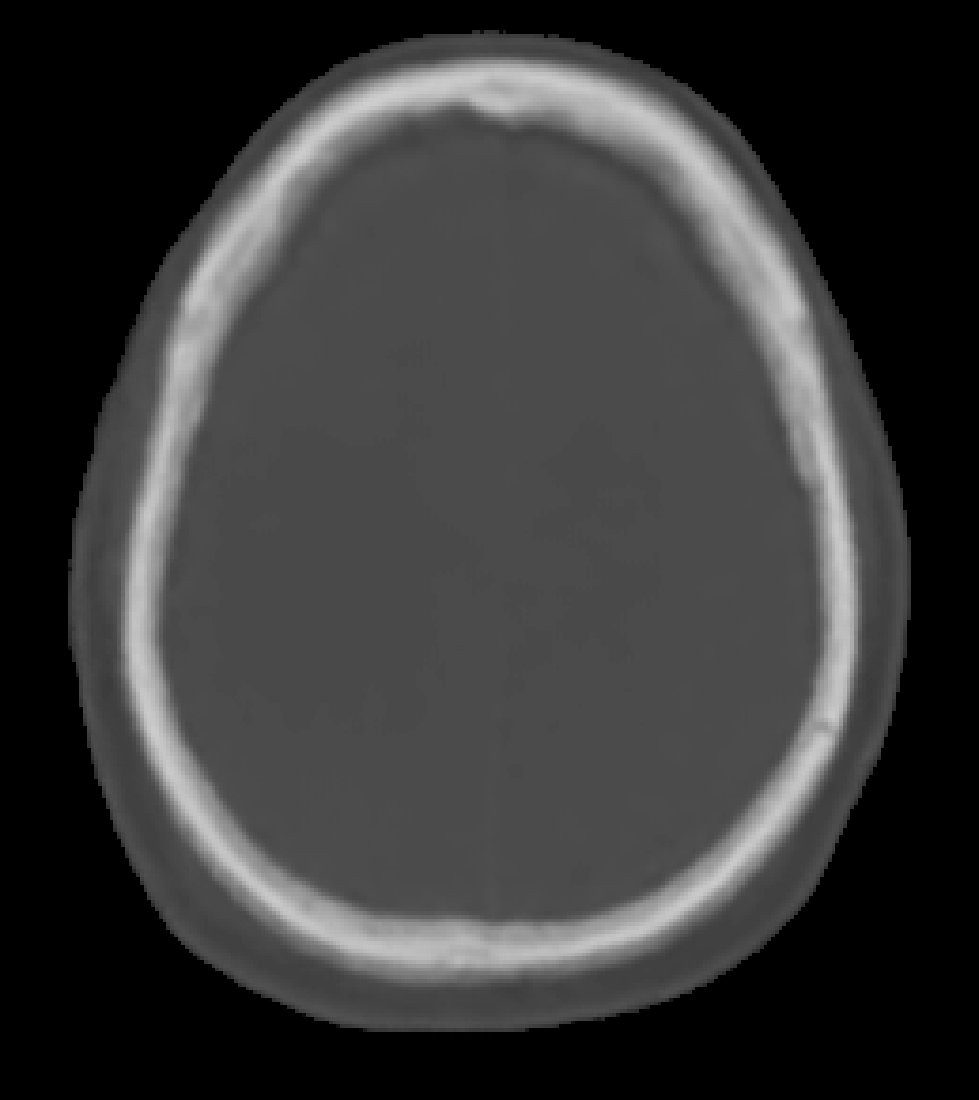}}%
\end{pgfscope}%
\end{pgfpicture}%
\makeatother%
\endgroup%
  \end{adjustbox}
  \caption{Transverse views for the u-net model trained with \acrshort{mae}
    loss compared with the pixtopix model trained with least-square adversarial
    loss, evaluated on the training dataset.
  }\label{fig:unet_pixtopix:training}
\end{figure}

\begin{figure}[h]
  \centering
  \begin{adjustbox}{width=\linewidth}
\begingroup%
\makeatletter%
\begin{pgfpicture}%
\pgfpathrectangle{\pgfpointorigin}{\pgfqpoint{12.000000in}{16.000000in}}%
\pgfusepath{use as bounding box, clip}%
\begin{pgfscope}%
\pgfsetbuttcap%
\pgfsetmiterjoin%
\pgfsetlinewidth{0.000000pt}%
\definecolor{currentstroke}{rgb}{1.000000,1.000000,1.000000}%
\pgfsetstrokecolor{currentstroke}%
\pgfsetdash{}{0pt}%
\pgfpathmoveto{\pgfqpoint{0.000000in}{0.000000in}}%
\pgfpathlineto{\pgfqpoint{12.000000in}{0.000000in}}%
\pgfpathlineto{\pgfqpoint{12.000000in}{16.000000in}}%
\pgfpathlineto{\pgfqpoint{0.000000in}{16.000000in}}%
\pgfpathclose%
\pgfusepath{}%
\end{pgfscope}%
\begin{pgfscope}%
\pgfpathrectangle{\pgfqpoint{0.658303in}{11.374975in}}{\pgfqpoint{2.791814in}{3.665025in}}%
\pgfusepath{clip}%
\pgfsys@transformshift{0.656667in}{11.373333in}%
\pgftext[left,bottom]{\pgfimage[interpolate=true,width=2.793333in,height=3.666667in]{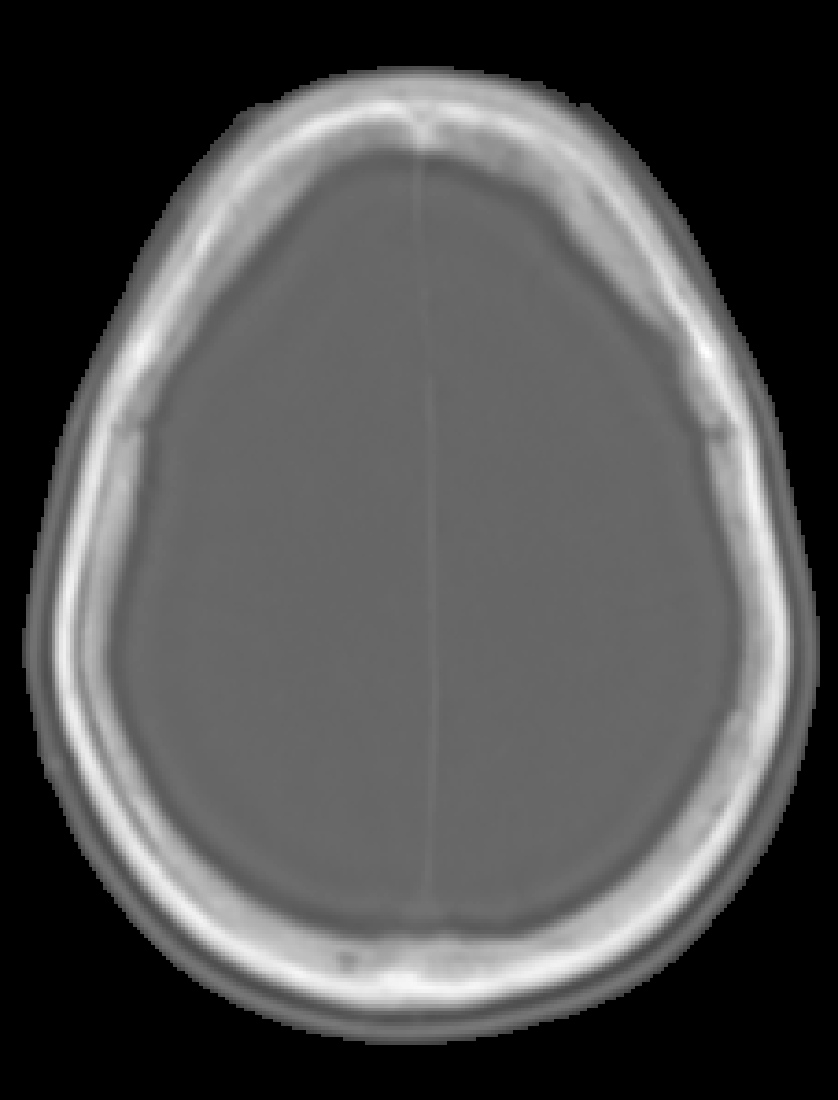}}%
\end{pgfscope}%
\begin{pgfscope}%
\pgftext[x=2.054211in,y=15.317778in,,base]{\fontsize{16.000000}{19.200000}\selectfont Ground Truth}%
\end{pgfscope}%
\begin{pgfscope}%
\pgfpathrectangle{\pgfqpoint{4.604093in}{11.374975in}}{\pgfqpoint{2.791814in}{3.665025in}}%
\pgfusepath{clip}%
\pgfsys@transformshift{4.603333in}{11.373333in}%
\pgftext[left,bottom]{\pgfimage[interpolate=true,width=2.793333in,height=3.666667in]{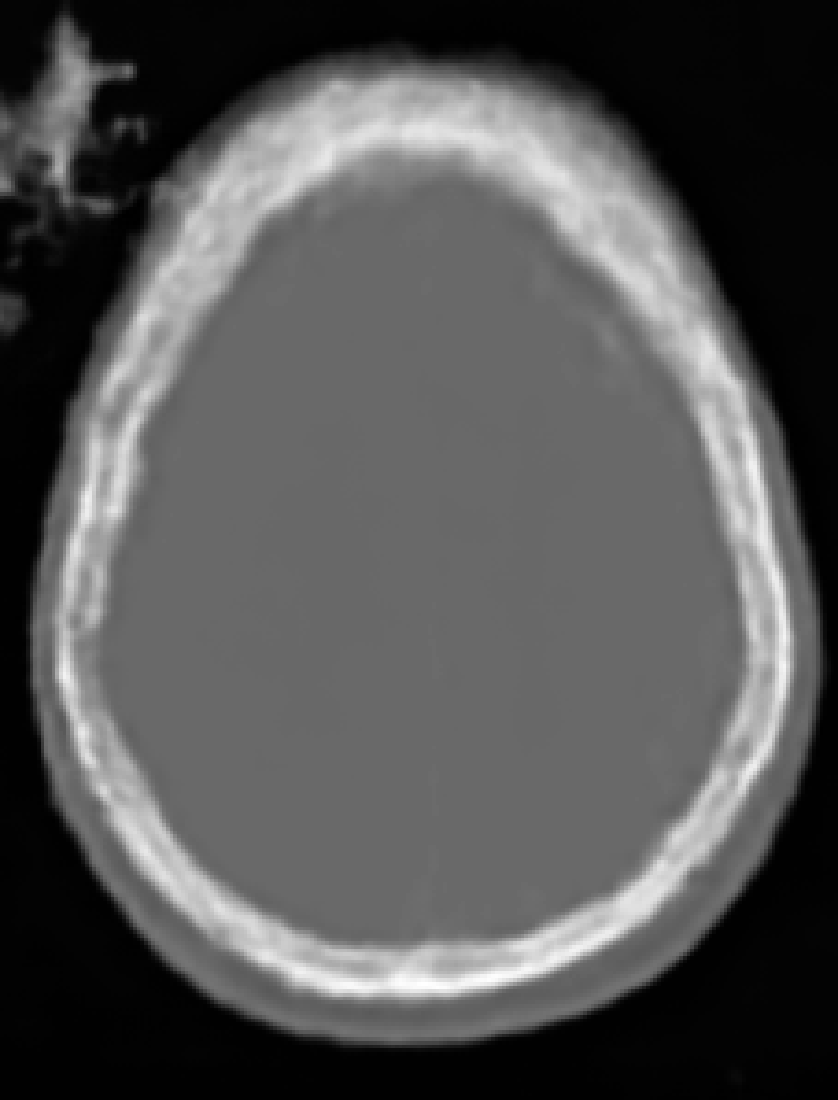}}%
\end{pgfscope}%
\begin{pgfscope}%
\pgftext[x=6.000000in,y=15.317778in,,base]{\fontsize{16.000000}{19.200000}\selectfont u-net}%
\end{pgfscope}%
\begin{pgfscope}%
\pgfpathrectangle{\pgfqpoint{8.549882in}{11.374975in}}{\pgfqpoint{2.791814in}{3.665025in}}%
\pgfusepath{clip}%
\pgfsys@transformshift{8.550000in}{11.373333in}%
\pgftext[left,bottom]{\pgfimage[interpolate=true,width=2.793333in,height=3.666667in]{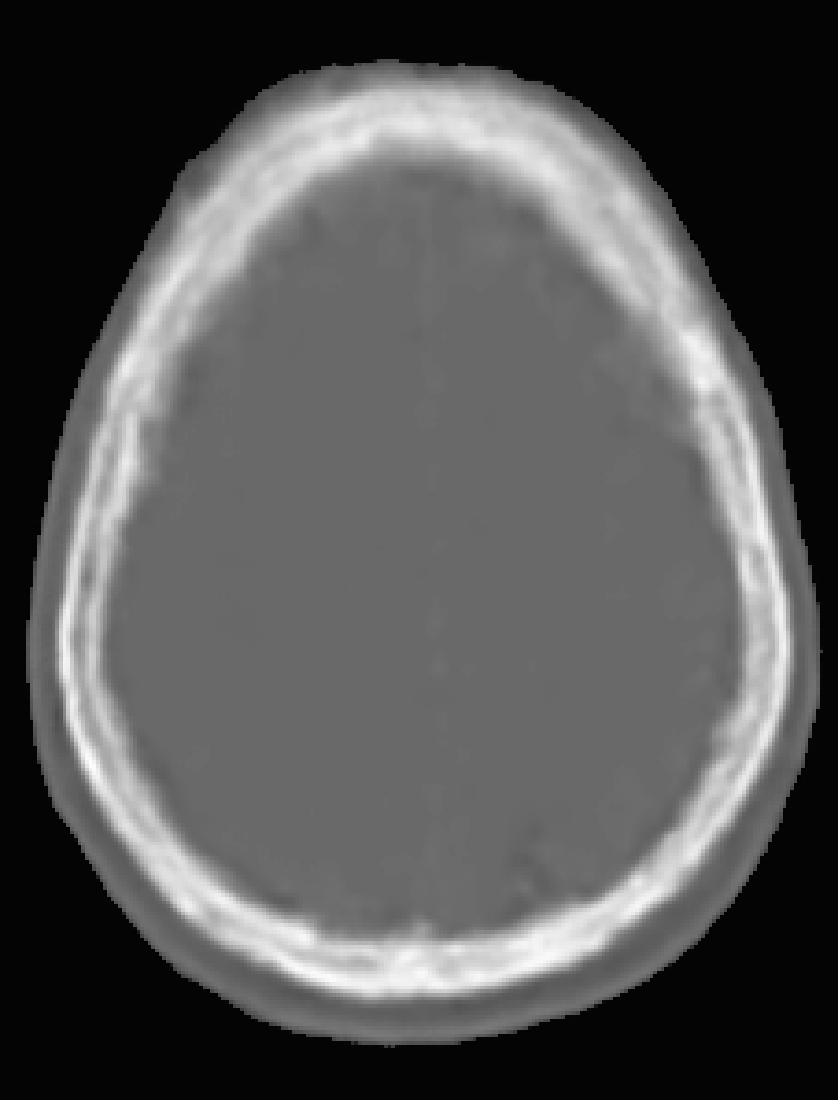}}%
\end{pgfscope}%
\begin{pgfscope}%
\pgftext[x=9.945789in,y=15.317778in,,base]{\fontsize{16.000000}{19.200000}\selectfont pixtopix}%
\end{pgfscope}%
\begin{pgfscope}%
\pgfpathrectangle{\pgfqpoint{0.478697in}{7.636650in}}{\pgfqpoint{3.151027in}{3.665025in}}%
\pgfusepath{clip}%
\pgfsys@transformshift{0.480000in}{7.636667in}%
\pgftext[left,bottom]{\pgfimage[interpolate=true,width=3.153333in,height=3.666667in]{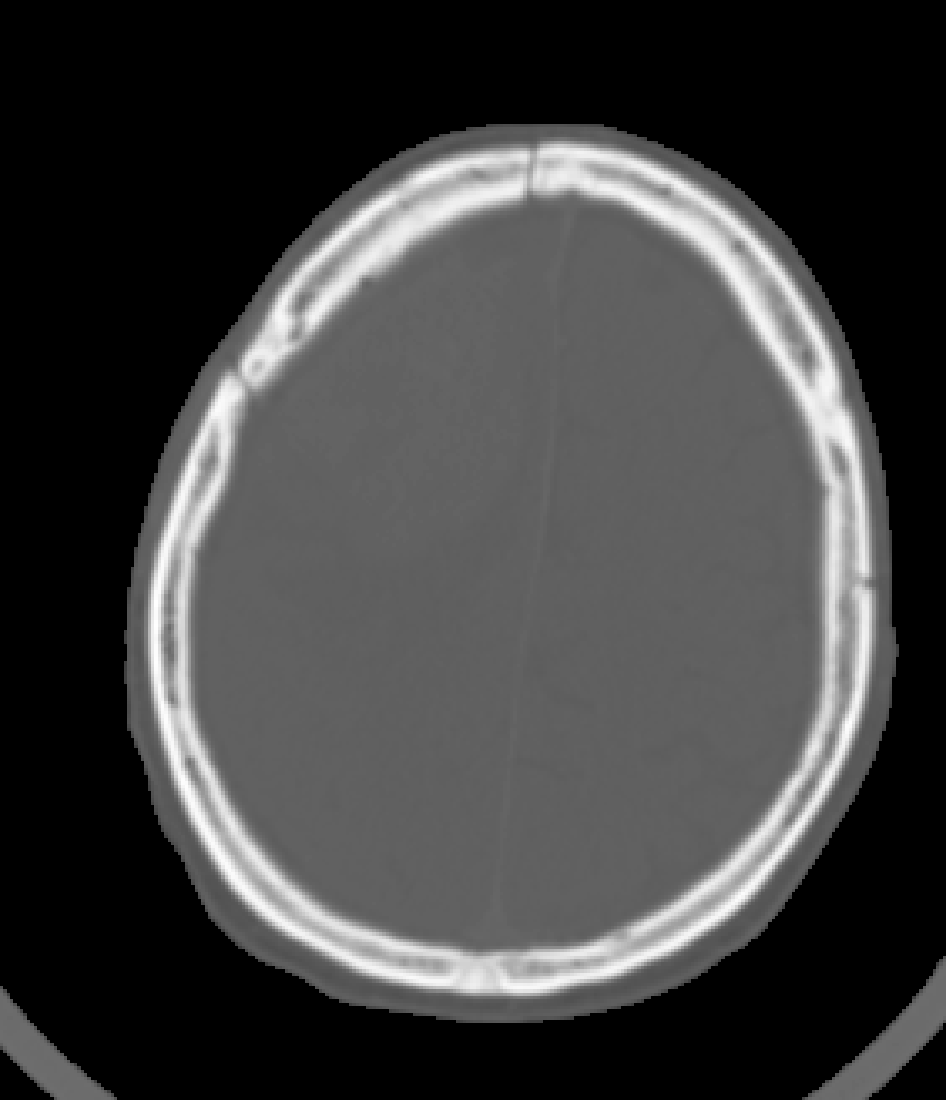}}%
\end{pgfscope}%
\begin{pgfscope}%
\pgfpathrectangle{\pgfqpoint{4.424486in}{7.636650in}}{\pgfqpoint{3.151027in}{3.665025in}}%
\pgfusepath{clip}%
\pgfsys@transformshift{4.423333in}{7.636667in}%
\pgftext[left,bottom]{\pgfimage[interpolate=true,width=3.153333in,height=3.666667in]{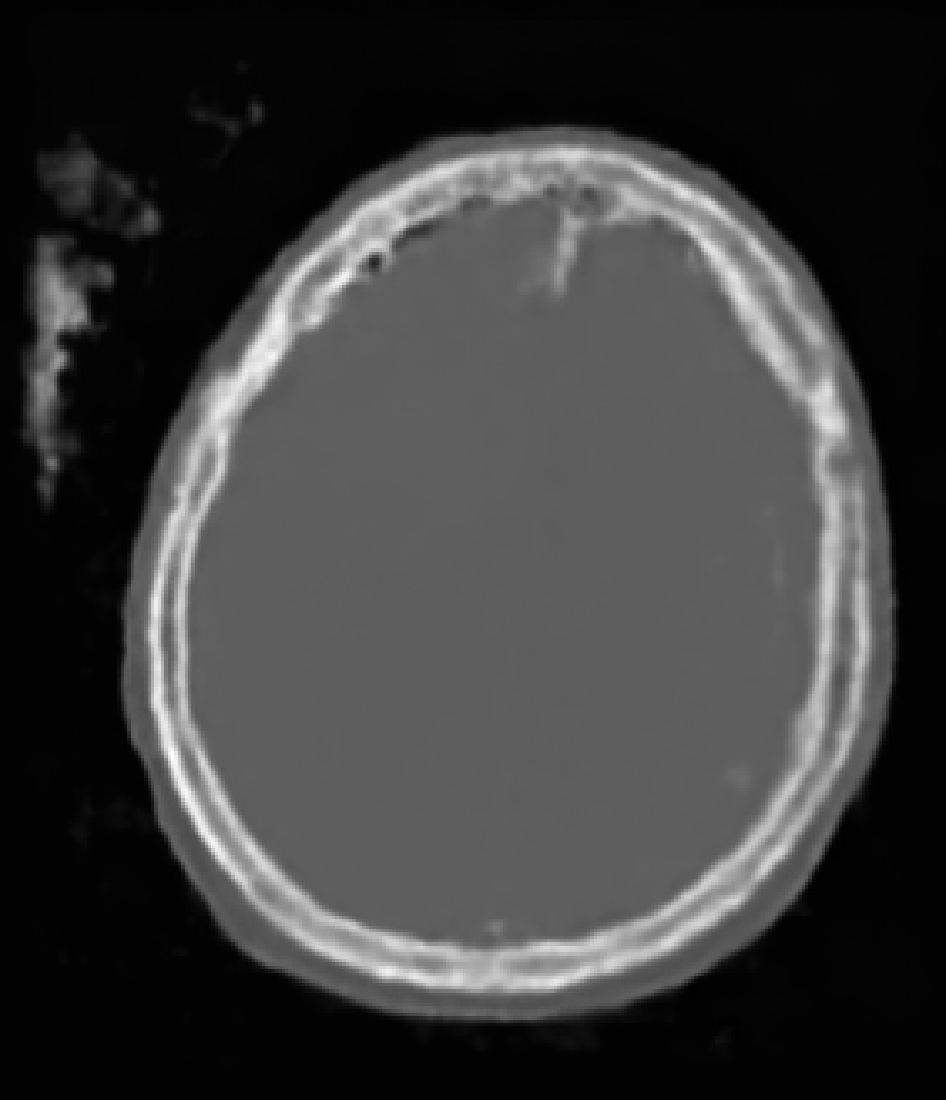}}%
\end{pgfscope}%
\begin{pgfscope}%
\pgfpathrectangle{\pgfqpoint{8.370276in}{7.636650in}}{\pgfqpoint{3.151027in}{3.665025in}}%
\pgfusepath{clip}%
\pgfsys@transformshift{8.370000in}{7.636667in}%
\pgftext[left,bottom]{\pgfimage[interpolate=true,width=3.153333in,height=3.666667in]{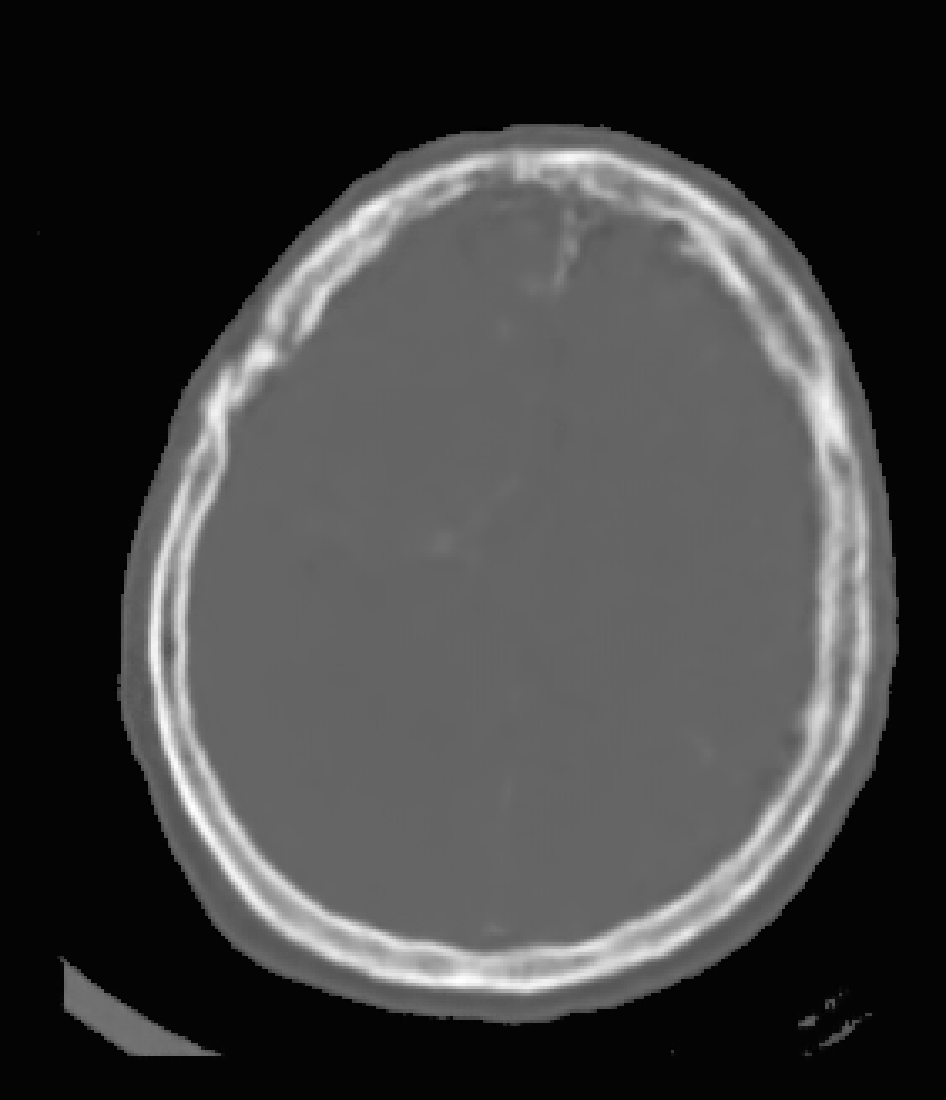}}%
\end{pgfscope}%
\begin{pgfscope}%
\pgfpathrectangle{\pgfqpoint{0.406740in}{3.898325in}}{\pgfqpoint{3.294941in}{3.665025in}}%
\pgfusepath{clip}%
\pgfsys@transformshift{0.406667in}{3.896667in}%
\pgftext[left,bottom]{\pgfimage[interpolate=true,width=3.296667in,height=3.663333in]{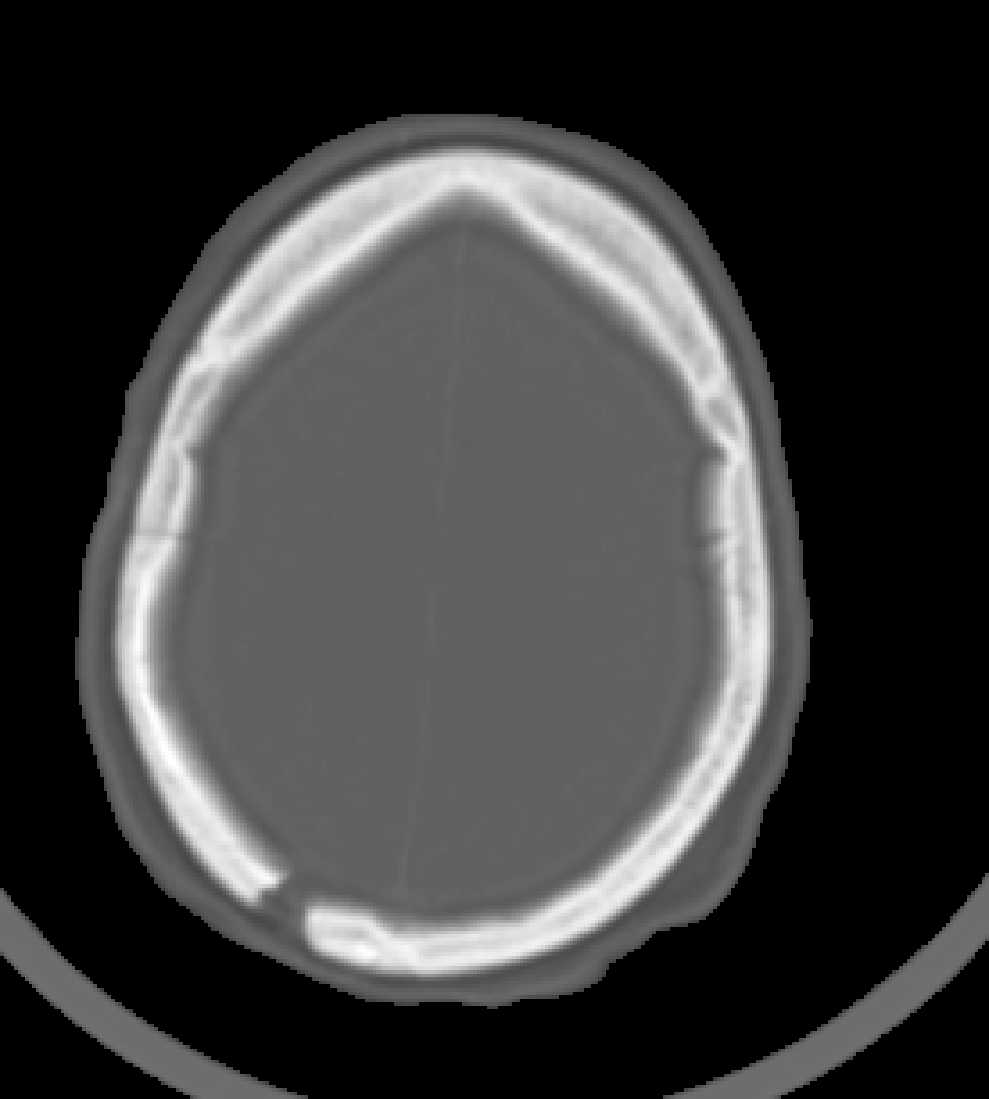}}%
\end{pgfscope}%
\begin{pgfscope}%
\pgfpathrectangle{\pgfqpoint{4.352530in}{3.898325in}}{\pgfqpoint{3.294941in}{3.665025in}}%
\pgfusepath{clip}%
\pgfsys@transformshift{4.353333in}{3.896667in}%
\pgftext[left,bottom]{\pgfimage[interpolate=true,width=3.296667in,height=3.663333in]{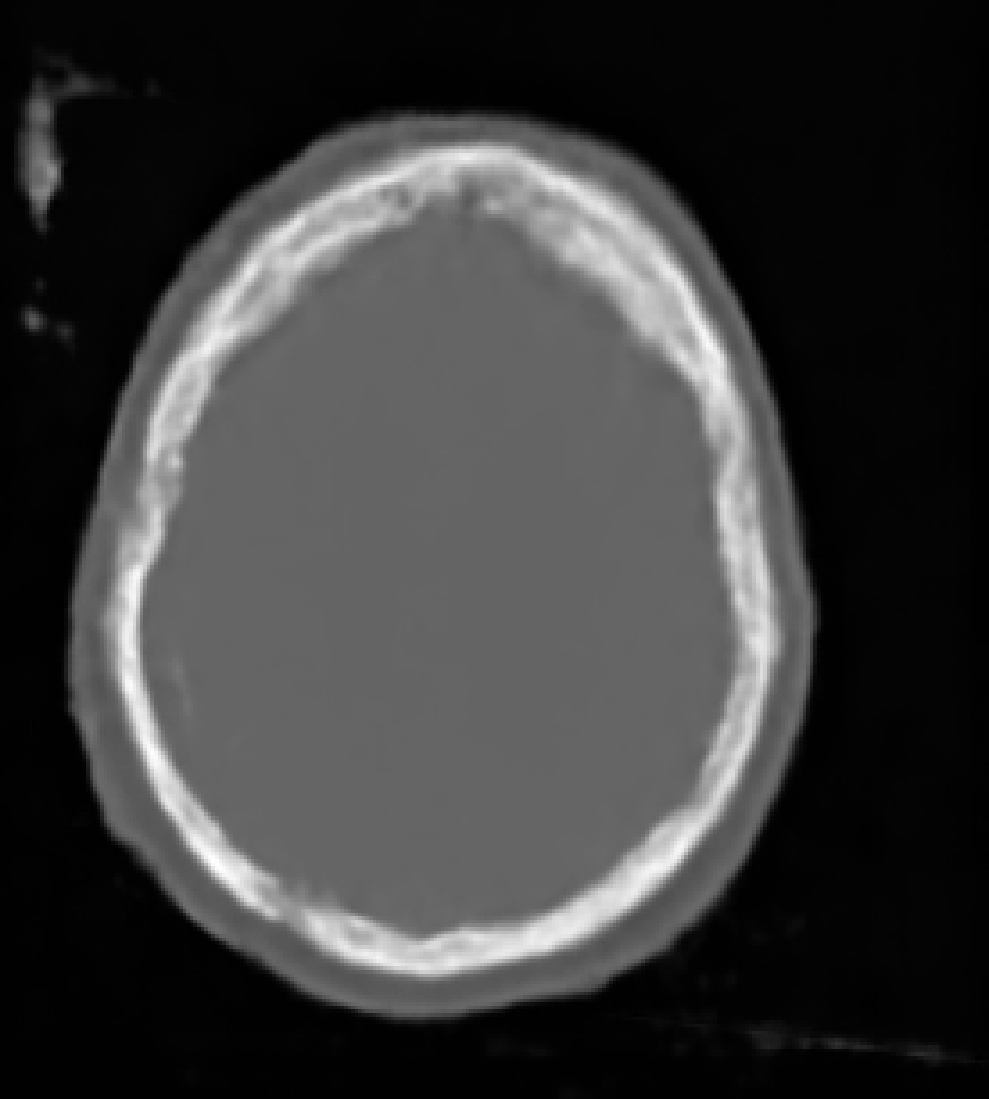}}%
\end{pgfscope}%
\begin{pgfscope}%
\pgfpathrectangle{\pgfqpoint{8.298319in}{3.898325in}}{\pgfqpoint{3.294941in}{3.665025in}}%
\pgfusepath{clip}%
\pgfsys@transformshift{8.296667in}{3.896667in}%
\pgftext[left,bottom]{\pgfimage[interpolate=true,width=3.296667in,height=3.663333in]{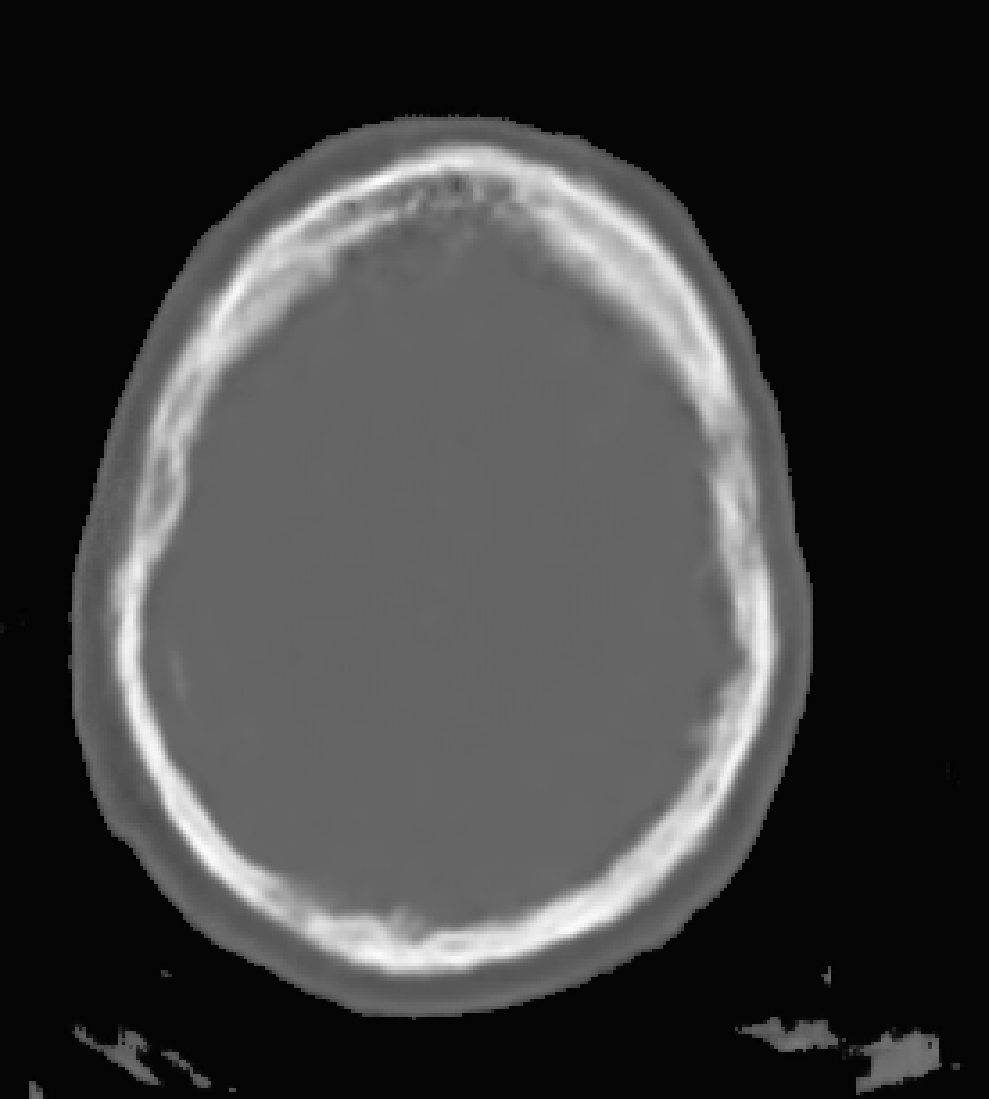}}%
\end{pgfscope}%
\begin{pgfscope}%
\pgfpathrectangle{\pgfqpoint{0.594885in}{0.160000in}}{\pgfqpoint{2.918652in}{3.665025in}}%
\pgfusepath{clip}%
\pgfsys@transformshift{0.593333in}{0.160000in}%
\pgftext[left,bottom]{\pgfimage[interpolate=true,width=2.916667in,height=3.666667in]{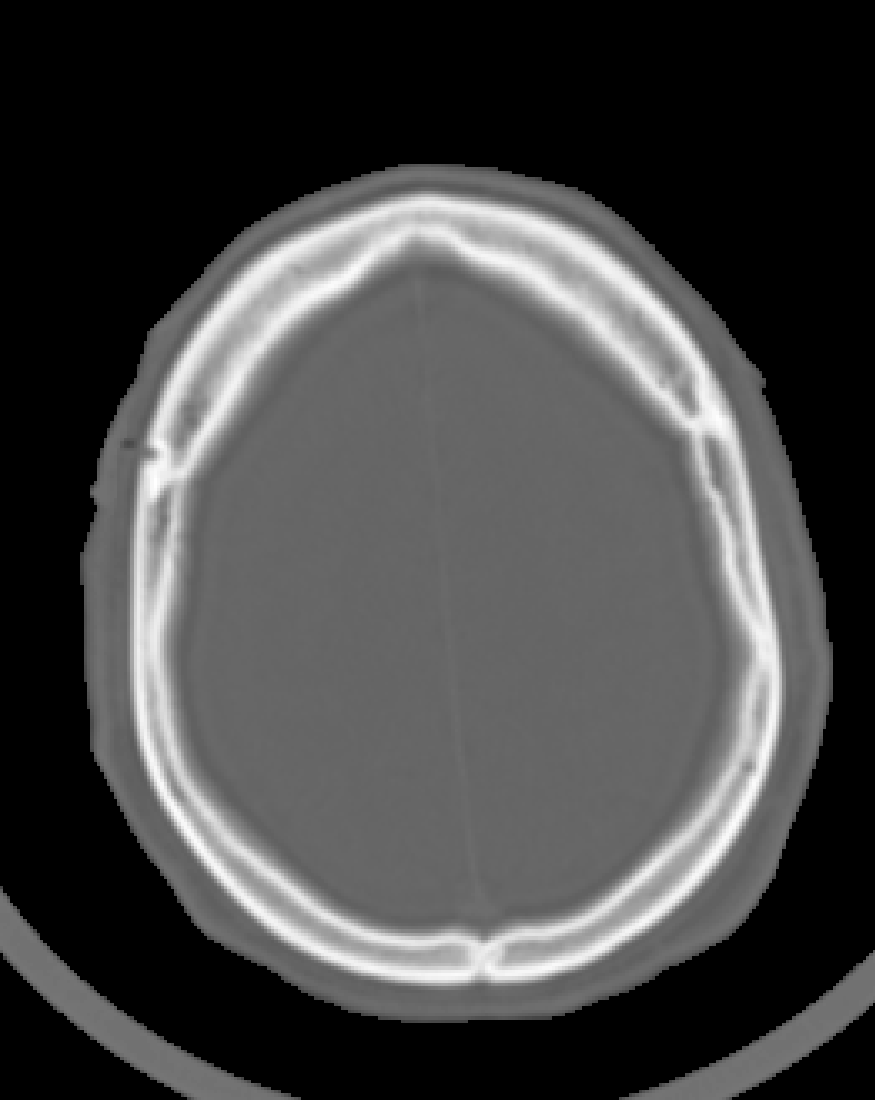}}%
\end{pgfscope}%
\begin{pgfscope}%
\pgfpathrectangle{\pgfqpoint{4.540674in}{0.160000in}}{\pgfqpoint{2.918652in}{3.665025in}}%
\pgfusepath{clip}%
\pgfsys@transformshift{4.540000in}{0.160000in}%
\pgftext[left,bottom]{\pgfimage[interpolate=true,width=2.920000in,height=3.666667in]{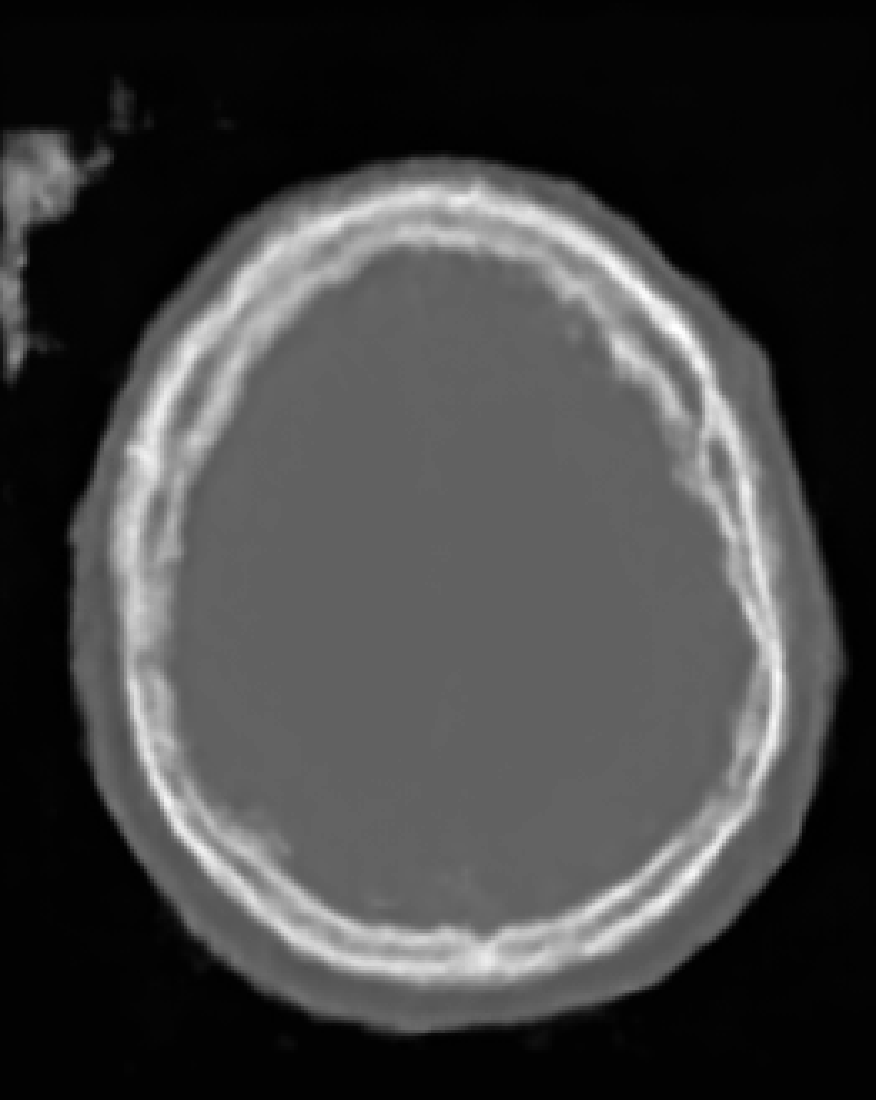}}%
\end{pgfscope}%
\begin{pgfscope}%
\pgfpathrectangle{\pgfqpoint{8.486464in}{0.160000in}}{\pgfqpoint{2.918652in}{3.665025in}}%
\pgfusepath{clip}%
\pgfsys@transformshift{8.486667in}{0.160000in}%
\pgftext[left,bottom]{\pgfimage[interpolate=true,width=2.920000in,height=3.666667in]{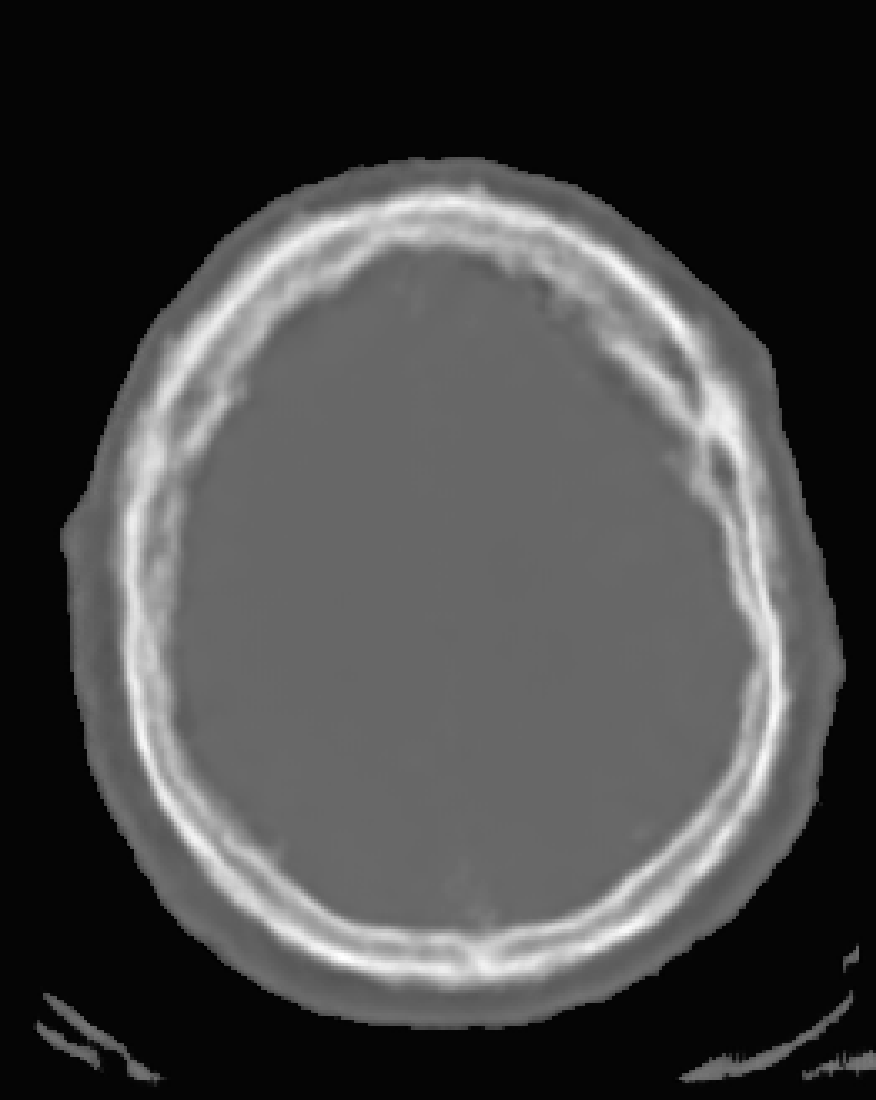}}%
\end{pgfscope}%
\end{pgfpicture}%
\makeatother%
\endgroup%
  \end{adjustbox}
  \caption{Transverse views for the u-net model trained with \acrshort{mae}
    loss compared with the pixtopix model trained with least-square adversarial
    loss, evaluated on the validation dataset.
  }\label{fig:unet_pixtopix:validation}
\end{figure}

Additionaly we manualy removed incomplete slices from the dataset. These
incomplete slices arise from the coregistration routine when one volume is
tilted but does not cover the same region of the fixed volume because of
different resolution. In \Cref{tab:unet_pixtopix:dataset} we present the
volumes shapes of the cleaned dataset. In comparison to
\Cref{tab:unet:dataset} the transverse (depth) resolution has been reduced
by the incomplete transverse slices we removed.
\begin{table}[h]
  \centering
  \begin{tabular}{ccc}
    \toprule
    Dataset & Subject & Shape \\
    \midrule
    Training & \num{1} & \num{137x320x250x1} \\
    Training & \num{2} & \num{130x328x250x1} \\
    Training & \num{3} & \num{111x303x281x1} \\
    Training & \num{4} & \num{143x291x259x1} \\
    Training & \num{5} & \num{141x307x284x1} \\
    Training & \num{6} & \num{148x278x267x1} \\
    Training & \num{7} & \num{198x289x268x1} \\
    Training & \num{8} & \num{208x282x238x1} \\
    Training & \num{9} & \num{162x322x252x1} \\
    Training & \num{10} & \num{185x303x243x1} \\
    Training & \num{11} & \num{180x317x239x1} \\
    Training & \num{12} & \num{184x310x248x1} \\
    Training & \num{13} & \num{93x313x238x1} \\
    Validation & \num{1} & \num{105x298x227x1} \\
    Validation & \num{2} & \num{190x328x282x1} \\
    Validation & \num{3} & \num{202x307x276x1} \\
    Validation & \num{4} & \num{198x329x262x1} \\
    \bottomrule
  \end{tabular}
  \caption{Training and validation dataset volumes used in this section. The
    dimensions of the shape correspond to depth, height and width.
  }\label{tab:unet_pixtopix:dataset}
\end{table}
\Cref{tab:unet_pixtopix:params} lists the training parameters used in the
following experiments. The pixtopix model required more training steps to
converge.
\begin{table}[h]
  \centering
  \begin{tabular}{cccccc}
    \toprule
    & &
    \multicolumn{2}{c}{u-net (\gls{mae})} &
    \multicolumn{2}{c}{pixtopix} \\
    Image Slices & Batch Size & Steps & Epochs & Steps & Epochs \\
    \midrule
    \num{2020} & \num{16} & \num{22080} & \num{174} & \num{37832} & \num{299} \\
    \bottomrule
  \end{tabular}
  \caption{Training parameters used for the u-net and pixtopix comparison.
  }\label{tab:unet_pixtopix:params}
\end{table}
\Cref{tab:unet_pixtopix:training} summarizes the evaluation metrics obtained
for the training dataset. Comparison to \Cref{tab:unet:training} has to be
done carefully as we used differently preprocessed datasets.
\begin{table}[h]
  \centering
  \begin{tabular}{ccccc}
    \toprule
    Model & Loss &
    \acrshort{mae} &
    \acrshort{mse} &
    \acrshort{psnr} \\
    \midrule
    u-net & \acrshort{mae} & \num{90.5} & \num{61853} & \num{49.4} \\
    pixtopix & least-square & \num{21.6} & \num{4210} & \num{60.4} \\
    \bottomrule
  \end{tabular}
  \caption{Distance metrics for the u-net model trained with \acrshort{mae}
    loss compared with the pixtopix model trained with least-square adversarial
    loss, evaluated on the training dataset.
  }\label{tab:unet_pixtopix:training}
\end{table}
\Cref{tab:unet_pixtopix:validation} lists the evaluation metrics obtained from
the validation dataset. Compared to \Cref{tab:unet_pixtopix:training} the
u-net metrics differ not as much in our former experiments with only the
u-net architecture. Furthermore we see a large decrease of the pixtopix
performance on the validation dataset.
\begin{table}[h]
  \centering
  \begin{tabular}{ccccc}
    \toprule
    Model & Loss &
    \acrshort{mae} &
    \acrshort{mse} &
    \acrshort{psnr} \\
    \midrule
    u-net & \acrshort{mae} & \num{136.9} & \num{101943} & \num{46.77} \\
    pixtopix & least-square & \num{112.7} & \num{82173} & \num{47.55} \\
    \bottomrule
  \end{tabular}
  \caption{Distance metrics for the u-net model trained with \acrshort{mae}
    loss compared with the pixtopix model trained with least-square adversarial
    loss, evaluated on the validation dataset.
  }\label{tab:unet_pixtopix:training}
\end{table}
The visual comparison of the u-net and pixtopix model on the training dataset,
see \Cref{fig:unet_pixtopix:training} shows very good results for both models.
The pixtopix model, however, does not show artifacts. Overall the hard and
soft matter tissue look very similar to the ground truth.
The visual comparison of the u-net and pixtopix model on the validation
dataset, see \Cref{fig:unet_pixtopix:validation} shows that even though the
metrics on the validation dataset decreased much more relative to the u-net
metrics, the visual results are much better. For ananatomical characteristics
general to the human head the pixtopix model is able to successfully reproduce
\gls{ct} representation from \gls{mri}, however for anatomical features that
differ greatly between subjects, results are not good.
Overall we can confirm a large improvement of the \gls{gan} approach compared
with \gls{cnn} approach. Altough both architecture use the same network for
the prediction, the adversarial loss term in pixtopix is able to guide the
optimizer to a better local extrema.

\subsection{Gradient Boost}

As a third experiment we wanted to improve the soft tissue structure of the
synthesized \gls{ct}s. Therefore we used skull extraction to create masks of
the brain volume. These masks were then used to increase gradient weight in
fine-tuning the pixtopix model.
\begin{table}[h]
  \centering
  \begin{tabular}{cccccc}
    \toprule
    & &
    \multicolumn{2}{c}{u-net} &
    \multicolumn{2}{c}{pixtopix} \\
    Image Slices & Batch Size & Steps & Epochs & Steps & Epochs \\
    \midrule
    \num{2020} & \num{16} & \num{37832} & \num{299} & \num{51911} & \num{411} \\
    \bottomrule
  \end{tabular}
  \caption{Training parameters used for pixtopix and gradient boosted
    fine-tuned pixtopix model.
  }\label{tab:pixtopix:params}
\end{table}
In \Cref{tab:pixtopix:params} we summarized the training parameters for the
fine-tuning experiment. The untuned pixtopix, described in the prevous section,
was trained for about \num{299} epochs. Then we amended the gradient
calculation to increase weight of the soft-tissue area and proceeded to train
for about \num{112} more epochs.
\begin{table}[h]
  \centering
  \begin{tabular}{cccc}
    \toprule
    Model & \acrshort{mae} & \acrshort{mse} & \acrshort{psnr} \\
    \midrule
    pixtopix & \num{21.56} & \num{4210} & \num{60.38} \\
    pixtopix (fine-tuned) & \num{23.37} & \num{4140} & \num{60.30} \\
    \bottomrule
  \end{tabular}
  \caption{Distance metrics for the pixtopix model trained with least-square
    adversarial loss compared to fine-tuned with gradient boost, evaluated on
    the training dataset.
  }\label{tab:pixtopix:training}
\end{table}
\begin{table}[h]
  \centering
  \begin{tabular}{cccc}
    \toprule
    Model & \acrshort{mae} & \acrshort{mse} & \acrshort{psnr} \\
    \midrule
    pixtopix & \num{112.73} & \num{82173} & \num{47.55} \\
    pixtopix (fine-tuned) & \num{112.23} & \num{79296} & \num{47.68} \\
    \bottomrule
  \end{tabular}
  \caption{Distance metrics for the pixtopix model trained with least-square
    adversarial loss compared to fine-tuned with gradient boost, evaluated on
    the validation dataset.
  }\label{tab:pixtopix:validation}
\end{table}
In \Cref{tab:pixtopix:training} and \Cref{tab:pixtopix:validation} the
evaluation metrics on the training and validation dataset comparing the
pixtopix model with the fine-tuned pixtopix model are presented. For the
\gls{mae} and \gls{mse} we see a small improvement of the fine-tuned model
on the validation dataset.
\begin{figure}[h]
  \centering
  \begin{adjustbox}{width=\linewidth}
\begingroup%
\makeatletter%
\begin{pgfpicture}%
\pgfpathrectangle{\pgfpointorigin}{\pgfqpoint{12.000000in}{16.000000in}}%
\pgfusepath{use as bounding box, clip}%
\begin{pgfscope}%
\pgfsetbuttcap%
\pgfsetmiterjoin%
\pgfsetlinewidth{0.000000pt}%
\definecolor{currentstroke}{rgb}{1.000000,1.000000,1.000000}%
\pgfsetstrokecolor{currentstroke}%
\pgfsetdash{}{0pt}%
\pgfpathmoveto{\pgfqpoint{0.000000in}{0.000000in}}%
\pgfpathlineto{\pgfqpoint{12.000000in}{0.000000in}}%
\pgfpathlineto{\pgfqpoint{12.000000in}{16.000000in}}%
\pgfpathlineto{\pgfqpoint{0.000000in}{16.000000in}}%
\pgfpathclose%
\pgfusepath{}%
\end{pgfscope}%
\begin{pgfscope}%
\pgfpathrectangle{\pgfqpoint{0.622560in}{11.374975in}}{\pgfqpoint{2.863300in}{3.665025in}}%
\pgfusepath{clip}%
\pgfsys@transformshift{0.623333in}{11.373333in}%
\pgftext[left,bottom]{\pgfimage[interpolate=true,width=2.860000in,height=3.666667in]{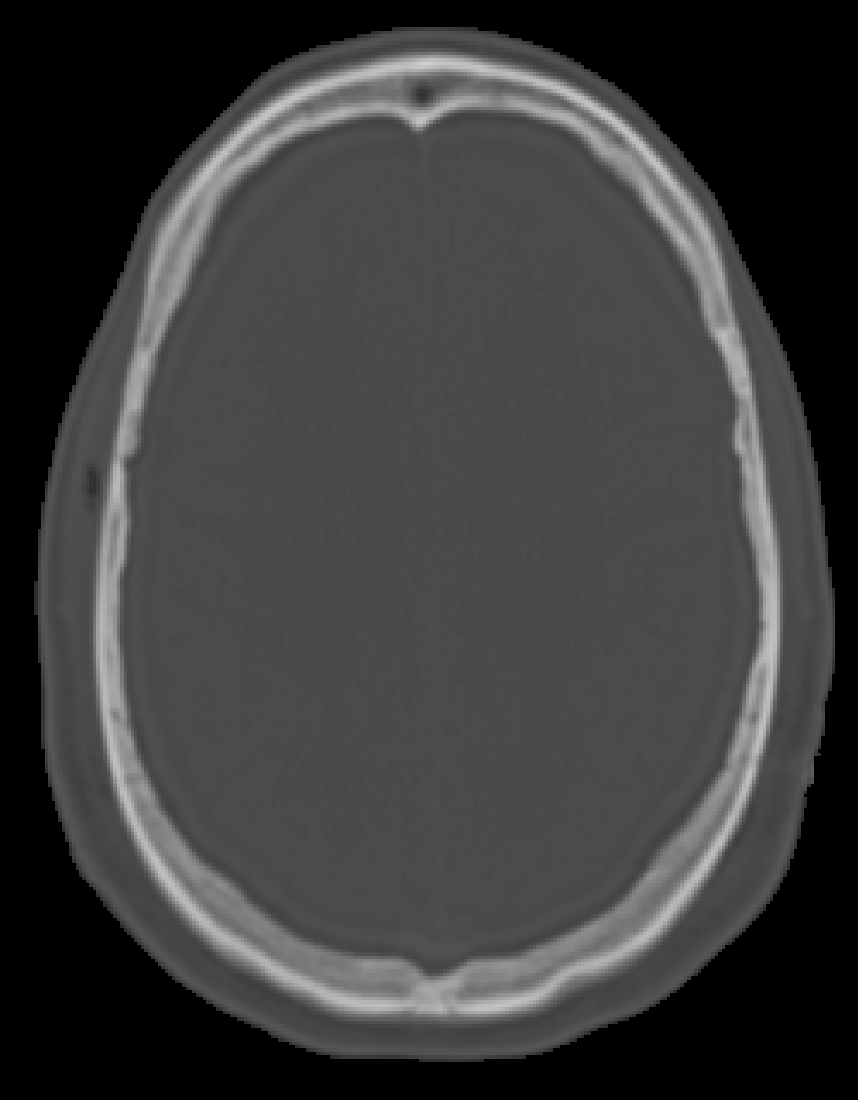}}%
\end{pgfscope}%
\begin{pgfscope}%
\pgftext[x=2.054211in,y=15.317778in,,base]{\fontsize{16.000000}{19.200000}\selectfont Ground Truth}%
\end{pgfscope}%
\begin{pgfscope}%
\pgfpathrectangle{\pgfqpoint{4.568350in}{11.374975in}}{\pgfqpoint{2.863300in}{3.665025in}}%
\pgfusepath{clip}%
\pgfsys@transformshift{4.570000in}{11.373333in}%
\pgftext[left,bottom]{\pgfimage[interpolate=true,width=2.863333in,height=3.666667in]{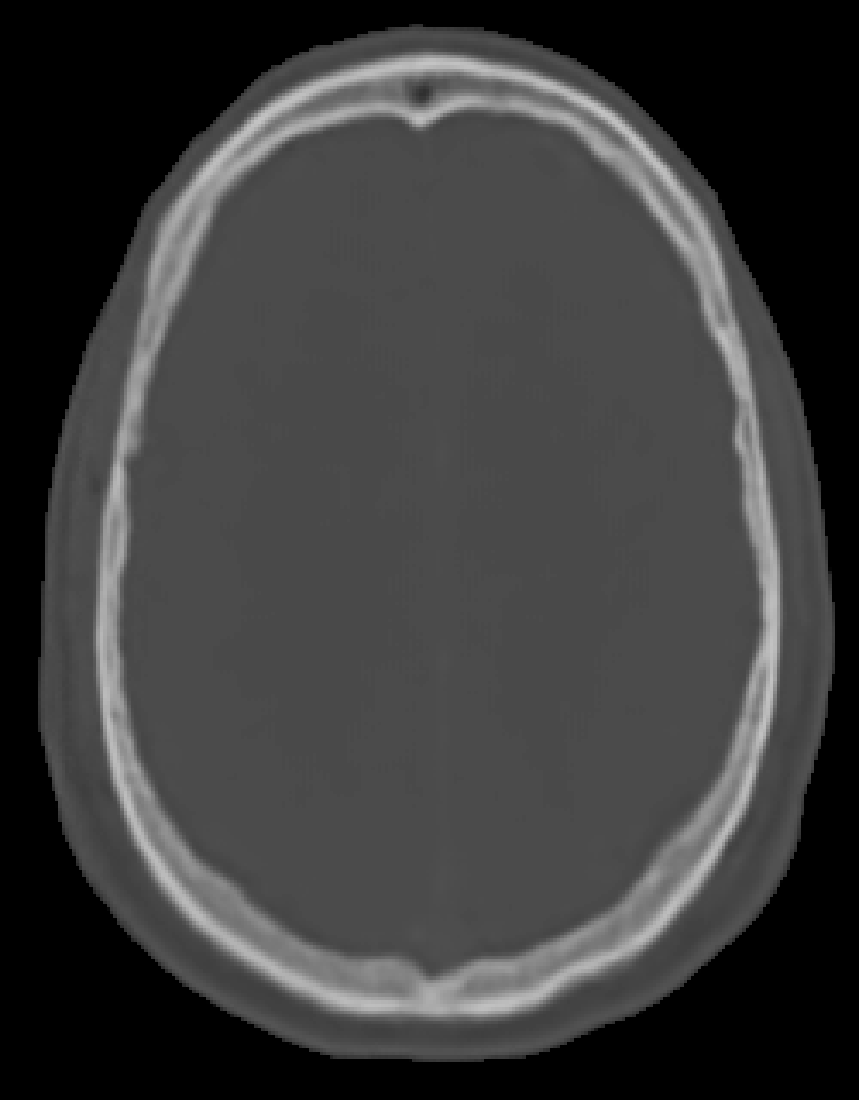}}%
\end{pgfscope}%
\begin{pgfscope}%
\pgftext[x=6.000000in,y=15.317778in,,base]{\fontsize{16.000000}{19.200000}\selectfont pixtopix}%
\end{pgfscope}%
\begin{pgfscope}%
\pgfpathrectangle{\pgfqpoint{8.514139in}{11.374975in}}{\pgfqpoint{2.863300in}{3.665025in}}%
\pgfusepath{clip}%
\pgfsys@transformshift{8.513333in}{11.373333in}%
\pgftext[left,bottom]{\pgfimage[interpolate=true,width=2.863333in,height=3.666667in]{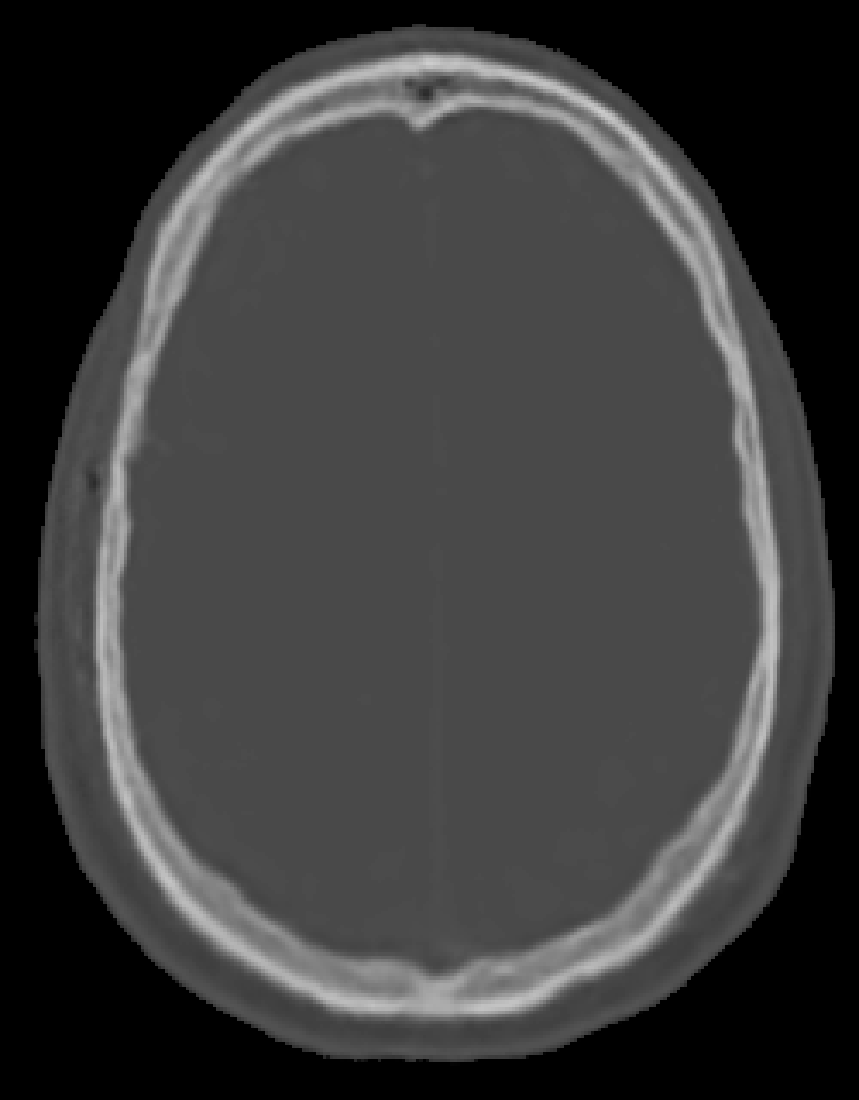}}%
\end{pgfscope}%
\begin{pgfscope}%
\pgftext[x=9.945789in,y=15.317778in,,base]{\fontsize{16.000000}{19.200000}\selectfont pixtopix (boosted)}%
\end{pgfscope}%
\begin{pgfscope}%
\pgfpathrectangle{\pgfqpoint{0.657479in}{7.636650in}}{\pgfqpoint{2.793464in}{3.665025in}}%
\pgfusepath{clip}%
\pgfsys@transformshift{0.656667in}{7.636667in}%
\pgftext[left,bottom]{\pgfimage[interpolate=true,width=2.796667in,height=3.666667in]{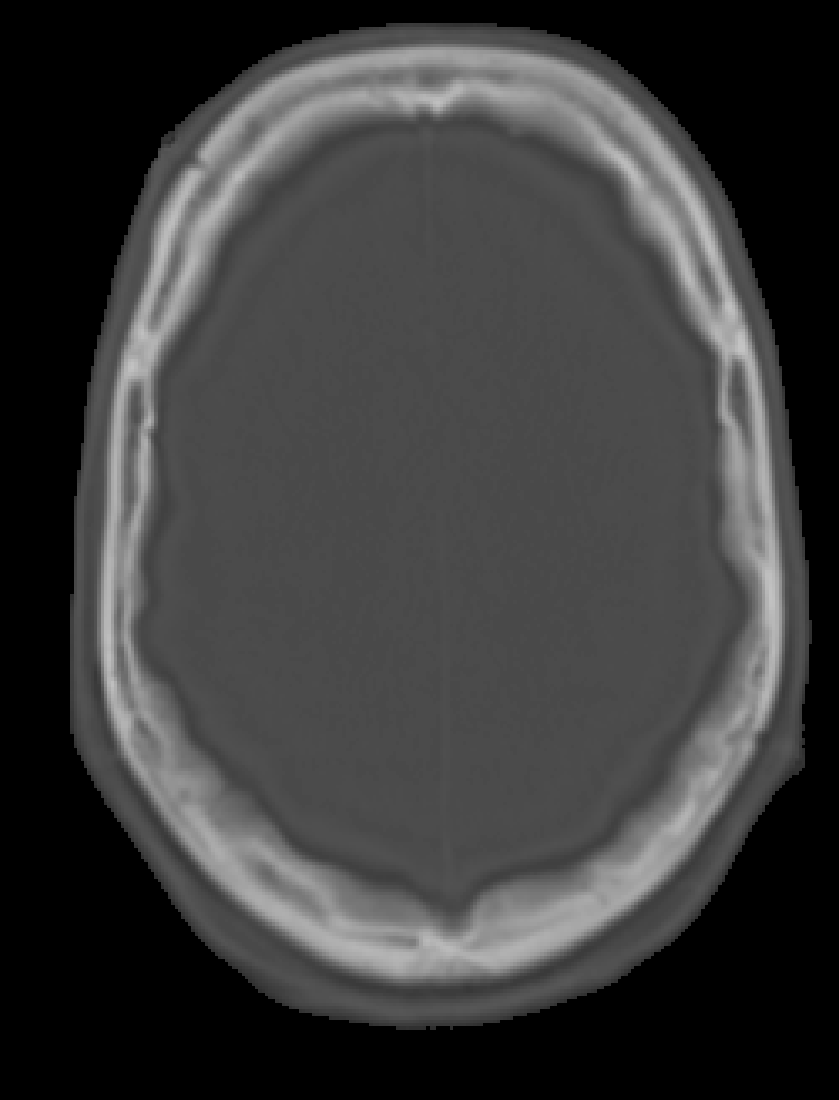}}%
\end{pgfscope}%
\begin{pgfscope}%
\pgfpathrectangle{\pgfqpoint{4.603268in}{7.636650in}}{\pgfqpoint{2.793464in}{3.665025in}}%
\pgfusepath{clip}%
\pgfsys@transformshift{4.603333in}{7.636667in}%
\pgftext[left,bottom]{\pgfimage[interpolate=true,width=2.796667in,height=3.666667in]{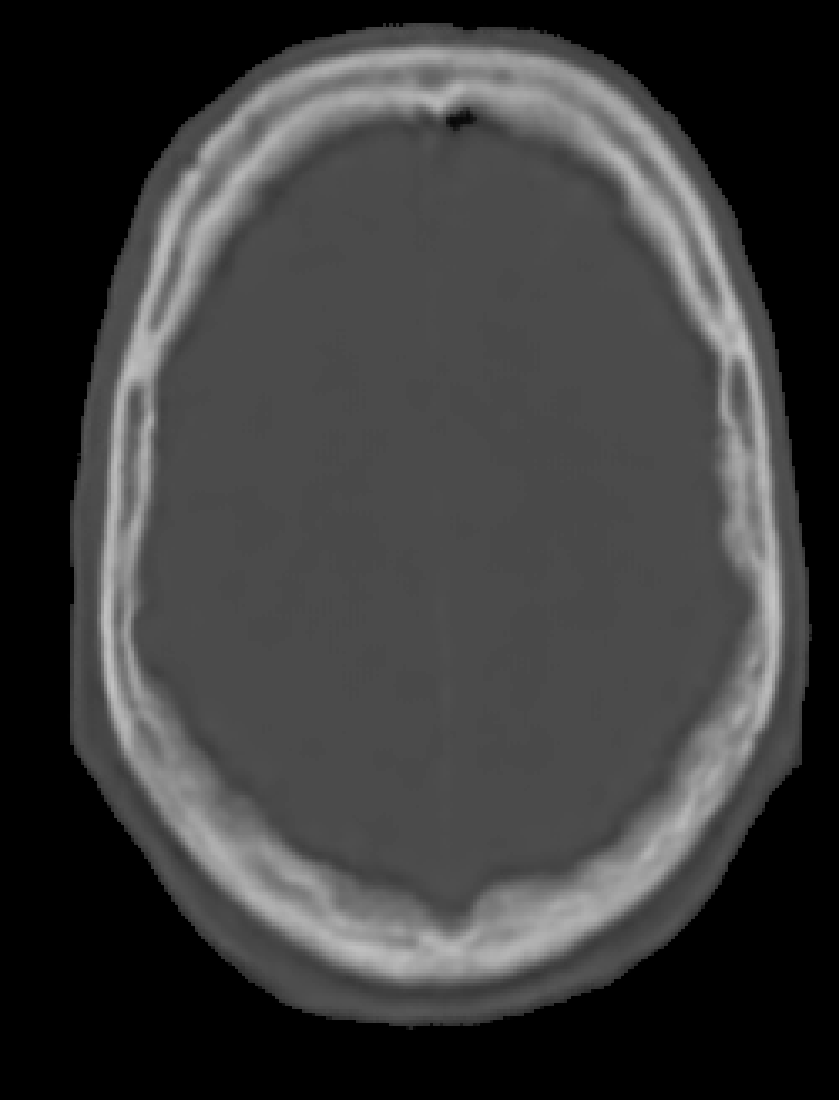}}%
\end{pgfscope}%
\begin{pgfscope}%
\pgfpathrectangle{\pgfqpoint{8.549058in}{7.636650in}}{\pgfqpoint{2.793464in}{3.665025in}}%
\pgfusepath{clip}%
\pgfsys@transformshift{8.550000in}{7.636667in}%
\pgftext[left,bottom]{\pgfimage[interpolate=true,width=2.796667in,height=3.666667in]{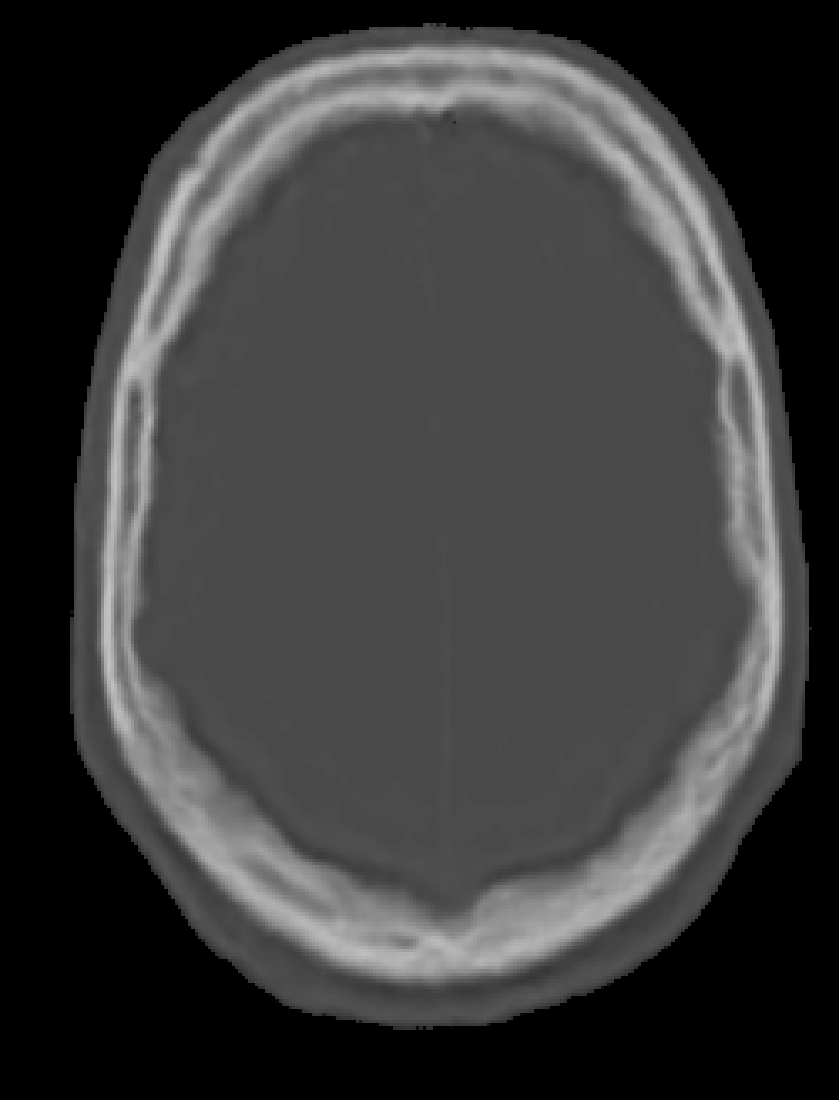}}%
\end{pgfscope}%
\begin{pgfscope}%
\pgfpathrectangle{\pgfqpoint{0.354752in}{3.898325in}}{\pgfqpoint{3.398917in}{3.665025in}}%
\pgfusepath{clip}%
\pgfsys@transformshift{0.353333in}{3.896667in}%
\pgftext[left,bottom]{\pgfimage[interpolate=true,width=3.400000in,height=3.663333in]{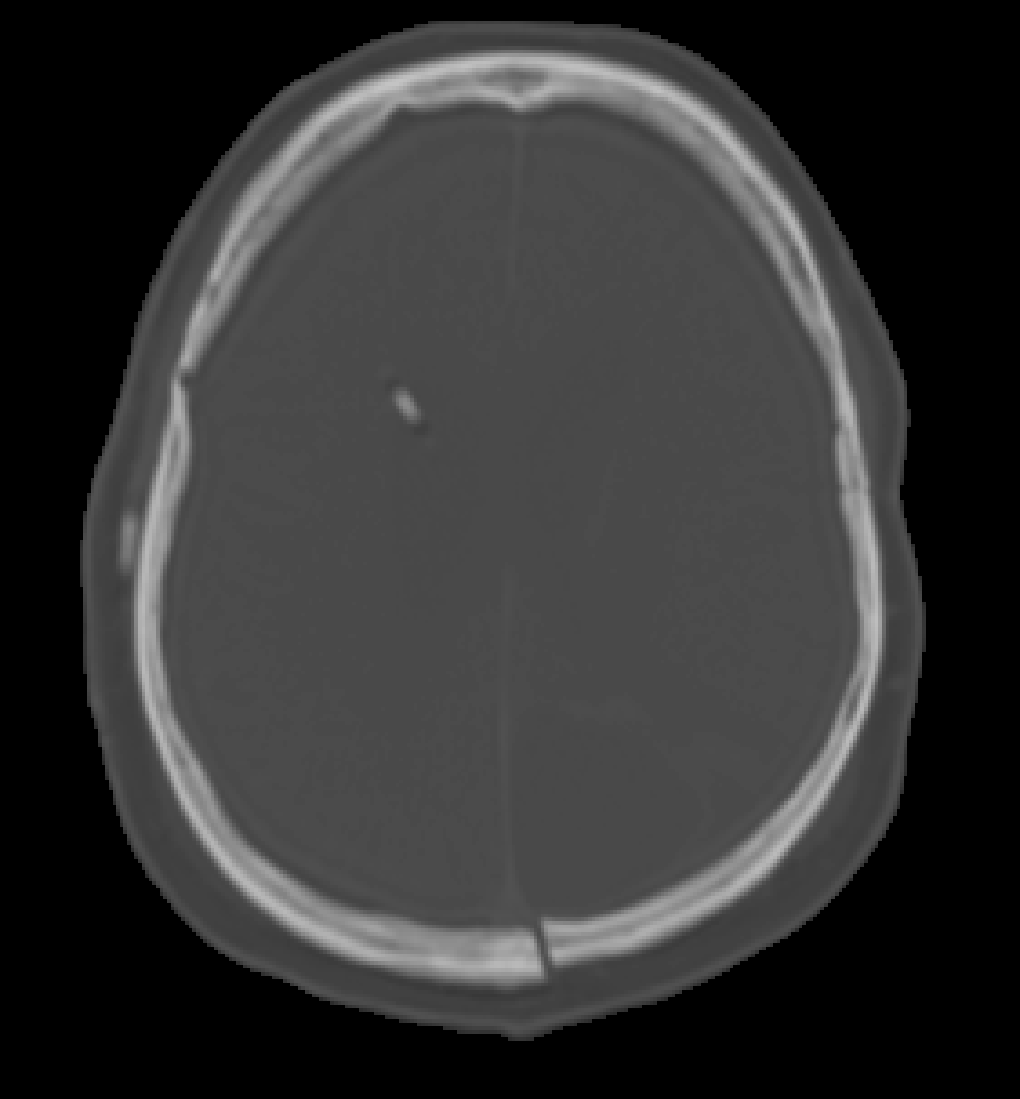}}%
\end{pgfscope}%
\begin{pgfscope}%
\pgfpathrectangle{\pgfqpoint{4.300541in}{3.898325in}}{\pgfqpoint{3.398917in}{3.665025in}}%
\pgfusepath{clip}%
\pgfsys@transformshift{4.300000in}{3.896667in}%
\pgftext[left,bottom]{\pgfimage[interpolate=true,width=3.400000in,height=3.663333in]{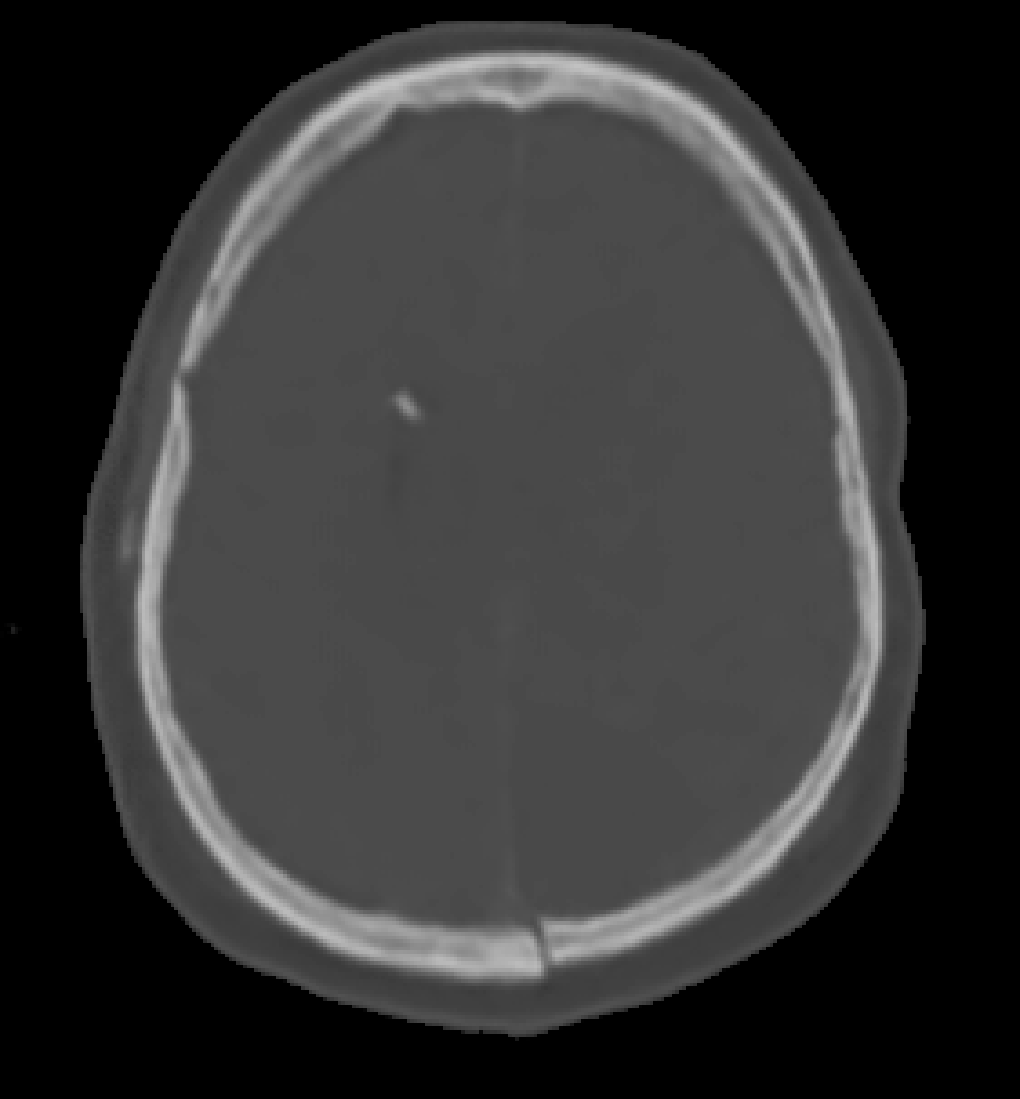}}%
\end{pgfscope}%
\begin{pgfscope}%
\pgfpathrectangle{\pgfqpoint{8.246331in}{3.898325in}}{\pgfqpoint{3.398917in}{3.665025in}}%
\pgfusepath{clip}%
\pgfsys@transformshift{8.246667in}{3.896667in}%
\pgftext[left,bottom]{\pgfimage[interpolate=true,width=3.400000in,height=3.663333in]{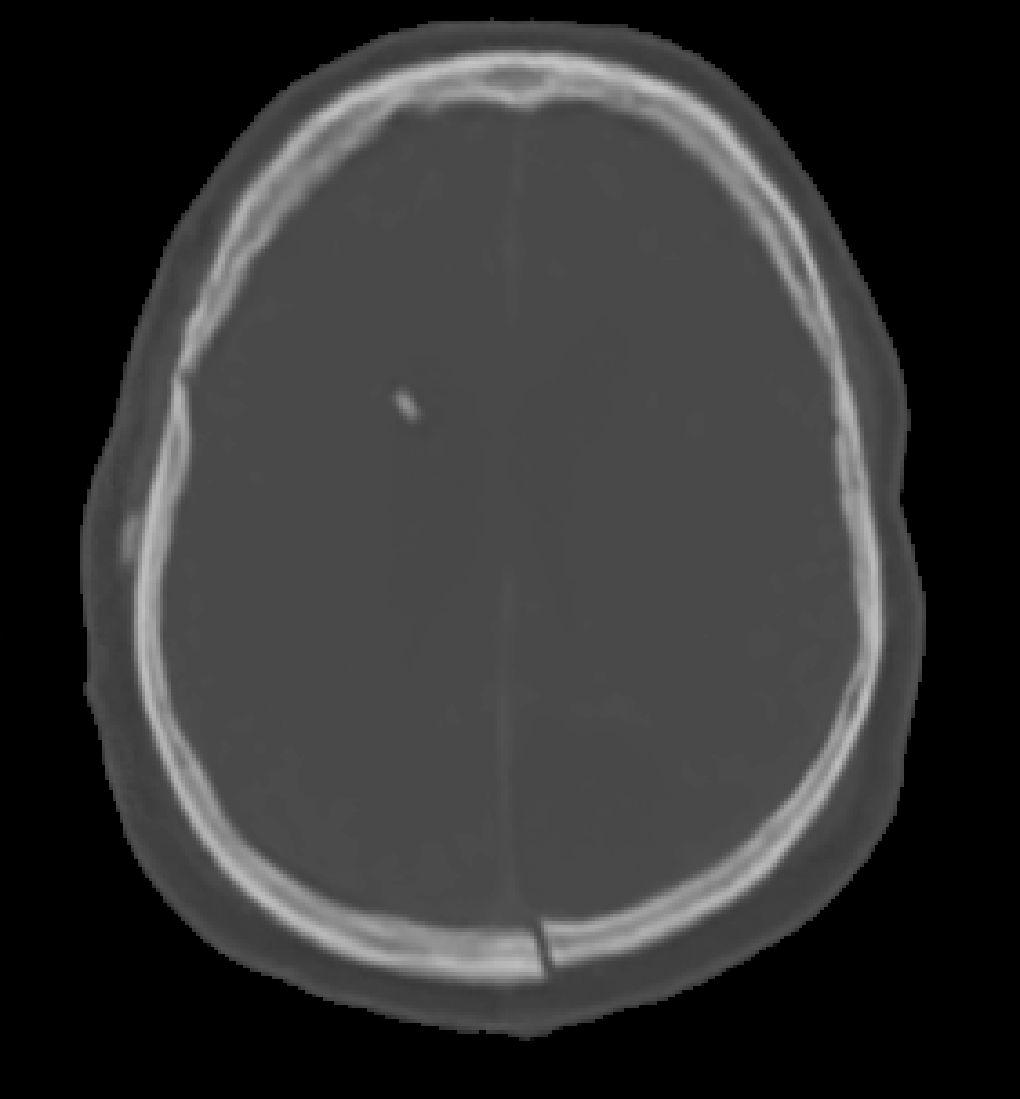}}%
\end{pgfscope}%
\begin{pgfscope}%
\pgfpathrectangle{\pgfqpoint{0.423212in}{0.160000in}}{\pgfqpoint{3.261998in}{3.665025in}}%
\pgfusepath{clip}%
\pgfsys@transformshift{0.423333in}{0.160000in}%
\pgftext[left,bottom]{\pgfimage[interpolate=true,width=3.263333in,height=3.666667in]{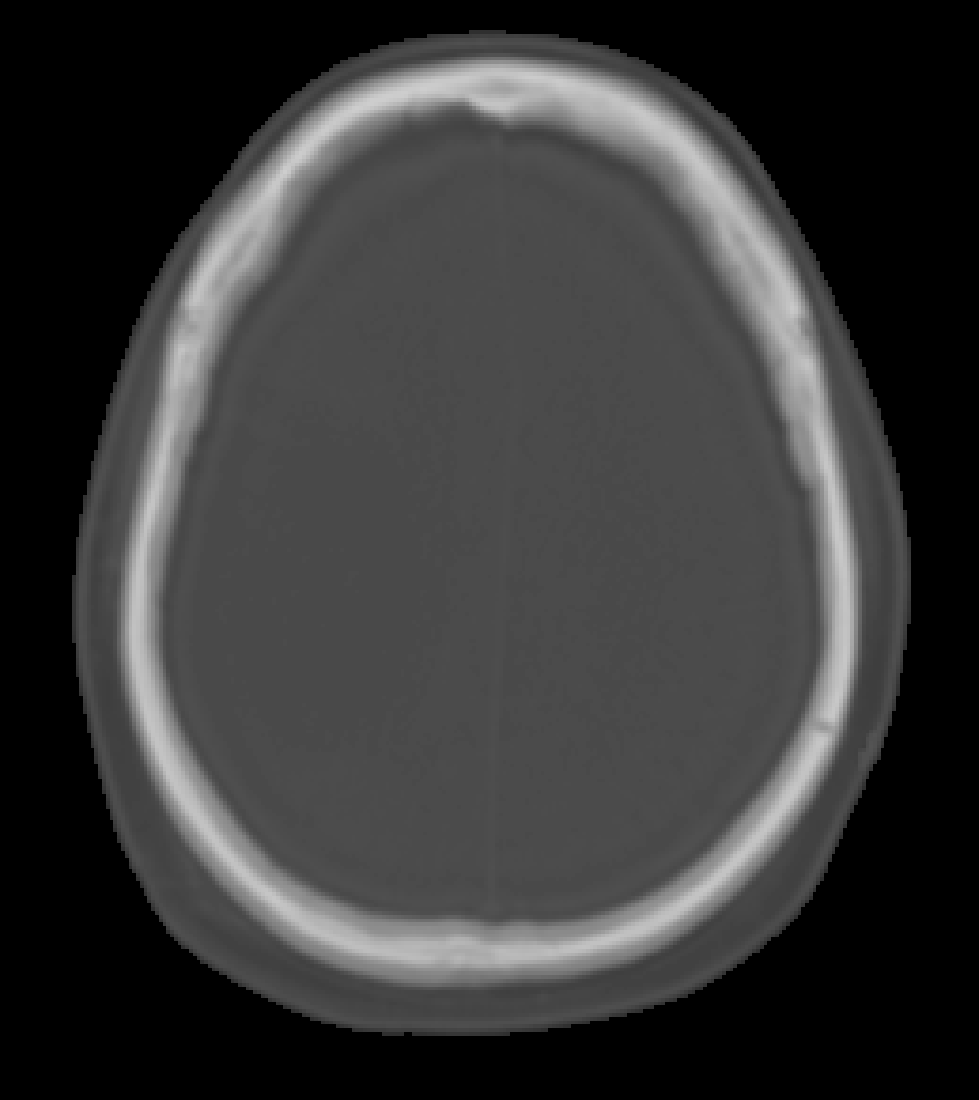}}%
\end{pgfscope}%
\begin{pgfscope}%
\pgfpathrectangle{\pgfqpoint{4.369001in}{0.160000in}}{\pgfqpoint{3.261998in}{3.665025in}}%
\pgfusepath{clip}%
\pgfsys@transformshift{4.370000in}{0.160000in}%
\pgftext[left,bottom]{\pgfimage[interpolate=true,width=3.260000in,height=3.666667in]{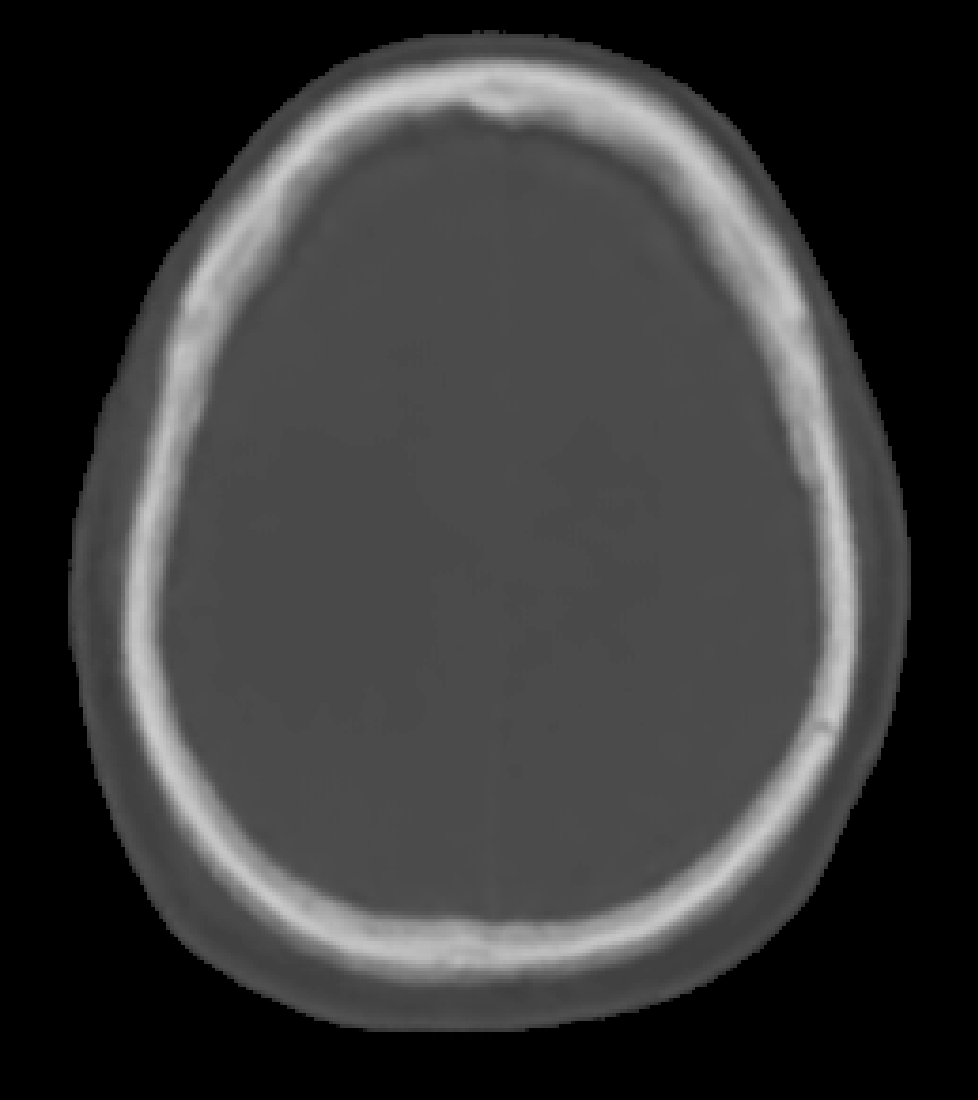}}%
\end{pgfscope}%
\begin{pgfscope}%
\pgfpathrectangle{\pgfqpoint{8.314791in}{0.160000in}}{\pgfqpoint{3.261998in}{3.665025in}}%
\pgfusepath{clip}%
\pgfsys@transformshift{8.313333in}{0.160000in}%
\pgftext[left,bottom]{\pgfimage[interpolate=true,width=3.263333in,height=3.666667in]{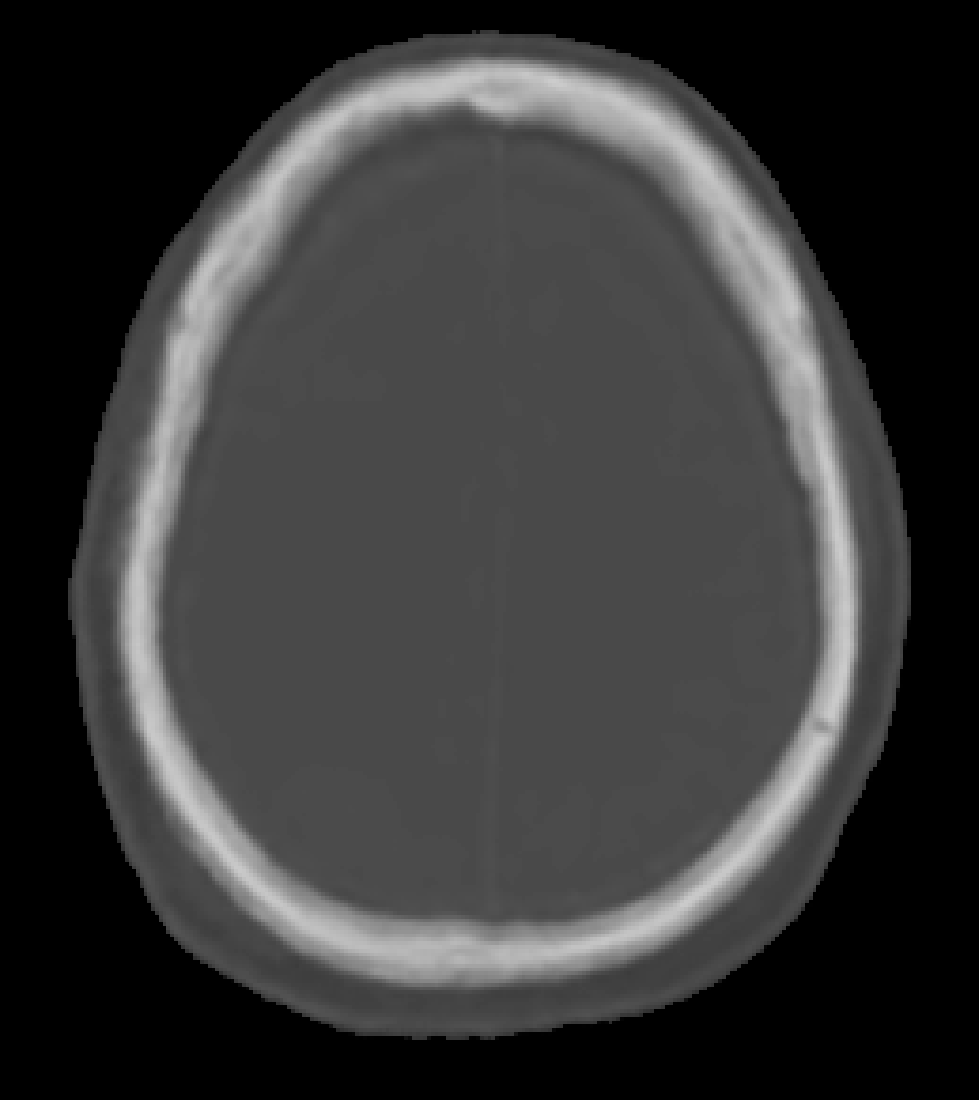}}%
\end{pgfscope}%
\end{pgfpicture}%
\makeatother%
\endgroup%
  \end{adjustbox}
  \caption{Transverse views for the pixtopix model trained with least-square
    adversarial loss compared to fine-tuned with gradient boost, evaluated on
    the validation dataset.
  }\label{fig:pixtopix:training}
\end{figure}
\begin{figure}[h]
  \centering
  \begin{adjustbox}{width=\linewidth}
\begingroup%
\makeatletter%
\begin{pgfpicture}%
\pgfpathrectangle{\pgfpointorigin}{\pgfqpoint{12.000000in}{16.000000in}}%
\pgfusepath{use as bounding box, clip}%
\begin{pgfscope}%
\pgfsetbuttcap%
\pgfsetmiterjoin%
\pgfsetlinewidth{0.000000pt}%
\definecolor{currentstroke}{rgb}{1.000000,1.000000,1.000000}%
\pgfsetstrokecolor{currentstroke}%
\pgfsetdash{}{0pt}%
\pgfpathmoveto{\pgfqpoint{0.000000in}{0.000000in}}%
\pgfpathlineto{\pgfqpoint{12.000000in}{0.000000in}}%
\pgfpathlineto{\pgfqpoint{12.000000in}{16.000000in}}%
\pgfpathlineto{\pgfqpoint{0.000000in}{16.000000in}}%
\pgfpathclose%
\pgfusepath{}%
\end{pgfscope}%
\begin{pgfscope}%
\pgfpathrectangle{\pgfqpoint{0.658303in}{11.374975in}}{\pgfqpoint{2.791814in}{3.665025in}}%
\pgfusepath{clip}%
\pgfsys@transformshift{0.656667in}{11.373333in}%
\pgftext[left,bottom]{\pgfimage[interpolate=true,width=2.793333in,height=3.666667in]{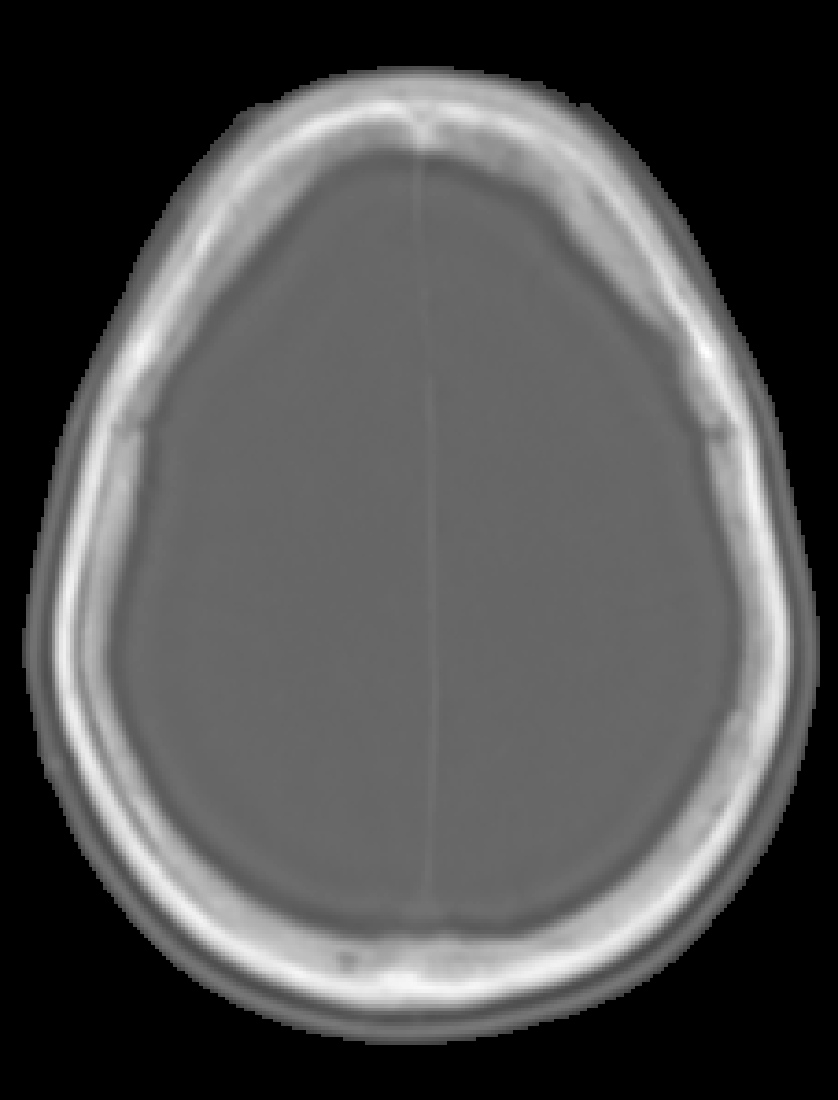}}%
\end{pgfscope}%
\begin{pgfscope}%
\pgftext[x=2.054211in,y=15.317778in,,base]{\fontsize{16.000000}{19.200000}\selectfont Ground Truth}%
\end{pgfscope}%
\begin{pgfscope}%
\pgfpathrectangle{\pgfqpoint{4.604093in}{11.374975in}}{\pgfqpoint{2.791814in}{3.665025in}}%
\pgfusepath{clip}%
\pgfsys@transformshift{4.603333in}{11.373333in}%
\pgftext[left,bottom]{\pgfimage[interpolate=true,width=2.793333in,height=3.666667in]{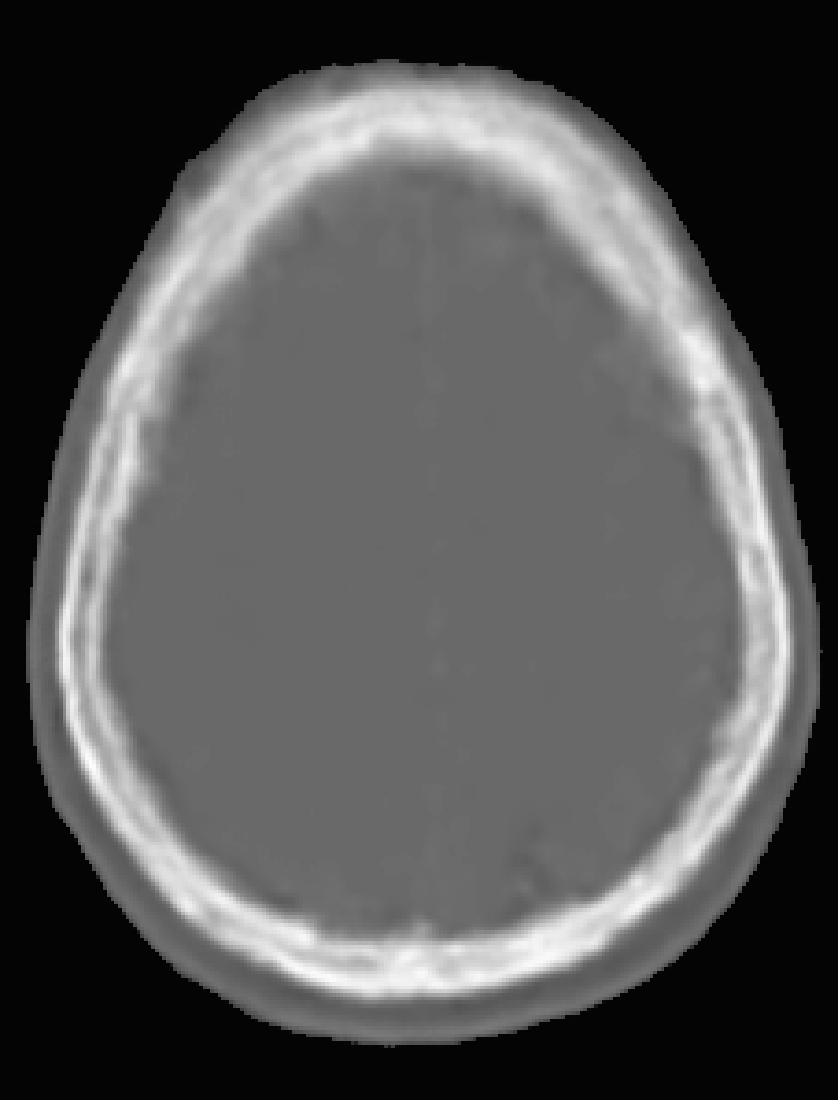}}%
\end{pgfscope}%
\begin{pgfscope}%
\pgftext[x=6.000000in,y=15.317778in,,base]{\fontsize{16.000000}{19.200000}\selectfont pixtopix}%
\end{pgfscope}%
\begin{pgfscope}%
\pgfpathrectangle{\pgfqpoint{8.549882in}{11.374975in}}{\pgfqpoint{2.791814in}{3.665025in}}%
\pgfusepath{clip}%
\pgfsys@transformshift{8.550000in}{11.373333in}%
\pgftext[left,bottom]{\pgfimage[interpolate=true,width=2.793333in,height=3.666667in]{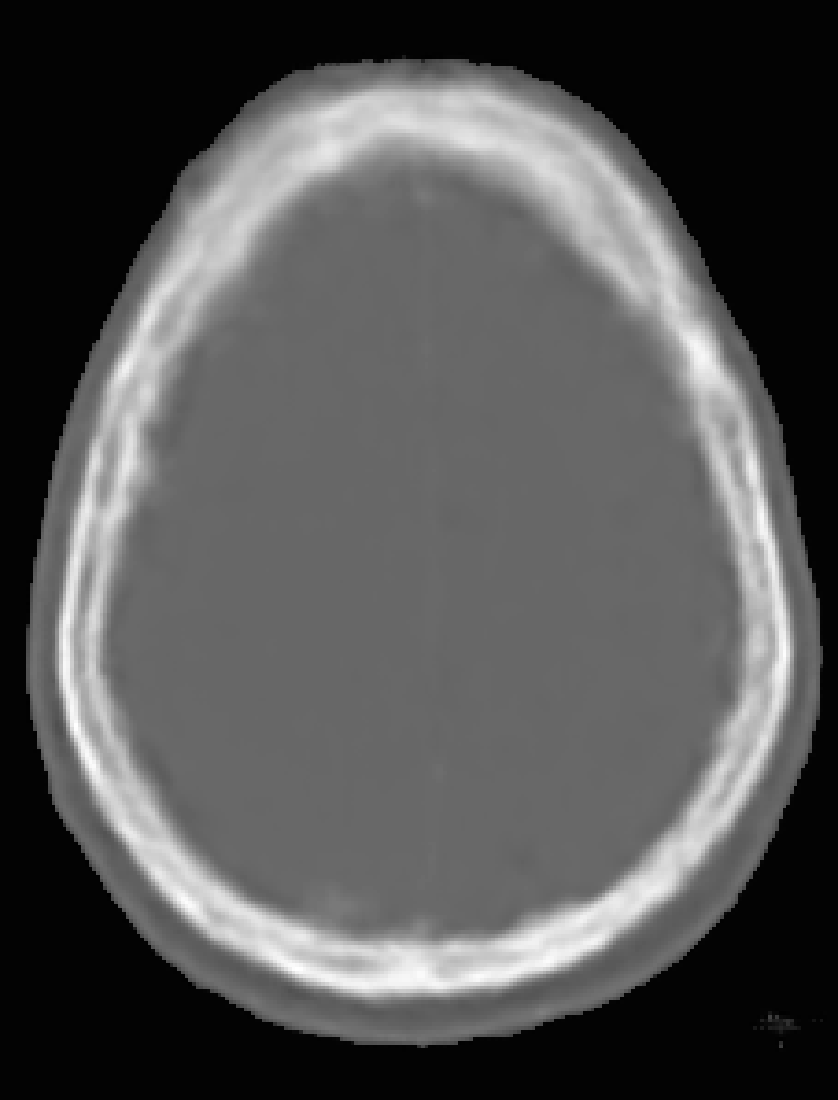}}%
\end{pgfscope}%
\begin{pgfscope}%
\pgftext[x=9.945789in,y=15.317778in,,base]{\fontsize{16.000000}{19.200000}\selectfont pixtopix (boosted)}%
\end{pgfscope}%
\begin{pgfscope}%
\pgfpathrectangle{\pgfqpoint{0.478697in}{7.636650in}}{\pgfqpoint{3.151027in}{3.665025in}}%
\pgfusepath{clip}%
\pgfsys@transformshift{0.480000in}{7.636667in}%
\pgftext[left,bottom]{\pgfimage[interpolate=true,width=3.153333in,height=3.666667in]{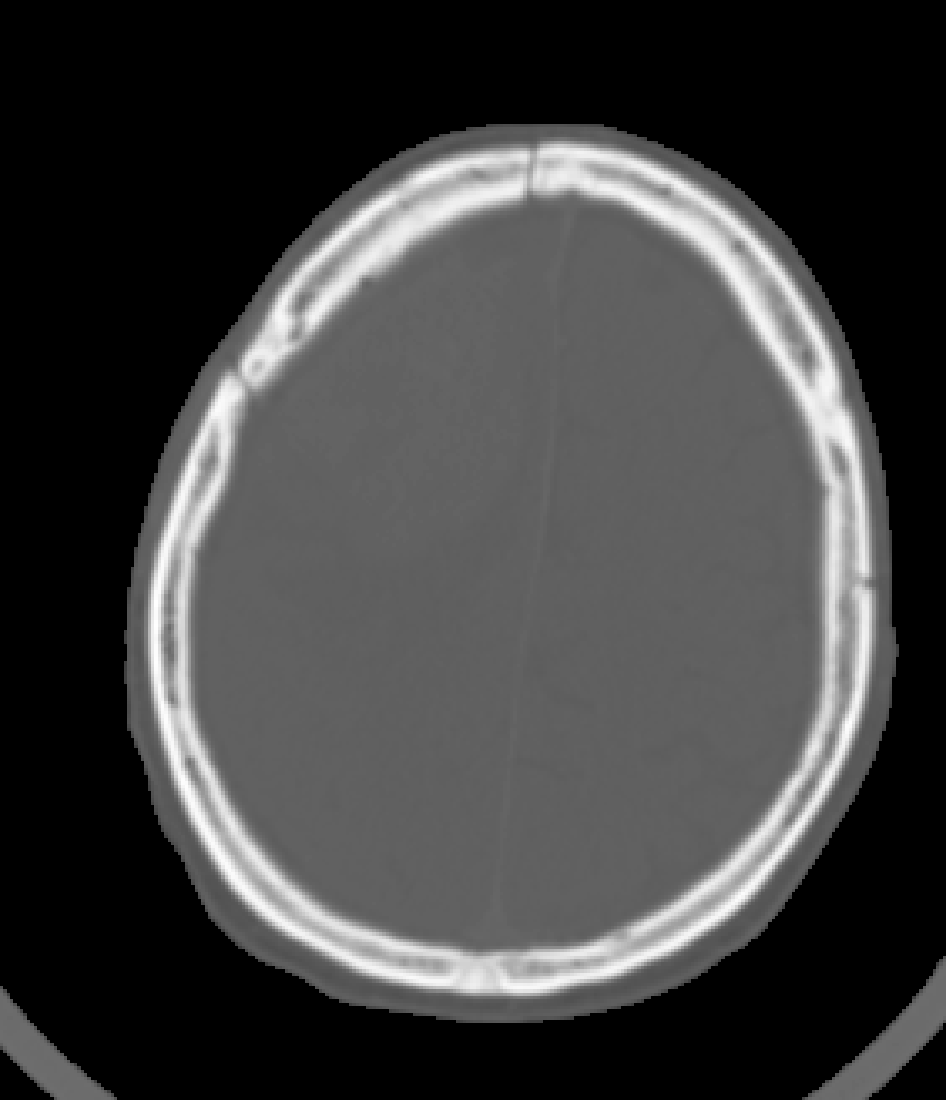}}%
\end{pgfscope}%
\begin{pgfscope}%
\pgfpathrectangle{\pgfqpoint{4.424486in}{7.636650in}}{\pgfqpoint{3.151027in}{3.665025in}}%
\pgfusepath{clip}%
\pgfsys@transformshift{4.423333in}{7.636667in}%
\pgftext[left,bottom]{\pgfimage[interpolate=true,width=3.153333in,height=3.666667in]{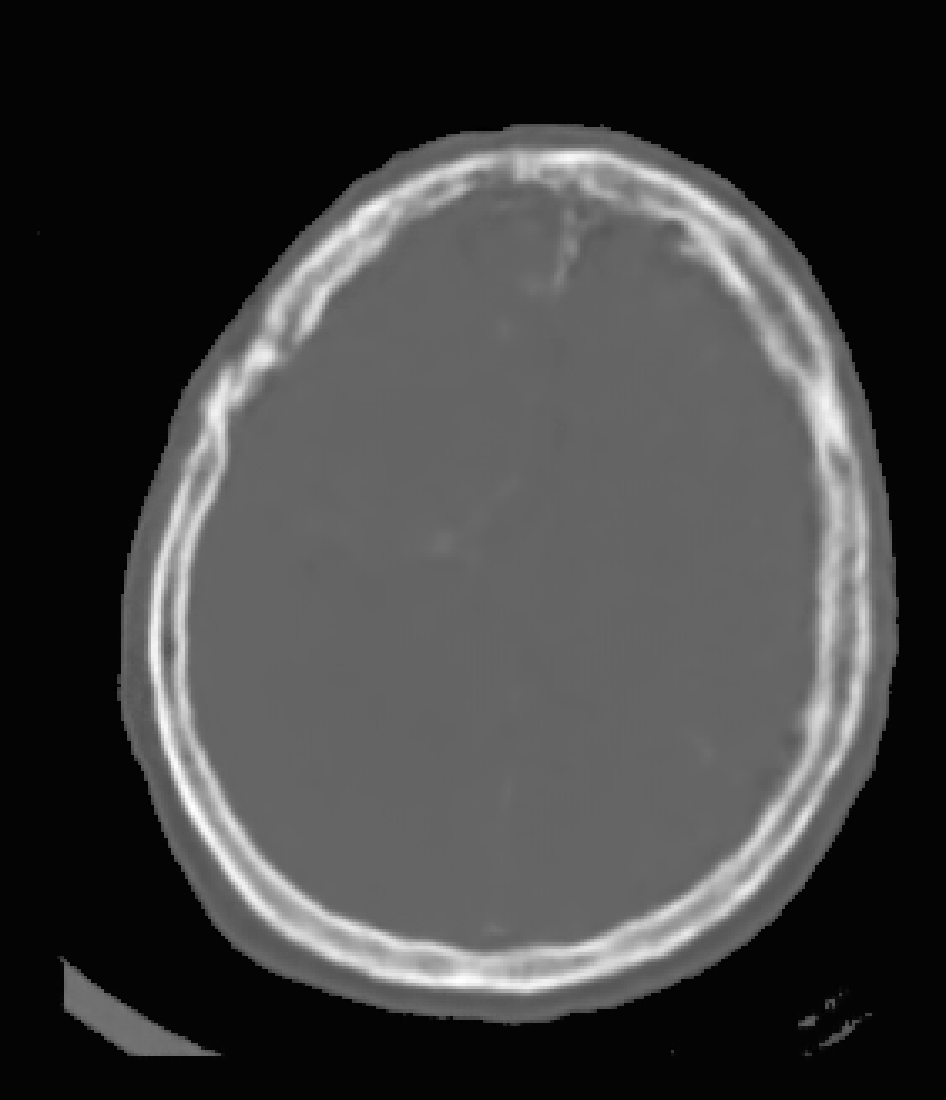}}%
\end{pgfscope}%
\begin{pgfscope}%
\pgfpathrectangle{\pgfqpoint{8.370276in}{7.636650in}}{\pgfqpoint{3.151027in}{3.665025in}}%
\pgfusepath{clip}%
\pgfsys@transformshift{8.370000in}{7.636667in}%
\pgftext[left,bottom]{\pgfimage[interpolate=true,width=3.153333in,height=3.666667in]{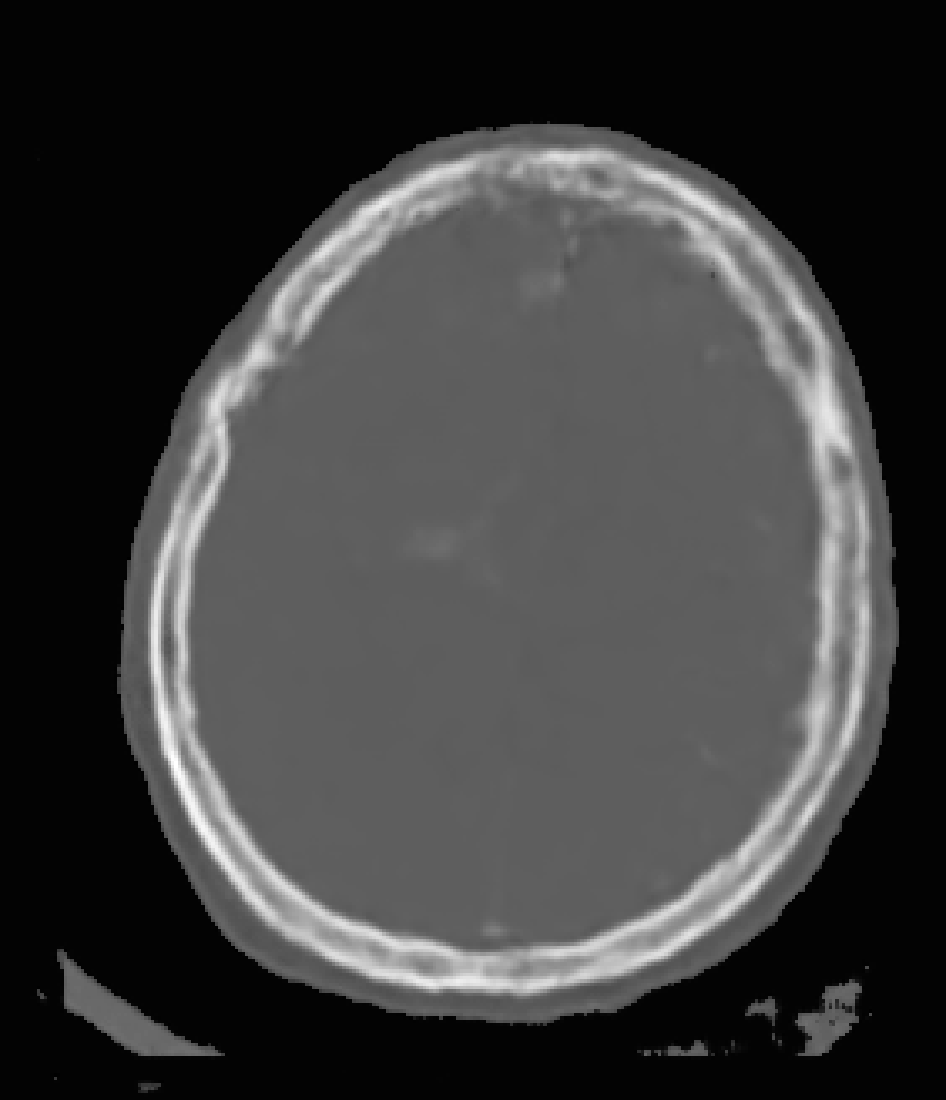}}%
\end{pgfscope}%
\begin{pgfscope}%
\pgfpathrectangle{\pgfqpoint{0.406740in}{3.898325in}}{\pgfqpoint{3.294941in}{3.665025in}}%
\pgfusepath{clip}%
\pgfsys@transformshift{0.406667in}{3.896667in}%
\pgftext[left,bottom]{\pgfimage[interpolate=true,width=3.296667in,height=3.663333in]{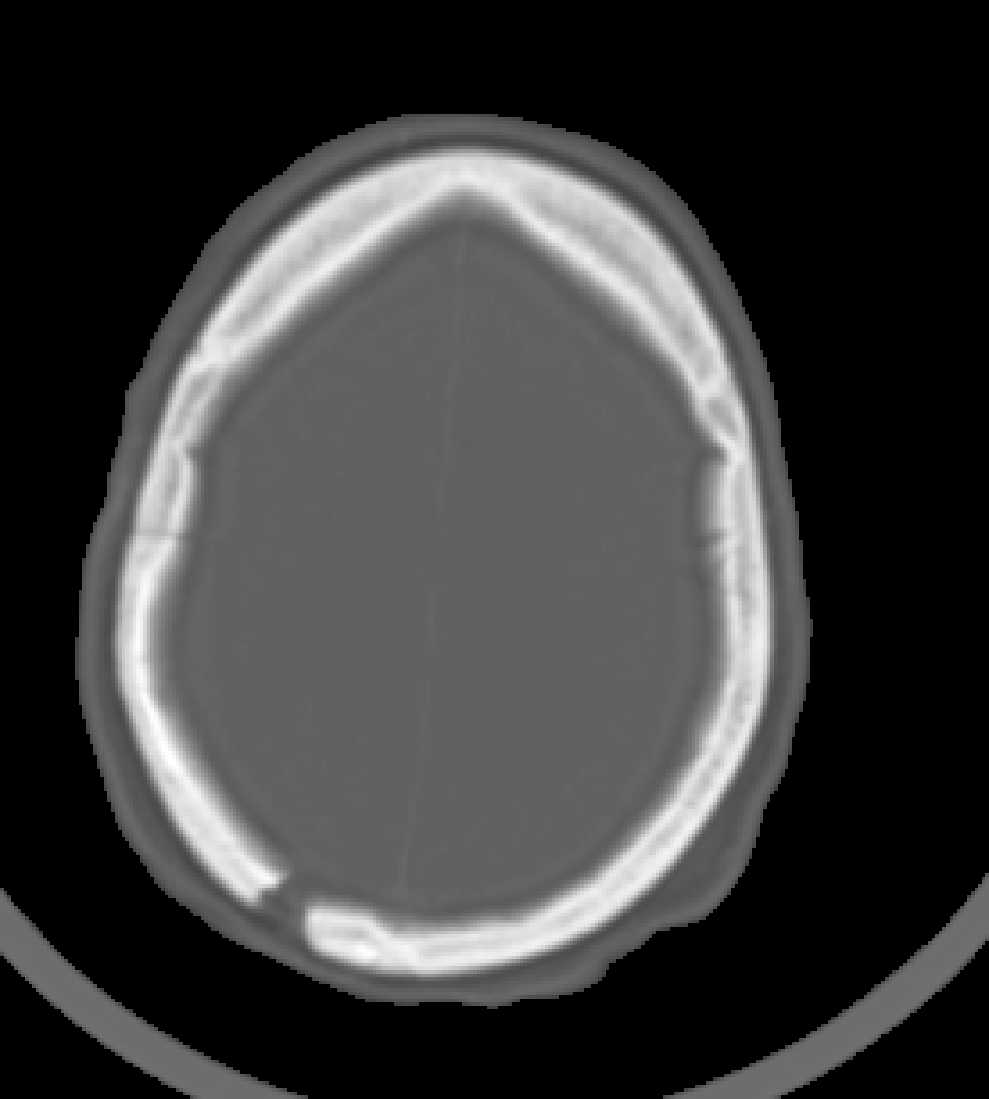}}%
\end{pgfscope}%
\begin{pgfscope}%
\pgfpathrectangle{\pgfqpoint{4.352530in}{3.898325in}}{\pgfqpoint{3.294941in}{3.665025in}}%
\pgfusepath{clip}%
\pgfsys@transformshift{4.353333in}{3.896667in}%
\pgftext[left,bottom]{\pgfimage[interpolate=true,width=3.296667in,height=3.663333in]{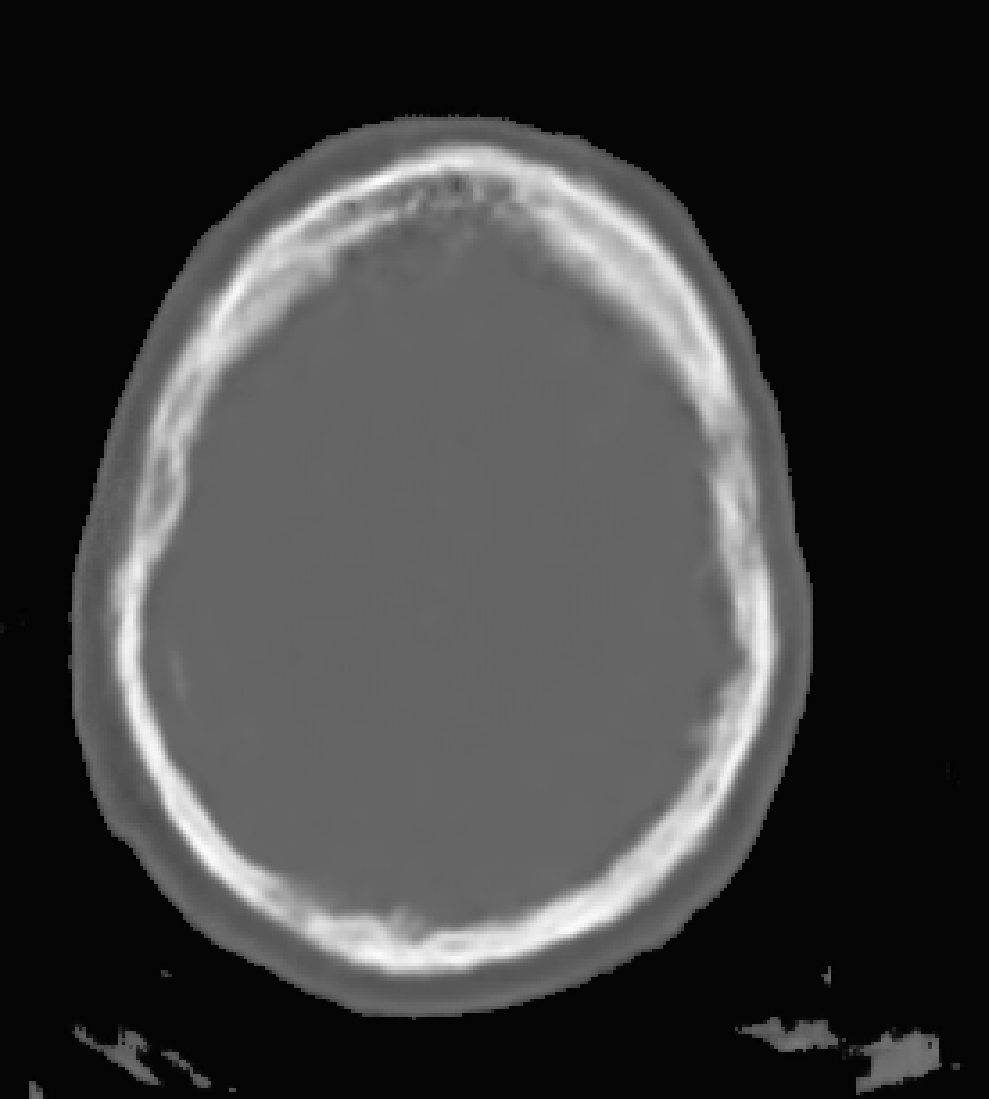}}%
\end{pgfscope}%
\begin{pgfscope}%
\pgfpathrectangle{\pgfqpoint{8.298319in}{3.898325in}}{\pgfqpoint{3.294941in}{3.665025in}}%
\pgfusepath{clip}%
\pgfsys@transformshift{8.296667in}{3.896667in}%
\pgftext[left,bottom]{\pgfimage[interpolate=true,width=3.296667in,height=3.663333in]{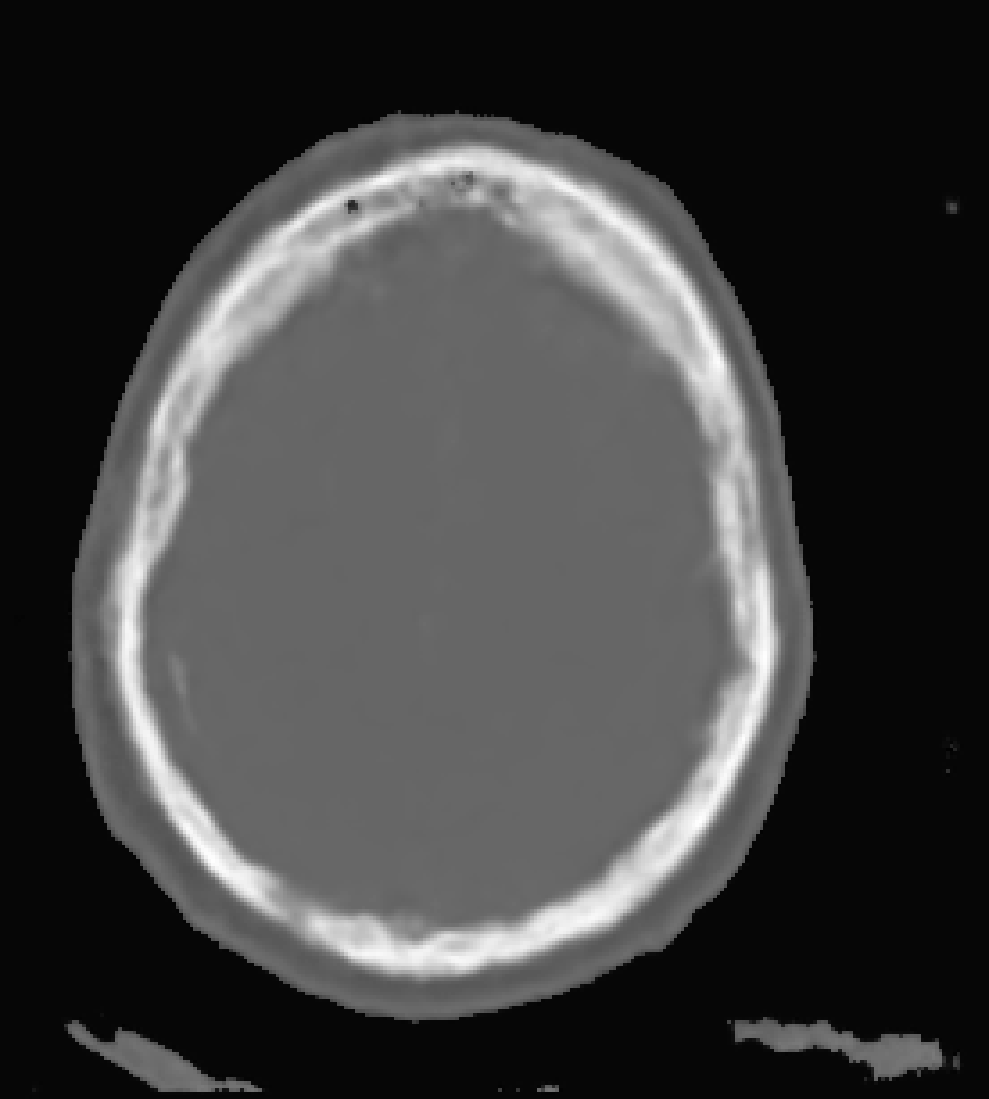}}%
\end{pgfscope}%
\begin{pgfscope}%
\pgfpathrectangle{\pgfqpoint{0.594885in}{0.160000in}}{\pgfqpoint{2.918652in}{3.665025in}}%
\pgfusepath{clip}%
\pgfsys@transformshift{0.593333in}{0.160000in}%
\pgftext[left,bottom]{\pgfimage[interpolate=true,width=2.916667in,height=3.666667in]{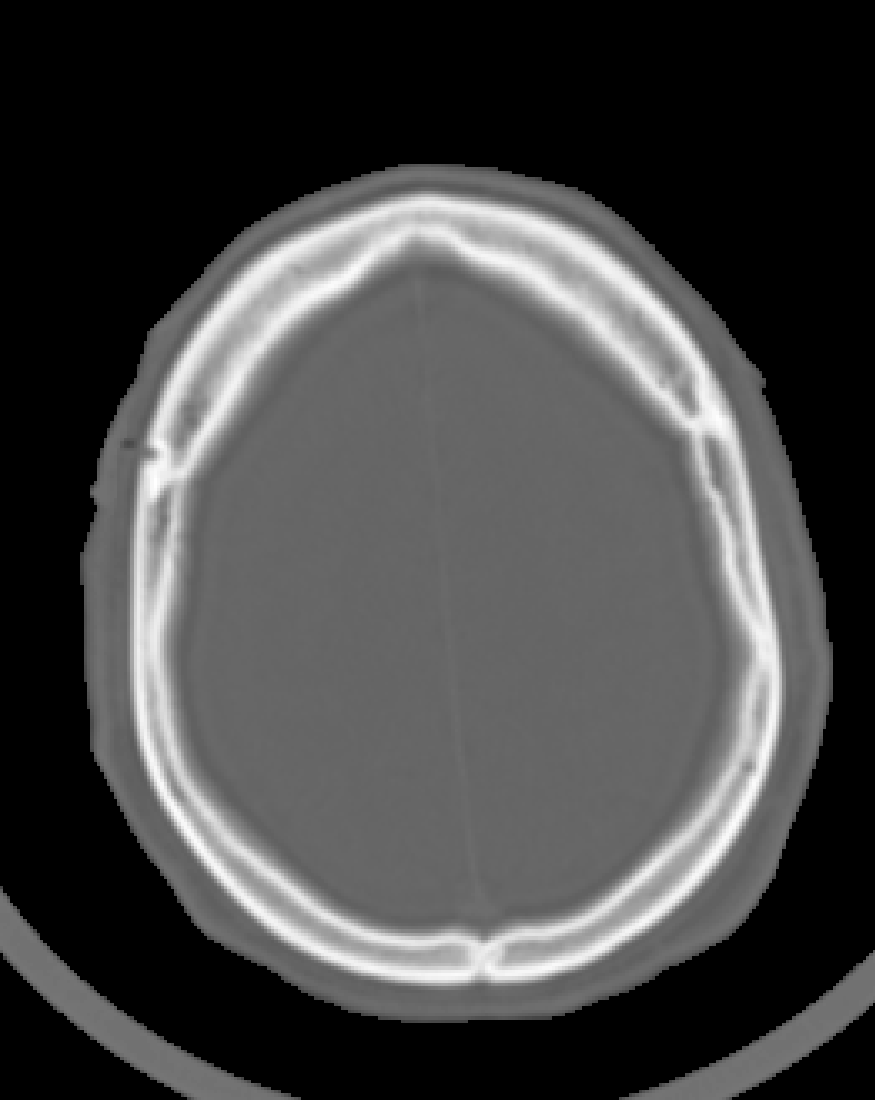}}%
\end{pgfscope}%
\begin{pgfscope}%
\pgfpathrectangle{\pgfqpoint{4.540674in}{0.160000in}}{\pgfqpoint{2.918652in}{3.665025in}}%
\pgfusepath{clip}%
\pgfsys@transformshift{4.540000in}{0.160000in}%
\pgftext[left,bottom]{\pgfimage[interpolate=true,width=2.920000in,height=3.666667in]{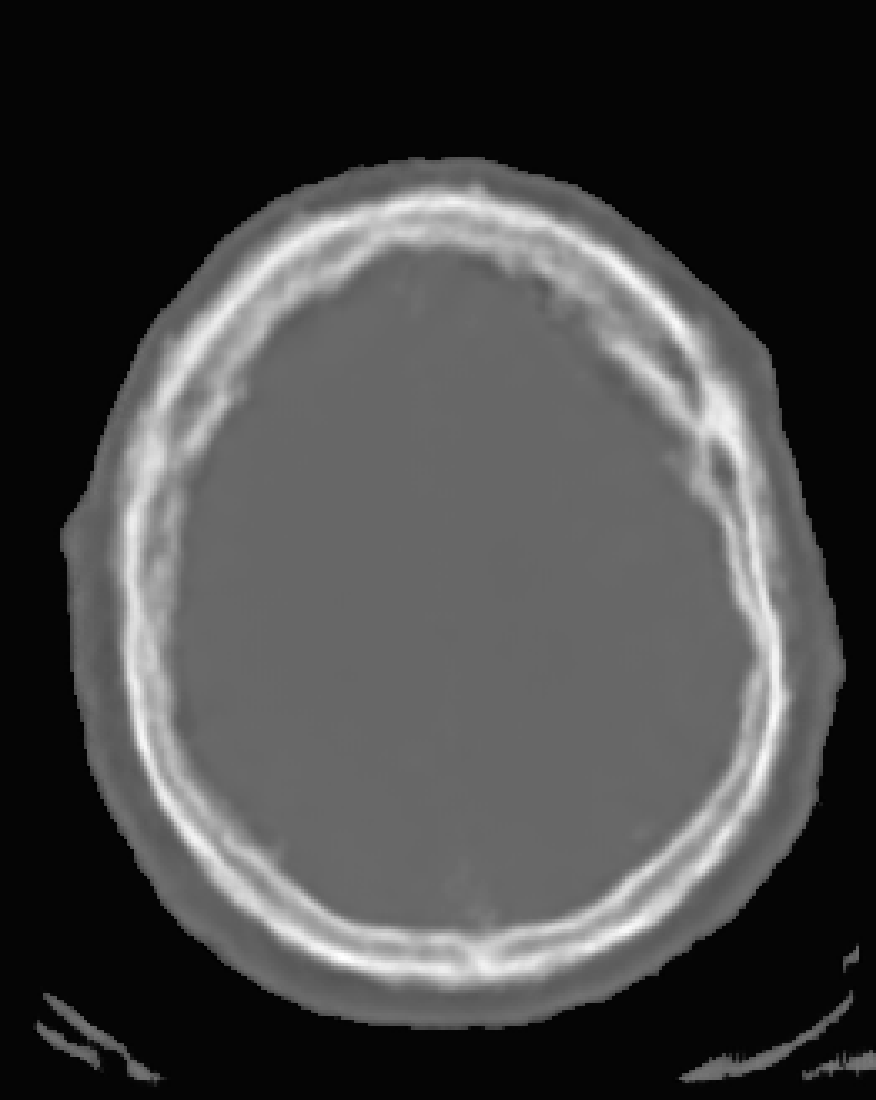}}%
\end{pgfscope}%
\begin{pgfscope}%
\pgfpathrectangle{\pgfqpoint{8.486464in}{0.160000in}}{\pgfqpoint{2.918652in}{3.665025in}}%
\pgfusepath{clip}%
\pgfsys@transformshift{8.486667in}{0.160000in}%
\pgftext[left,bottom]{\pgfimage[interpolate=true,width=2.920000in,height=3.666667in]{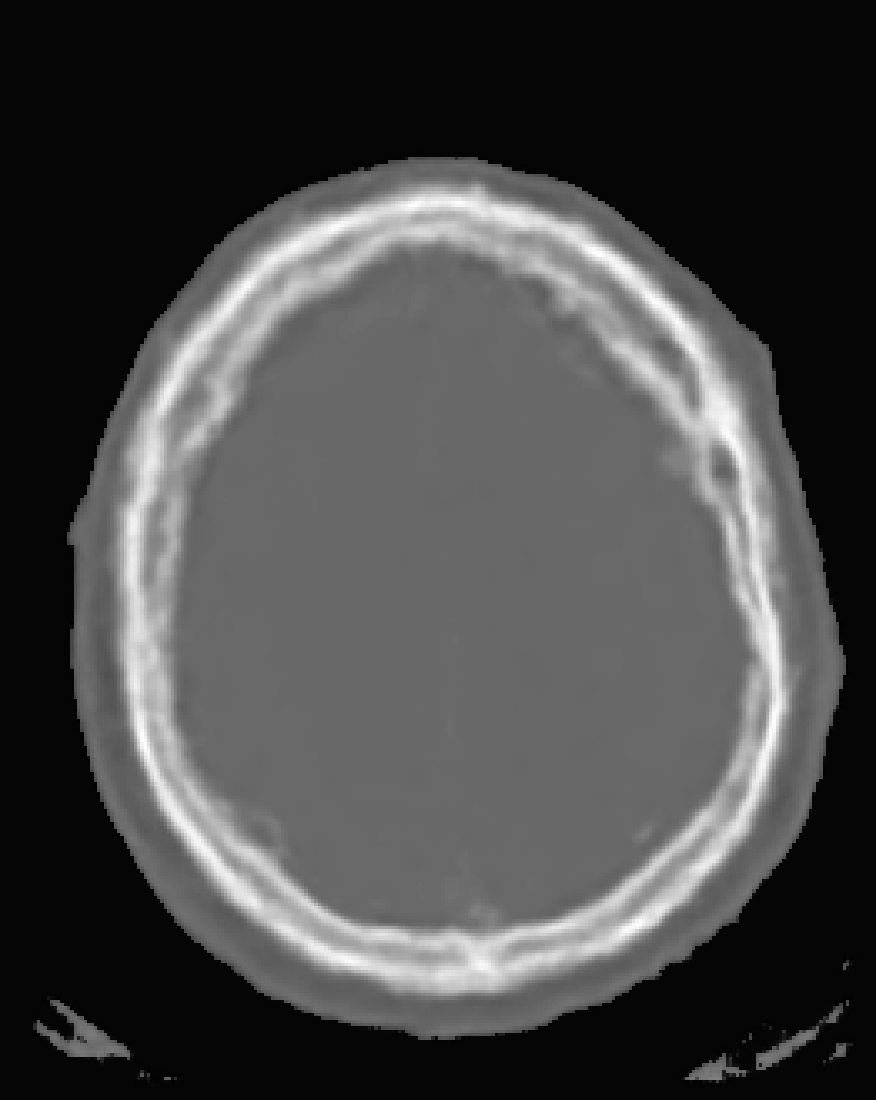}}%
\end{pgfscope}%
\end{pgfpicture}%
\makeatother%
\endgroup%
  \end{adjustbox}
  \caption{Transverse views for the pixtopix model trained with least-square
    adversarial loss compared to fine-tuned with gradient boost, evaluated on
    the validation dataset.
  }\label{fig:pixtopix:validation}
\end{figure}
In \Cref{fig:pixtopix:training} and \Cref{fig:pixtopix:validation} we show
the visual results of the two pixtopix variants. Altough some results, for
instance the last row from the training results, suggest an improved fine
structure of the soft tissue, we cannot definetly conclude that fine-tuning
with incresed soft-tissue weights yields better soft-tissue results, however,
we should keep in mind that we might not found the correct fine-tuning
parameters and that a further increase in order to compensate for the
exponential decay of the optimizer is necessary.

\section{Summary and outlook}

We outline in detail the preprocessing steps necessary to prepare a public
available dataset for the task of \gls{mri} to \gls{ct} translation.
Furtheremore we provide a reference implementation of different state of the
art models and compare obtained statistics and visual results. We believe
that the lack of sufficient (public) data is still a major holdback to this
specific computer vision task, which, however can be circumvented to a degree
through the input of more domain knowledge.

\section*{Acknowledgements}

With sincere gratitude, I want to thank Shadi Albarqouni from the Chair for
Computer Aided Medical Procedures at the Technical University Munich for the
fruitful discussions and guidance on this project.

\printbibliography{}

\end{document}